\documentclass[journal]{IEEEtran}
\usepackage[utf8]{inputenc}
\usepackage{amsmath,amsfonts,amssymb}
\usepackage{algorithmic}
\usepackage{algorithm}
\usepackage{array}
\usepackage{textcomp}
\usepackage{stfloats}
\usepackage{url}
\usepackage{verbatim}
\usepackage{graphicx}
\usepackage[backend=biber,style=ieee,sorting=none,bibencoding=utf8,doi=false]{biblatex}
\usepackage[nolist]{acronym}
\usepackage{hyperref}
\usepackage{bm}
\usepackage{siunitx}
\usepackage{xcolor}
\usepackage{tabularx}
\usepackage{threeparttable}
\usepackage{multirow}
\usepackage{multicol}
\usepackage{booktabs} % rulers 
\usepackage{arydshln} % hdashline etc...
\usepackage{listings}
\usepackage{matlab-prettifier}
\usepackage{pifont} % cmarks and xmarks
\usepackage[normalem]{ulem} % uwave for \pc
\usepackage[compact]{titlesec}
\usepackage{pgfplots}
\usepackage{pgfplotstable} % for pgfplotstabletranspose
\pgfplotsset{compat=newest}
\usepackage{tikz,tikz-3dplot}
\usetikzlibrary{fit,positioning,shapes,shapes.geometric,calc,backgrounds}
\usetikzlibrary{arrows,arrows.meta}
\usetikzlibrary{decorations.markings,decorations.pathreplacing,decorations.softclip,decorations.pathmorphing} % no fadings !
\usetikzlibrary{pgfplots.colorbrewer}
\usetikzlibrary{fadings}
\usetikzlibrary{patterns,patterns.meta}
\usepackage{balance}

\def\WithAuthorInfo{1}
\def\WithFigures{1}

\begin{acronym}[LEGO-LOAM]
\acro{ATE}{absolute trajectory error}
\acro{AVX}{advanced vector extensions}
\acro{CT-ICP}{Continuous-Time \acs{ICP}}
\acro{deque}{double-ended queue}
\acro{dof}[DoF]{degrees of freedom}
\acrodefplural{dof}{degrees of freedom}
\acro{DLO}{direct LiDAR odometry}
\acro{DLIO}{direct LiDAR-inertial odometry}
\acro{KF}{Kalman filter}
\acro{EKF}{extended \ac{KF}}
\acro{EM}{expectation maximization}
\acro{EWMA}{exponentially weighted moving average}
\acro{FMA}{Fused Multiply-Add}
\acro{fov}[FoV]{field-of-view}
\acro{ICP}{iterative closest point}
\acro{GICP}{generalized \acs{ICP}}
\acro{GMM}{Gaussian mixture model}
\acro{GN}{Gauss-Newton}
\acro{GP}{Gaussian process}
\acrodefplural{GP}{Gaussian processes}
\acro{gnss}[GNSS]{global navigation satellite system}
\acro{IMU}{inertial measurement unit}
\acro{iekf}[IEKF]{iterated error-state Kalman filter}
\acro{iou}[IoU]{intersection over union}
\acro{KLD}{Kullback-Leibler divergence}
\acro{LM}{Levenberg-Marquardt}
\acro{LO}{LiDAR odometry}
\acro{LIO}{LiDAR-inertial odometry}
\acro{LIO-MARS}{LiDAR-inertial odometry with MARS maps}
\acro{LIC}{LiDAR-inertial-camera odometry}
\acro{LiDAR}{Light Detection and Ranging}
\acro{LOAM}{localization and mapping}
\acro{MARS}{Multi-Adaptive-Resolution-Surfel}
\acro{mocap}[MoCap]{Motion Capture}
\acro{NDT}[NDT]{Normal distribution transform}
\acro{RAM}{random access memory}
\acro{RMS}{root-mean-squared}
\acro{RMSE}{\acs{RMS} error}
\acro{ROS}{robot operating system}
\acro{SAR}{search \& rescue}
\acro{SDE}{stochastic differential equations}
\acro{SIMD}{single instruction, multiple data}
\acro{SLAM}{simultaneous localization and mapping}
\acro{slerp}{spherical linear interpolation}
\acro{spd}[s.p.d.]{symmetric positive semidefinite}
\acro{sdf}[SDF]{signed distance function}
\acro{SE-LIO}{Semi-Elastic-\acs{LIO}}
\acro{surfel}{surface element}
\acro{TLS}{terrestrial \acs{LiDAR} scanner}
\acro{UAV}{unmanned aerial vehicle}
\acro{UGV}{unmanned ground vehicle}
\acro{UT}{unscented transform}
\acro{VIO}[VIO]{visual-inertial odometry}
\acro{VO}[VO]{visual odometry}
\acro{voxel}{volume element}
\end{acronym} 

\DeclareSIUnit\fps{fps}
\DeclareSIUnit\px{px}
\DeclareSIUnit{\nothing}{\relax}

\newcommand{\cmark}{\ding{51}}%
\newcommand{\xmark}{ }
\newcommand{\tbfX}{\textbf{X}}

\newcommand{\tbfZ}{\textbf{0}}

\newcommand{\tcb}[1]{\uline{#1}}
\newcommand{\tbf}[1]{\textbf{#1}}

\newcommand{\pa}[1]{\tbf{#1}}
\newcommand{\pb}[1]{\tcb{#1}}
\newcommand{\pc}[1]{\dotuline{#1}}

\newcommand{\norm}[1]{\left\lVert#1\right\rVert}
\newcommand{\abs}[1]{\lvert#1\rvert}

\DeclareMathOperator*{\argmin}{arg\,min}

\DeclareMathOperator{\diag}{diag}

\newcommand{\Reffig}[1]{Figure~\ref{#1}}

\newcommand{\Reftab}[1]{Table~\ref{#1}}

\newcommand{\reffig}[1]{Fig.~\ref{#1}}
\newcommand{\refsec}[1]{Sec.~\ref{#1}}

\newcommand{\reftab}[1]{Tab.~\ref{#1}}
\newcommand{\refeq}[1]{\eqref{#1}}

\newcommand{\citep}[1]{(\cite{#1})}

\newcommand{\wrt}{w.r.t.~}

\newcommand{\eg}{e.g.,\ }

\newcommand{\RX}{\mathbb{R}^9}
\newcommand{\RE}{\mathbb{R}^8}
\newcommand{\RD}{\mathbb{R}^3}
\newcommand{\SeD}{\mathrm{SE}(3)}
\newcommand{\SoD}{\mathrm{SO}(3)}
\newcommand{\RDSoR}{\RD\!\times\!\SoD\!\times\!\RD}
\newcommand{\SoR}{\SoD\!\times\!\RD}
\newcommand{\LOGD}{\mathrm{Log}_{D}}

\newcommand{\relPosPreN}{\Delta\widetilde{\bm{p}}}
\newcommand{\relPosPre}{\relPosPreN}

\newcommand{\relPosN}{\Delta{\bm{p}}}
\newcommand{\relPos}{\relPosN}

\newcommand{\relRotPreN}{\Delta\widetilde{R}}
\newcommand{\relRotPre}{\relRotPreN}

\newcommand{\relVelPreN}{\Delta\widetilde{\bm{v}}}
\newcommand{\relVelPre}{\relVelPreN}

\newcommand{\angRotM}{\bm{\omega}_{\mathrm{m}}}

\newcommand{\angAccZ}{\bm{\alpha}_\mathrm{z}}

\newcommand{\linAccM}{\bm{a}_{\mathrm{m}}}
\newcommand{\linAccZ}{\bm{a}_{\mathrm{z}}}

\newcommand{\drelPosPre}{\bm{d}_{\relPosPre}}
\newcommand{\drelRotPre}{\bm{d}_{\relRotPre}}
\newcommand{\drelVelPre}{\bm{d}_{\relVelPre}}

\newcommand{\JfXdX}[2]{\bm{J}^{#1}_{#2}}
\newcommand{\Jfxdx}{\JfXdX{f(X)}{X}}
\newcommand{\JrfxdX}[1]{\JfXdX{\mathrm{g}}{#1}}
\newcommand{\JrfxdXkij}[1]{\JfXdX{#1}{X_k}}

\newcommand{\limTau}[1]{\lim_{\bm{\tau} \rightarrow 0} \frac{#1}{\bm{\tau}}} 
\newcommand{\Rk}{R_{k}}
\newcommand{\Ri}{R_{i}}
\newcommand{\Rj}{R_{j}}

\newcommand{\Rit}{R_{i}^\intercal}
\newcommand{\Rjt}{R_{j}^\intercal}
\newcommand{\pos}{\bm{p}}
\newcommand{\posk}{\bm{p}_{k}}
\newcommand{\posi}{\bm{p}_{i}}
\newcommand{\posj}{\bm{p}_{j}}

\newcommand{\veli}{\bm{v}_{i}}
\newcommand{\velj}{\bm{v}_{j}}
\newcommand{\Dpm}{\Delta\bm{p}_\mathrm{m}}

\newcommand{\DRmt}{\Delta R_\mathrm{m}^\intercal}

\newcommand{\Dp}{\Delta\bm{p}}

\newcommand{\DR}{\Delta R}

\newcommand{\DT}{\Delta T}
\newcommand{\DTm}{\Delta T_\mathrm{m}}
\newcommand{\dDT}{\bm{d}_{\DT}}

\newcommand{\inv}{^{\scriptscriptstyle\text{-}1}}

\usepackage{adjustbox}
\newcolumntype{R}[2]{%
    >{\adjustbox{angle=#1,lap=\width-(#2)}\bgroup}%
    l%
    <{\egroup}%
}
\newcommand\RotD[1]{\rotatebox[origin=r]{90}{#1}}

\definecolor{my_darkgray}{RGB}{70, 70, 70}
\definecolor{Bed}{rgb}{0.         0.50196078 0.50196078}
\definecolor{Bookshelf}{rgb}{0.98039216 0.19607843 0.19607843}
\definecolor{Ceiling}{rgb}{0.4        0.         0.8       }
\definecolor{Chair}{rgb}{0.19607843 0.19607843 0.98039216}
\definecolor{Floor}{rgb}{0.8627451  0.8627451  0.8627451 }
\definecolor{Furniture}{rgb}{1.         0.27058824 0.07843137}
\definecolor{Objects}{rgb}{1.         0.07843137 0.49803922}
\definecolor{Picture}{rgb}{0.19607843 0.19607843 0.58823529}
\definecolor{Sofa}{rgb}{0.87058824 0.70588235 0.54901961}
\definecolor{Table}{rgb}{0.19607843 0.98039216 0.19607843}
\definecolor{Tv}{rgb}{1.         0.84313725 0.        }
\definecolor{Wall}{rgb}{0.58823529 0.58823529 0.58823529}
\definecolor{Window}{rgb}{0.         1.         1.        }

\renewcommand{\hdashline}{}
\renewcommand{\cdashline}[1]{}

\begin{document}
\title{\huge LIO-MARS: Non-uniform Continuous-time Trajectories\\for Real-time LiDAR-Inertial-Odometry}
\if\WithAuthorInfo1
\author{Jan Quenzel~${}^{a,b}$ and Sven Behnke~${}^{a,b,c}$%,~\IEEEmembership{~IEEE,}% <-this % stops a space
\thanks{${}^a$~Autonomous Intelligent Systems Group, Computer Science Institute VI -- Intelligent Systems and Robotics -- and ${}^b$~Center for Robotics and Lamarr Institute for Machine Learning and Artificial Intelligence, University of Bonn, Germany;
${}^c$~Fraunhofer IAIS, Germany;
{\tt\small quenzel@ais.uni-bonn.de}}% <-this % stops a space
\thanks{Manuscript received October XX, 2025; revised XX, 2025.}
\\\vspace*{-1cm}
}% <-this % stops a space
\else
\author{Anonymous Authors% <-this % stops a space
\thanks{Anonymous Group, Anonymous City;
{\tt\small anonymous email}}% <-this % stops a space
\thanks{Manuscript received October XX, 2025; revised XX, 2025.}
\\\vspace*{-1cm}
}% <-this % stops a space
\fi

%% The paper headers
\markboth{Journal of \LaTeX\ Class Files,~Vol.~X, No.~X, X~2025}%
{%
LIO-MARS: Non-uniform Continuous-time Trajectories\\for real-time LiDAR-Inertial-Odometry}
\IEEEpubid{0000--0000/00\$00.00~\copyright~2025 IEEE}
%% Remember, if you use this you must call \IEEEpubidadjcol in the second
%% column for its text to clear the IEEEpubid mark.
\maketitle
\begin{abstract}
Autonomous robotic systems heavily rely on environment knowledge to safely navigate. For search \& rescue, a flying robot requires robust real-time perception, enabled by complementary sensors. IMU data constrains acceleration and rotation, whereas LiDAR measures accurate distances around the robot.
Building upon the LiDAR odometry MARS, our LiDAR-inertial odometry (LIO) jointly aligns multi-resolution surfel maps with a Gaussian mixture model (GMM) using a continuous-time B-spline trajectory.
Our new scan window uses non-uniform temporal knot placement to ensure continuity over the whole trajectory without additional scan delay. Moreover, we accelerate essential covariance and GMM computations with Kronecker sums and products by a factor of 3.3. An unscented transform de-skews surfels, while a splitting into intra-scan segments facilitates motion compensation during spline optimization.
Complementary soft constraints on relative poses and preintegrated IMU pseudo-measurements further improve robustness and accuracy.
Extensive evaluation showcases the state-of-the-art quality of our LIO-MARS \wrt recent LIO systems on various handheld, ground and aerial vehicle-based datasets.
\end{abstract}


\section{Introduction}
\IEEEPARstart{R}{eliable} real-time perception is essential for robotic autonomy.
In particular, accurate mapping and ego-motion estimation are key components for safe interaction in complex and unstructured environments.
Due to their precision and measurement density, modern {LiDARs} are often used in these scenarios, \eg in the DARPA Subterranean Challenge~\cite{CerberusSubT,ExplorerSubT}.

Sensor motion during scanning distorts the point cloud and degrades the quality of the map.
This intra-scan motion is either compensated by de-skewing prior to registration~\cite{quenzel2021mars,vizzo2022kissicp,xu2021fastlio2,chen2023dlio} or by modeling it with a continuous-time trajectory~\cite{dellenbach2021ct,lv2021clins,david2018icra}.
The former uses the previous state estimate and, optionally, an \acs{IMU} to predict the motion and transform points to a common reference time.
The latter approach optimizes the trajectory directly at intermediate time steps.
However, this comes at the cost of reduced real-time capability and requires either costly reintegration of \acsp{surfel}~\cite{david2018icra} or a limited number of selected pointwise features [\eg \acs{CT-ICP}~\cite{dellenbach2021ct}, CLINS~\cite{lv2021clins}].

\begin{figure}
\centering
\if\WithFigures1
\resizebox{1.0\linewidth}{!}{\begin{tikzpicture}[	>={Stealth[inset=0pt,length=4pt,angle'=45]}]
\definecolor{red}{rgb}{0.7,0.0,0.0}
\definecolor{blue}{rgb}{0.2,0.2,0.7}
% trim={<left> <lower> <right> <upper>}

\pgfmathsetmacro{\dH}{1.5cm}
\pgfmathsetmacro{\dV}{0.75cm}
\pgfmathsetmacro{\dr}{0.082*\dV}

%https://tex.stackexchange.com/questions/141378/path-following-color-gradient-in-tikz
\definecolor{p0}{rgb}{0.050383,0.029803,0.527975}
\definecolor{p1}{rgb}{0.267703,0.012716,0.620346}
\definecolor{p2}{rgb}{0.435734,0.001127,0.659797}
\definecolor{p3}{rgb}{0.589719,0.072878,0.630408}
\definecolor{p4}{rgb}{0.714883,0.187299,0.546338}
\definecolor{p5}{rgb}{0.819651,0.306812,0.448306}
\definecolor{p6}{rgb}{0.904601,0.429797,0.356329}
\definecolor{p7}{rgb}{0.966798,0.564396,0.265118}
\definecolor{p8}{rgb}{0.994324,0.716681,0.177208}
\definecolor{p9}{rgb}{0.940015,0.975158,0.131326}
\pgfdeclareverticalshading{plasma}{100bp}{
  color(0bp)=(p0);color(11bp)=(p1);color(21bp)=(p2);
  color(31bp)=(p3);color(42bp)=(p4);color(53bp)=(p5);
  color(63bp)=(p6);color(74bp)=(p7); color(84bp)=(p8); color(100bp)=(p9)
}

\node(P0)[fill=none] at (0,0) {};

\pgfmathsetmacro{\mv}{-20}
\pgfmathsetmacro{\sv}{1.25}
\pgfmathsetmacro{\rv}{4}
\pgfmathsetmacro{\lv}{9}

\pgfmathsetmacro{\cx}{0} 
\pgfmathsetmacro{\cy}{-1.55} 
\pgfmathsetmacro{\mvi}{9*\lv+4}
\pgfmathsetmacro{\llv}{18*\lv-\mvi-5}
\foreach \i in {\llv,...,\lv}
{
  \pgfmathsetmacro{\t}{-\sv*(\i+\mv+\mvi)}
  \pgfmathsetmacro{\x}{\rv*cos(\t)        /(1+sin(\t)*sin(\t)) + \cx}
  \pgfmathsetmacro{\y}{\rv*sin(\t)*cos(\t)/(1+sin(\t)*sin(\t)) + \cy}
  \node (O\i)[draw=none,fill=none,outer sep=0em,inner sep=0em] at (\x,\y){};
}

\foreach \z in {4,...,6}
{
 \foreach \xi/\mc in {2/0,3/1,4/3,5/5,6/7,7/8}
 {
  \begin{scope}
  \pgfmathtruncatemacro{\x}{\xi + \z * \lv }
  \pgfmathtruncatemacro{\mj}{\mc + 1 }
  \pgfmathtruncatemacro{\a}{\x - 2 }
  \pgfmathtruncatemacro{\b}{\x - 1 }
  \pgfmathtruncatemacro{\c}{\x + 1 }
  \pgfmathtruncatemacro{\d}{\x + 2 }
  \node(A0)[scale=0.5,below left of=O\b]{};
  \node(A2)[scale=0.5,above right of=O\c]{};
  \draw [clip,draw=none] (A0) rectangle (A2);
  \draw[p\mc,path fading=east,postaction={draw, p\mj, path fading=west},very thick] 
  plot [smooth,tension=1] coordinates {(O\a)(O\b)(O\x)(O\c)(O\d)};
  \end{scope}
 }
}

\node(Ps0)[fill=none] at (O18){};
\node(Ps1)[fill=none] at (O27){};
\node(Ps2)[fill=none] at (O36){};
\node(Ps3)[fill=none] at (O45){};
\node(Ps4)[fill=none] at (O54){};
\node(Ps5)[fill=none] at (O63){};
\foreach \i in {2,...,7}
{
 \pgfmathtruncatemacro{\n}{\i * \lv}
 \draw[fill=blue] (O\n) circle (0.2em);
}

\foreach \z in {2,3}%0,1}
{
 \foreach \xi in {2,...,7}
 {
  \begin{scope}
  \pgfmathtruncatemacro{\x}{\xi + \z * \lv }
  \pgfmathtruncatemacro{\a}{\x - 2 }
  \pgfmathtruncatemacro{\b}{\x - 1 }
  \pgfmathtruncatemacro{\c}{\x + 1 }
  \pgfmathtruncatemacro{\d}{\x + 2 }
  \node(A0)[scale=0.5,below left of=O\b]{};
  \node(A2)[scale=0.5,above right of=O\c]{};
  \draw [clip,draw=none] (A0) rectangle (A2);
  \draw[blue,very thick] plot [smooth,tension=1] coordinates {(O\a)(O\b)(O\x)(O\c)(O\d)};
  \end{scope}
 }
}

\begin{scope}
 \pgfmathtruncatemacro{\x}{69}
 \pgfmathtruncatemacro{\r}{\x - 5 }
 \pgfmathtruncatemacro{\q}{\x - 4 }
 \pgfmathtruncatemacro{\o}{\x - 3 }
 \pgfmathtruncatemacro{\a}{\x - 2 }
 \pgfmathtruncatemacro{\b}{\x - 1 }
 \pgfmathtruncatemacro{\c}{\x + 1 }
 \pgfmathtruncatemacro{\d}{\x + 2 }
 \node(A0)[xshift=-0.1*\dH,yshift=-0.25*\dV] at (O\x){};
 \node(A2)[xshift=0.25*\dH,yshift=0.5*\dV] at (O\x){};
 \draw [clip,draw=none] (A0) rectangle (A2);
 \draw[->,green!60!black,very thick] plot [smooth,tension=1] 
 coordinates {(O\r)(O\q)(O\o)(O\a)(O\b)(O\x)(O\c)(O\d)};
\end{scope}

\node(CurScan)[draw=blue,ellipse,rotate=15,fit=(O54)(O63),outer sep=0em,inner sep=0em]{};

\pgfmathsetmacro{\cx}{0}
\pgfmathsetmacro{\cy}{-0.65}

\pgfmathsetmacro{\crv}{25}
\pgfmathsetmacro{\ccx}{-8.5 + \cx}
\pgfmathsetmacro{\ccy}{-7 + \cy}
\foreach \i in {52,...,64}
{
  \pgfmathsetmacro{\t}{\sv*(\i+\mv)}
  \pgfmathsetmacro{\x}{\crv*cos(\t)        /(1+sin(\t)*sin(\t)) + \ccx}
  \pgfmathsetmacro{\y}{\crv*sin(\t)*cos(\t)/(1+sin(\t)*sin(\t)) + \ccy}
  \node (H\i)[draw=none,fill=none,outer sep=0em,inner sep=0em] at (\x,\y){};
}

\begin{scope}
 \pgfmathsetmacro{\z}{6}
 \foreach \xi in {0,...,8} 
 {
  \begin{scope}
   \pgfmathsetmacro{\mc}{\xi}
   \pgfmathtruncatemacro{\x}{\xi + \z * \lv }
   \pgfmathtruncatemacro{\mj}{\mc + 1 }
   \pgfmathtruncatemacro{\a}{\x - 2 }
   \pgfmathtruncatemacro{\b}{\x - 1 }
   \pgfmathtruncatemacro{\c}{\x + 1 }
   \pgfmathtruncatemacro{\d}{\x + 2 }
   \node(A0) at (H\b){};
   \node(A2) at (H\c){};
      
   \draw [clip,draw=none] (A0) rectangle (A2);
   \draw[p\mc,path fading=east,postaction={draw, p\mj, path fading=west},very thick]
   plot [smooth,tension=1] coordinates {(H\a)(H\b)(H\x)(H\c)(H\d)};
  \end{scope}
 }
\end{scope}

\node(Hs4)[fill=none,font=\tiny\sffamily, xshift=0.15*\dH, yshift=0.3*\dV] at (H54){$s_{j-1}$};
\node(Hs5)[fill=none,font=\tiny\sffamily, xshift=0.05*\dH, yshift=0.3*\dV] at (H63){$s_j$};
\foreach \i in {55,...,62}
 \draw[fill=red,draw=none] (H\i) circle (0.15em);
\draw[fill=blue,draw=none] (H54) circle (0.3em);
\draw[fill=blue,draw=none] (H63) circle (0.3em);

% rectangle on surfels/scan

% rounded braces... https://tex.stackexchange.com/questions/492887/curved-brace-decoration
\draw[-,very thick,decorate,decoration={brace, mirror,raise=0.15cm},font=\tiny\sffamily] (O63) -- node[below,midway,yshift=-0.2cm]{optimized scans}(O36);
\draw[-,very thick,decorate,decoration={brace,raise=0.2cm},font=\tiny\sffamily] (O36) -- node[above,midway,xshift=0.25*\dH,yshift=0.25*\dV]{fixed}(O18);

\node[inner sep=0,outer sep=0,xshift=0*\dH,yshift=-1.*\dV,font=\tiny\sffamily] at (O63.south east) {\underline{Trajectory}};

\node(H60S)[xshift=0.475*\dH,yshift=0.2*\dV] at(H60.center){};
\draw[<->,shorten >= 2pt,shorten <= 4pt,thick] (H54.180) to [out=165,in=75] node[midway,above,xshift=-0.2*\dV,font=\tiny\sffamily]{$\Delta R, \Delta \bm{p}, \Delta \bm{v}$}(H60S.90);
\node(H61S)[xshift=0.375*\dH,yshift=0.5*\dV] at(H61.center){};
\draw[<->,shorten >= 2pt,shorten <= 2pt,thick] (H53.270) to [out=270,in=320] node[midway,below,yshift=-0.1*\dV,font=\tiny\sffamily]{$\Delta T$}(H61S);

\draw[-,thick,decorate,decoration={brace,raise=0.1cm}] (H56) -- node[below,midway,yshift=-0.2*\dV,font=\tiny\sffamily]{segment}(H57);

\node(M7)[fill=none,above of=H61,xshift=-0.15*\dH,yshift=-0.0*\dV,font=\tiny\sffamily] {$\bm{\alpha},\bm{\omega}$};
\node(P7)[fill=none,below of=M7,yshift=-0.3*\dV]{};
\node(P75)[fill=none,yshift=-0.2*\dV,xshift=-0.25*\dH] at (H62) {};
\node(M75)[fill=none,yshift=-1.0*\dV,xshift=0.15*\dH,font=\tiny\sffamily] at (P75){$\bm{\alpha}_0$};
\node(M53)[fill=none,yshift=0.5*\dV] at (H53) {};

\draw[->,shorten >= 4pt,thick] (M7.270) to [out=270,in=90] (P7.90);
\draw[->,thick] (M75) to [out=90,in=270]([xshift=0.15*\dH]P75);

\node(CloseUp)[draw=blue,rectangle,fit=(H52)(P75)(M75)(H63)(M53),outer sep=0em,inner sep=0em]{};
\node(Constraints)[rectangle,anchor=south,font=\tiny\sffamily] at (CloseUp.south) {\underline{Constraints}};

\draw[->,draw=blue,thick] (CloseUp.west) to [out=180,in=20](CurScan);

% https://tex.stackexchange.com/questions/66490/drawing-a-tikz-arc-specifying-the-center
% https://tex.stackexchange.com/questions/123158/tikz-using-the-ellipse-command-with-a-start-and-end-angle-instead-of-an-arc
\tikzset{
  pics/carc/.style args={#1:#2:#3}{
    code={
      \draw[pic actions] (#1:#3) arc(#1:#2:#3);
    }
  },
  partial ellipse/.style args={#1:#2:#3}{
        insert path={+ (#1:#3) arc (#1:#2:#3)}
  }
}

% spinning LiDAR for time
\pgfmathsetmacro{\dr}{0.05*\dV}
\pgfmathsetmacro{\dq}{0.4*\dr}
\node(SC)[fill=none,above= of P0,yshift=-0.35*\dV,xshift=-1.75*\dH] {};
\node(SCB)[fill=none,yshift=-0.6*\dV] at (SC) {};
\node(SCU)[fill=none,yshift=0.6*\dV] at (SC) {};
\draw[-,densely dashed](SCB)--(SCU);
\node(LIDAR)[cylinder,draw,yshift=-0.1*\dV,aspect=0.5,rotate=90] at (SC.center){};
\node(MU)[fill=none,xshift=\dr cm,yshift=\dq cm] at (SC.center){};
\node(MB)[fill=none,xshift=\dr cm,yshift=-\dq cm] at (SC.center){};
\foreach \i in {-4,-2,0,2,4}
{
 \pgfmathsetmacro{\dqv}{\dq/4 * (\i)}
 \node(MB\i)[yshift=\dqv cm,xshift=\dr cm + 0.5*\dV] at (LIDAR.center) {};
 \draw[black!50!white,-{Stealth[inset=0pt,length=2pt,angle'=45]},thin]
 (LIDAR.center) -- (MB\i);
}
\foreach \i in {0,...,\lv}
{
 \pgfmathsetmacro{\doff}{3}
 \pgfmathtruncatemacro{\do}{(360/\lv)}
 \pgfmathtruncatemacro{\da}{-(\i+\doff) * \do}
 \pgfmathtruncatemacro{\de}{-(\i+1+\doff) * \do)}
 \draw[-,ultra thick,opacity=1,draw=p\i] (SC.center) 
 [partial ellipse=\da:\de: \dr cm and \dq cm];
 \pgfmathparse{\i==1}
 \ifnum\pgfmathresult=1 
  \draw[->,thick,opacity=1] (SC.center) [partial ellipse=\da:\de-\do:0.8*\dr cm and 0.8*\dq cm];
 \fi
}
\node[xshift=-0.8*\dV,yshift=0.16*\dV,font=\tiny\sffamily] at (LIDAR){$t$};
\node[anchor=north west,xshift=-0.975*\dH,yshift=-0.3*\dV,font=\tiny\sffamily] at (LIDAR.south west){\underline {Scan time}};

\end{tikzpicture}}
\fi
\caption[Non-uniform continuous-time trajectory and motion compensation]{Joint registration of \acs{LiDAR} scans embedded in multi-resolution \acs{surfel} maps (colored by relative scan time) optimizes a \textit{non-uniform} continuous-time B-spline trajectory. An \ac{UT} compensates motion within a \acs{surfel}, while a temporal partitioning into segments enables optimization of motion distortion between \acsp{surfel} intrinsically. Inclusion of relative ($\Delta R,\Delta \bm{v},\Delta\bm{p},\Delta T$) and absolute ($\bm{\alpha}_0,\bm{\alpha},\bm{\omega}$) soft-constraints further improves robustness in challenging situations.}
\label{fig:lio_teaser}
\end{figure}
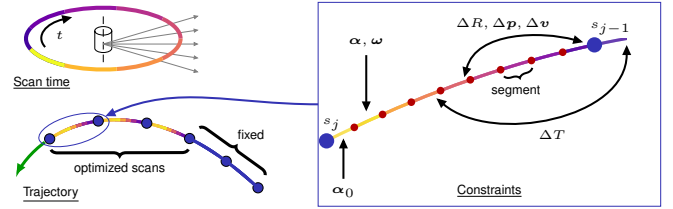

To overcome these limitations of continuous-time methods, our novel real-time \ac{LIO} jointly optimizes temporally partitioned scan segments (\reffig{fig:lio_teaser}) by registering multi-resolution \acs{surfel} maps while an \acf{UT} compensates the intra-\acs{surfel} motion.
Rephrasing the computation of the \ac{GMM} and \acsp{surfel} using vectorized Kronecker sums and products~\cite{lancaster1985kron,KRON} reduces redundancy and improves processing speed.
We introduce \textit{relative} inertial and motion constraints from complementary modalities, such as \acs{IMU} and robot odometry, and derive their analytic Jacobians \wrt the spline knots to increase robustness and accuracy.
Furthermore, we use a \textit{non-uniform} continuous-time B-spline trajectory as an elegant solution to address variations in scan time without increased delay [\eg as in Coco-\acs{LIC}~\cite{lang2023cocolic}] at greater numerical stability compared to its uniform counterpart.

\noindent We thoroughly evaluate the proposed \acs{LIO-MARS} to support our key claims, which are:
\begin{enumerate}
\setlength{\itemindent}{0em}
\item The non-uniform spline has improved numerical stability in real robotics applications.
\item An \acf{UT} enables motion compensation for individual \acsp{surfel}.
\item Temporal separation into intra-scan segments facilitates motion compensation at optimization time.
\item Leveraging relative inertial and motion constraints improves accuracy.
\item Rephrasing the \acf{GMM} and \acs{surfel} covariances with Kronecker sum and products improves parallelization.
\end{enumerate}
\if\WithAuthorInfo1
We will open-source \acs{LIO-MARS} at: \url{https://github.com/AIS-Bonn/lio_mars}.
\else
We will open-source \acs{LIO-MARS}.
\fi
\IEEEpubidadjcol
\section{Related Work}
\begin{figure*}
\centering
\if\WithFigures1
 \resizebox{1.0\linewidth}{!}{\begin{tikzpicture}
[content_node/.append style={font=\sffamily,minimum size=1.5em,minimum width=7.1em,draw,align=center,rounded corners,scale=0.65},
vis_node/.append style={font=\sffamily,minimum size=1.5em,minimum width=1.0em,draw,align=center,rounded corners,scale=0.65},
label_node/.append style={font=\sffamily,scale=0.5},
group_node/.append style={font=\sffamily,dotted,align=center,rounded corners,inner sep=0.1em,thick,opacity=0.25},
>={Stealth[inset=0pt,length=4pt,angle'=45]},
node distance=1.5cm]

\pgfmathsetmacro{\dH}{1.50cm}
\pgfmathsetmacro{\dV}{0.75cm}

\definecolor{red}{rgb}     {0.5,0.0,0.0}
\definecolor{green}{rgb}   {0.0,0.5,0.0}
\definecolor{blue}{rgb}    {0.0,0.0,0.5}
\definecolor{grey}{rgb}    {0.5,0.5,0.5}

\node(LIDAR)[content_node,fill=green!15!white] at (0,0) {LiDAR};
\node(IMU)[content_node,fill=green!15!white, right of=LIDAR,xshift=1.75*\dH] {IMU};
\node(ODOM)[content_node,fill=green!15!white,right of=IMU,xshift=1.75*\dH] {Odometry};

\node(ORICOMP)[content_node,fill=blue!15!white,below of=LIDAR,yshift=-0.25*\dV] {Orientation\\Compensation};
\node[group_node,fill=none,fit={(ORICOMP)}] {};

\node(LATTICE)[content_node,fill=blue!15!white,below of=ORICOMP,yshift=-0.25*\dV] {Lattice\\Embedding};
\node[group_node,fill=none,fit={(LATTICE)}] {};

\node(PRIOR)[content_node,fill=blue!15!white,below of=IMU,yshift=-0.25*\dV] {Prior\\Generation};
\node(INIT)[content_node,fill=blue!15!white,below of=ODOM,yshift=-0.25*\dV] {Spline\\Initialization};

\node(SCENE)[content_node,fill=blue!15!white,below of=PRIOR,yshift=-0.25*\dV] {Registration\\Window};
\node(SN)[below=of SCENE.south west,anchor=north west,vis_node,minimum width=7.1em,fill=red!15!white,yshift=3*\dV] {$\mathcal{S}_{l-L+1} \ldots \mathcal{S}_l$};
\node[group_node,fill=none,fit={(SCENE)(SN)}] {};

\node(UT)[content_node,fill=blue!15!white,below of=INIT,yshift=-0.25*\dV] {Surfel\\Compensation};
\node(MAPH)[content_node,fill=blue!15!white,right of=INIT,xshift=1.75*\dH,yshift=2*\dV] {Local\\Map};
\node(KFX)[below=of MAPH.south,anchor=north,vis_node,fill=red!15!white,minimum width=7.1em,yshift=3*\dV] {\tiny $\left\lbrace \mathcal{S}_{K_x} \ldots \mathcal{S}_{K_y}\right\rbrace$};
\node(MAP)[group_node,fill=none,fit={(MAPH)(KFX)}] {};

\node(REG)[content_node,fill=blue!15!white,below of=MAPH,yshift=-1.*\dV] {Spline\\Registration};
\node(UPD)[content_node,fill=blue!15!white,right of=MAPH,xshift=1.75*\dH] {Map\\ Update?};
\node(KFDBH)[content_node,fill=blue!15!white,right of=UPD,xshift=1.75*\dH] {Keyframe\\Storage};
\node(KF)[content_node,fill=blue!15!white,below of=UPD,yshift=-3*\dV] {Create\\ Keyframe?};
\node(KFLE)[content_node,fill=blue!15!white,below of=KFDBH,yshift=-\dV] {Lattice\\ Embedding};
\node(PWC)[content_node,fill=blue!15!white,below of=KFLE,yshift=-0.*\dH] {Pointwise\\Compensation};
\node(DBSym)[below=of KFDBH.south,anchor=north,vis_node,fill=red!15!white,minimum width=7.1em,yshift=3*\dV] {};
%https://tex.stackexchange.com/questions/442991/database-shape-in-tikz
\node(DBS)[cylinder,draw,yshift=-0.05*\dV,aspect=0.5,rotate=90,path picture={
\draw (path picture bounding box.170) to[out=180,in=180] (path picture bounding
box.10);
\draw (path picture bounding box.190) to[out=180,in=180] (path picture bounding
box.350);},yscale=1.5,xscale=0.75,] at (DBSym.center){};
\node(KFDB)[group_node,fill=none,fit={(KFDBH)(DBSym)}] {};

\draw[->,thick] (LIDAR.270) --node[midway,right,xshift=-0.1cm]{\tiny $\mathcal{P}_l$} (ORICOMP.90);
\draw[->,thick] (IMU.270) to [out=220,in=30] node[midway,above,xshift=-0.05cm]{\tiny $\angRotM$}(ORICOMP.45);
\draw[->,thick] (IMU.270) -- node[midway,yshift=-0.05cm]{\tiny $\angRotM,\linAccM$}(PRIOR.90);
\draw[->,thick] (ODOM.270)  to [out=220,in=30] node[midway,above]{\tiny $T_{\mathrm{m}}$}(PRIOR.30);
\draw[->,thick] (ORICOMP.270) -- node[midway,left]{\tiny $\widetilde{\mathcal{P}}_l$}(LATTICE.90);
\draw[->,thick] (LATTICE.10) to [out=0,in=210] node[midway,above ,yshift=0.2cm]{\tiny $C_l$}(PRIOR.180);
\draw[->,thick] (LATTICE.0) -- node[midway,above,yshift=-0.1cm]{\tiny $\mathcal{S}_l$ }(SCENE.180);
\draw[->,thick] (PRIOR.0)  -- node[midway,above]{\tiny $\Delta T$} node[midway,below]{\tiny $\Delta_{pre}$}(INIT.180);
\draw[->,thick] (INIT.270) -- node[midway,right]{\tiny $\mathcal{X}$}(UT.90);
\draw[->,thick] (SCENE.0) -- node[midway,above]{\tiny $\mathcal{W}_l$}(UT.180);
\draw[->,thick] (UT.0) --  node[midway,above,xshift=-0.1cm]{\tiny $\overline{\mathcal{W}}_l$}(REG.180);
\draw[->,thick] (MAP.270) -- node[midway,right]{\tiny $\mathcal{M}$}(REG.90);
\draw[->,thick] (REG.0) -- (KF.170);
\draw[->,thick] (REG.0) to [out=45,in=270] (UPD.270);
\draw[->,thick] (KF.0) -- node[midway,above]{\tiny $\mathcal{P}_{l-L+1}$}(PWC.180);
\draw[->,thick] (SCENE.350) to [out=340,in=180] node[midway,above right]{\tiny $\mathcal{P}_{l-L+1}$}(KF.180);
\draw[->,thick] (PWC.90) -- node[midway,left,yshift=0.05cm]{\tiny $\overline{\mathcal{P}}_{l-L+1}$}(KFLE.270);
\draw[->,thick] (KFLE.90) -- node[midway,left,yshift=0.05cm]{\tiny $\overline{\mathcal{S}}_{l-L+1}$}(KFDB.270);
\draw[->,thick] (KFDBH.180) -- node[midway,below]{\tiny $\left\lbrace\mathcal{S}_{\mathrm{F}}\right\rbrace$}(UPD.0);
\draw[->,thick] (UPD.180) -- (MAPH.0);

\node(LORI)[fill=none,left of=ORICOMP,xshift=0.25*\dH] {};
\node(LLAT)[fill=none,left of=LATTICE,xshift=0.25*\dH,yshift=0.25*\dV] {};
\draw[->,thick] (LIDAR.270) to[out=210,in=90] (LORI.0) -- (LLAT.0) to [out=270,in=210] node[midway,above,yshift=-0.1cm]{\tiny $\mathcal{P}_l$}
(SCENE.190);

\end{tikzpicture}}
\fi
\vspace{-0.9cm}
\caption{%
System overview: A non-uniform continuous-time spline trajectory %$T_X(t)$
defined by knots $\mathcal{X}$ describes the sensor motion. 
A new raw scan $\mathcal{P}_l$ is preoriented with \acs{IMU} before lattice embedding into a multi-resolution \acs{surfel} map $\mathcal{S}_l$.
Motion priors ($\Delta T,\Delta_{pre}$) aid the spline initialization of the sliding registration window $\mathcal{W}_l$.
Sensor motion within \acsp{surfel} is compensated prior to alignment against a keyframe-based local \acs{surfel} map $\mathcal{M}$ under motion constraints.
After spline registration, a new pointwise undistorted keyframe is added to the storage if necessary; or the local map is updated with the closest keyframes.%
}
\label{fig:lio_system}
\end{figure*}
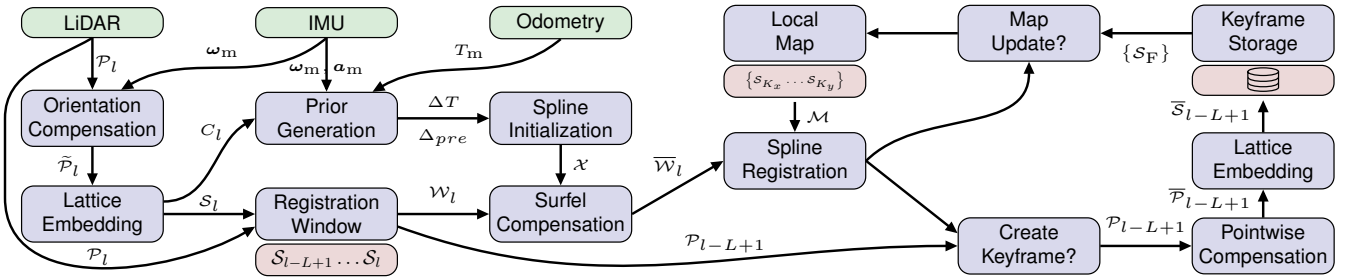

Point cloud registration is a well-researched topic and has wide applicability~\cite{HolzITRB15}.
The registration commonly involves an \ac{ICP}-variant~\cite{besl1992icp,segal2009gicp} to iteratively align a new scan with a model in two steps.
In the first step, the algorithm establishes correspondences between the both point clouds.
The second step calculates a transformation to reduce a respective distance, \eg point-to-point~\cite{besl1992icp,vizzo2022kissicp} or point-to-plane~\cite{xu2021fastlio,tuna2024xicp,lee2024genzicp} or plane-to-plane~\cite{segal2009gicp,behley2018suma,chen2021dlo}.
Commonly, both steps repeat until convergence or for a fixed number of iterations.

Since only local convergence is guaranteed, it is crucial to obtain a good initialization, \eg from preintegration of \acs{IMU} measurements~\cite{shan2020liosam,ExplorerSubT,chen2023dlio} or wheel/robot odometry~\cite{guadagnino2024kinicp}.
Despite the inclusion of priors, the registration may slip or diverge~\cite{zhang2016degeneracy}, \eg in featureless corridors~\cite{nashed2021rdslam} or in open areas~\cite{tuna2024xicp,tuna2024informed}.
KISS-\acs{ICP}~\cite{vizzo2022kissicp} robustifies point-to-point \ac{ICP} through a motion-dependent adaptive distance threshold for correspondences and optimizes with the outlier-robust Geman-McClure kernel~\cite{barron2019cvpr}.
Building upon this, GenZ-\acs{ICP}~\cite{lee2024genzicp} adaptively balances the respective influence of point-to-point constraints and additional point-to-plane factors to improve robustness.
Instead, RMS~\cite{petracek2024rms} selects informative points using a gradient flow heuristic that prefers points along edges and reduces redundant points on surfaces.
As an alternative, X-\acs{ICP}~\cite{tuna2024xicp} inspects the contribution of individual point-to-plane correspondences on the \acs{ICP}'s Hessian to filter uninformative pairs.
Constrained optimization further restricts slippage due to insufficient information, \eg in tunnels or long corridors.

While the inclusion of priors in a loosely-coupled approach~\cite{tang2015loosely,zhen2017loosely,zhao2019loosely,chen2021dlo,chen2023dlio} is easier to realize,
most \ac{LIO} systems~\cite{shan2020liosam,qin2020lins,xu2021fastlio,he2023pointlio,jung2023malio,zhang2024aslio,chen2024iglio} optimize \acs{LiDAR} jointly with \acs{IMU} at discrete timesteps (\eg end of scan) which requires temporal interpolation.
The widely adapted \ac{iekf} of Fast-\acs{LIO}~\cite{xu2021fastlio,xu2021fastlio2} propagates the state forward with \acs{IMU} measurements before back-propagating it to the respective point times to undistort the scan.
As an alternative to processing the accumulated scan, AS-LIO~\cite{zhang2024aslio} adapts the sliding window length depending on the spatial overlap between the map and the voxelized scan.
Point-\acs{LIO}~\cite{he2023pointlio} only processes points with the same timestamp at once, whereas FR-LIO~\cite{liu2024frlio} employs Kalman smoothing over the last scan.
Complementary to previous approaches, RI-\acs{LIO}~\cite{zhang2023rilio} and COIN-\acs{LIO}~\cite{pfreundschuh2023coinlio} exploit per-point reflectivity estimates from the \acs{LiDAR}.

A continuous-time trajectory representation facilitates the inclusion of multi-modal data.
\textcite{talbot2024ctreview} extensively review continuous-time state estimation, dividing methods into three groups.

The first group~\cite{lovegrove2013spline,park2017elastic,daun2021hectorgrapher} applies linear interpolation between consecutive states.
Here, a single relative transformation linearly interpolates the trajectory within the time segment, \eg using \ac{slerp}~\cite{shoemake1985slerp}, under an inherent constant-velocity assumption.
Per scan, the state typically includes the position and orientation with optional velocity and \acs{IMU} biases.
The \textit{offline} \ac{SLAM} method SLICT~\cite{nguyen2023slict} uses this trajectory representation to align a scan window using the point-to-plane error against a hierarchical \acs{surfel} map.
To prevent over-smoothing in \acs{CT-ICP}, \textcite{dellenbach2021ct} assume continuity during each scan, with residuals influencing the scan's start and end pose, but decouple consecutive scans from one another.
\acs{SE-LIO}~\cite{yuan2023selio} weakens this assumption by enforcing similarity between the end and subsequent start pose while restraining inter-scan motion via preintegrated \acs{IMU}.
On the contrary, Traj-\acs{LO}~\cite{zheng2024trajlo} adds smoothness constraints during sliding window optimization of the linear B-spline trajectory.
When sudden rotational changes deform a scan, a linear B-spline may be too inaccurate to represent the whole scan.
Hence, the adaptive temporal subdivision of \textcite{zhou2024atictlo} decreases the time interval between knots for higher angular velocities or increases interval time for underconstrained scan segments.

The second group describes the continuous-time trajectory using splines with temporal basis functions.
An example is \acs{MARS}~\cite{quenzel2021mars} which uses a flexible uniform $N$th order B-spline~\cite{sommer2020cvpr} for a sliding registration window.
The registration jointly optimizes multi-resolution \acs{surfel} maps for multiple scans without enforcing continuity outside the window.
In contrast, CLINS~\cite{lv2021clins} % https://arxiv.org/abs/2109.04687
targets \textit{offline} \ac{SLAM} and retains points on planar surfaces and edges~\cite{zhang2014loam}.
The full trajectory is represented with the same uniform B-spline as \acs{MARS} but minimizes errors on raw \acs{IMU} measurements and point-to-plane(-line) distance for undistorted planar (resp. edge) points with automatic differentiation using Ceres Solver~\cite{agarwal2022ceres}.
The follow-up work, CLIC~\cite{lv2023clic}, integrates analytical Jacobians and a camera-based frontend.
Concurrently, SLICT2~\cite{nguyen2024slict2} adopts the same trajectory as SLICT and follows the same iterated \ac{EM}-strategy as \acs{MARS} of alternating between correspondence search and spline optimization without multiple inner iterations.
Coco-\acs{LIC}~\cite{lang2023cocolic} further introduces a non-uniform cubic B-spline to adaptively select the number of knots per scan depending on the measured angular velocity and linear acceleration.
More recently, RESPLE~\cite{cao2025resple} recursively estimates the knots of a cubic B-spline with an iterated \acs{EKF} at a knot frequency of \SI{100}{\hertz}. 

The third group focuses on temporal \acp{GP} where the process model $\mathcal{GP}(\bm{\mu}(t),\Sigma(t_i,t_{i-1}))$ describes the transition between temporally adjacent states with prior mean $\bm{\mu}(t)$ and prior covariance $\Sigma(t_i,t_{i-1})$ functions.
Although \ac{GP}-based approaches may be seen as a weighted combination of infinite temporal basis functions~\cite{tong2013gp}, evaluating the trajectory at time $\tau\in[t_{i-1},t_i)$ involves only the states at $t_i$ and $t_{i-1}$ for priors based on linear, time-varying \ac{SDE}~\cite{anderson2015gpr}.
Common priors assume white noise on acceleration~\cite{talbot2024ctreview}, velocity~\cite{zheng2024trajlio}, or jerk~\cite{nguyen2024gptr}.
In contrast to spline-based methods, the state space typically contains the velocity in addition to the pose~\cite{wu2023steam,burnett2024steamlio}.
Traj-\acs{LIO}~\cite{zheng2024trajlio} replaces the linear interpolation within Traj-\acs{LO}~\cite{zheng2024trajlo} with \ac{GP} interpolation using various \ac{GP} motion priors.
Instead, \textcite{shen2024ctemlo} combine a \ac{GP} as their continuous-time representation with a \ac{KF}.

The aforementioned continuous-time methods expect regular scan input and degrade with missing or irregular intermediate scans.
In these situations, a non-uniform continuous-time B-spline adapts better while being equivalent to its uniform counterpart for regular scan input.
Hence, we perform real-time \ac{LIO} with non-uniform continuous-time B-splines with analytic Jacobians and motion compensation during optimization.
For this, we extend \acs{MARS}~\cite{quenzel2021mars} by introducing a non-uniform B-spline and tightly coupling \acs{LiDAR} with \acs{IMU}.
The analytical Jacobians for full relative motion constraints applied to the spline are derived and included to improve consistency and robustness while maintaining real-time processing.
Additionally, we introduce an \acf{UT} to enable motion compensation for individual \acsp{surfel} and employ a temporal separation into intra-scan segments to facilitate motion compensation at optimization time.
Moreover, we rephrase the \acf{GMM} and \acs{surfel} covariances of \acs{MARS} with Kronecker sum and products to improve parallelization.
The modified covariance computations are directly transferable to other LIO systems with plane-to-plane constraints (GICP).
\bgroup
\newcolumntype{C}[1]{>{\centering\arraybackslash\hspace{0pt}}m{#1}}
\newcolumntype{Z}{>{\centering\arraybackslash}X}
\newcolumntype{A}{C{1.1cm}}
\renewcommand{\arraystretch}{1.2}  %Vertical space: 1 is the default, change 
\begin{table}
\scriptsize%\tiny
\centering
\caption{Symbols}
\label{tab:lio_notation}
\setlength{\tabcolsep}{1pt}
\aboverulesep=0ex 
\belowrulesep=0ex
\begin{tabularx}{\linewidth}{AZ|AZ}
\toprule
Variable & Meaning & Variable & Meaning\\
\midrule
$\mathcal{P}$ & a point cloud & $\bm{p} \in \RD$ & a 3D point \\
$\mathcal{X}$ & the spline knots & $\mathcal{B}$ & the bias knots \\
$X_k\in \left(\SoR\right)$ & a spline knot & $B_k \in \left(\RD\!\times\!\RD\right)$ & a bias knot \\
$N$ & spline order for poses & $N_B$ & spline order for biases \\
%\midrule
$\mathcal{X}(t)$ & the spline knots with non-zero weight at time $t$ & $\kappa\left(\cdot\right)$ & condition number\\
$T_{\mathcal{X}}\left(t\right)$ & the spline pose at time $t$ & $t\left(\bm{p}\right)$ & time of the point \\
$R\left(t\right)$ & orientation at time $t$ & $\mathcal{S}^2$ & the unit sphere in $\RD$ \\
$\bm{p}\left(t\right)$ & position at time $t$ & $\bm{g}_{\mathrm{w}}$ & gravitational acceleration \\
\midrule
$\Delta t_e$ & duration of a single scan revolution & $O$ & number of segments \\
$t_l$ & end time of scan $l$ & $o$ & a segment\\
$t_{\mathrm{ref}}$ & reference time & $t_{\mathrm{seg}}\left(o\right)$ & reference time for segment $o$ \\
$\bm{\lambda}, V$ & Eigenvalues \& -vectors & $\bm{n}$ & a normal vector \\
$\bm{\mu}_s,\Sigma_s$ & a scene surfel's mean position \& covariance & $\bm{\mu}_m,\Sigma_m$ & a map surfel's mean position \& covariance\\
%\midrule
$\widetilde{\mathcal{P}}$ & a pre-oriented point cloud & $\overline{\mathcal{P}}$ & point-wise compensated point cloud\\
$\mathcal{W}_l$ & scan window at time $t_l$ & $\overline{\mathcal{W}}_l$ & compensated scan window at time $t_l$ \\
$\mathcal{S}_l$ & surfel map for scan at $t_l$ & $\overline{\mathcal{S}}_k$ & point-wise compensated keyframe surfel map\\
$\mathcal{M}$ & map surfels & $\mathcal{A}_s$ & associated surfels \\
\midrule
$\bm{d}_{f(\cdot)}$ & a residual for $f(\cdot)$ & $\mathrm{Log}\left(\cdot\right)$ & logarithm map for $\SoD$ \\
$\Jfxdx$ & (right) Jacobian for $f(X)$ \wrt $X$ & $\mathrm{Exp}\left(\cdot\right)$ & exponential map for $\SeD$ \\
$\bm{b}_{\mathrm{gyr}}$ & gyroscope bias & $\bm{b}_{\mathrm{acc}}$ & accelerometer bias \\
$\bm{a}_{\mathrm{m}}$ & measured linear acceleration & $\bm{\omega}_{\mathrm{m}}$ & measured angular velocity \\
$\overline{\bm{a}}$ & bias-corrected linear acceleration & $\overline{\bm{\omega}}$ & bias-corrected angular velocity \\
%\midrule
$\Delta R  \in \SoD$ & a relative orientation & $\Delta T \in \SeD$ & a relative pose\\
$\Delta\mathcal{R}_{\mathrm{IMU}}$ & set of relative orientation from IMU integration & $\Delta_{\mathrm{pre}}$ & preintegrated IMU pseudo-measurement \\
$\Delta\overline{R}$ & relative rotation & $\Delta h(\cdot)$ & relative position function\\
$\relRotPre$ & relative rotation difference \\
%\midrule
$\mathcal{L}_c$ & a constraint's cost function & $\mathcal{Y}_{\Sigma}$ & sigma points for \acs{UT} \\
$\mathcal{L}_{\mathrm{MARS}}$ & GMM cost function of MARS & $\bar{\mathcal{Y}}_{\Sigma}$ & sigma points after \acs{UT}\\
$\mathcal{L}_{\mathrm{marg}}$ & marginalization cost function & $\bm{a}_z$ & zero-acceleration\\
$\mathcal{L}_{\mathrm{IMU}}$ & IMU cost function & $\bm{\alpha}_z$ & zero-angular velocity \\
\midrule
$H_d$ & simplex & $c_i$ & cell side length \\
$\bm{y} \in H_d$ & closest remainder-0 point on simplex & $\gamma$ & EMWA weight for keyframe generation\\
$\bm{\nu}_k$ & a keyframe's simplex offset\\
\bottomrule
\end{tabularx}
\end{table}
\egroup

\section{Method}\label{sec:method}
We take a raw \acs{LiDAR} scan $\mathcal{P}_l$ in the sensor frame captured at time $t_l$ as input.
$\mathcal{P}_l$ consists of a set of measurements with range $r \in \mathbb{R}$ and direction $\overrightarrow{\bm{v}}\in\RD$.
This yields a point $\bm{p} = r \overrightarrow{\bm{v}}$ at time $t(\bm{p})\in(t_l-\Delta t_e,t_l]$ with scan duration $\Delta t_e$.
During one revolution, the \acs{LiDAR} measures $h$ ranges (\eg 128) simultaneously under $w$ varying directions (\eg 1024) with small fixed directional offsets $o_r$.
The scan timestamp $t_l$ corresponds to the last acquired point(s) within the \acs{LiDAR}'s revolution.
In the absence of a measured point time $t(\bm{p})$, we estimate $t(\bm{p})$ from the organized image-like structure ($h\times w$) using the column index $u_{\bm{p}}\in [0,w)$ of point $\bm{p}$. This directly relates to the time within a scan revolution as:
\begin{align}
%t(\bm{p}) &= t_{l}-\Delta t_e + \frac{\Delta t_e}{2}\left(\frac{\arctan(p_y/p_x)}{\pi} + 1 \right).\label{eq:PointTime}\\
t(\bm{p}) &= t_{l}-\Delta t_e + \frac{\Delta t_e}{w} u_{\bm{p}}.\label{eq:PointTime}
\end{align}
If the column index is unavailable, we compute $u_{\bm{p}}$ using the azimuthal angle in spherical sensor coordinates with sensor offset $o_s$:
\begin{align}
u(\bm{p},o_r) &= \frac{w}{2\pi} \arctan(p_y/p_x) + o_s + o_r.\label{eq:PointColumn}
\end{align}

During a revolution, the sensor motion continuously influences the scan origin and measurement direction.
We obtain an initial estimate of the sensor's change in orientation $\Delta R_{\mathrm{ref}}$ from integration of raw \acs{IMU} measurements of angular rate $\angRotM$ at $t_{\mathrm{m}_j}$ with the exp-map~\cite{sola2021lie,sommer2020cvpr}:
\begin{align}
\Delta R_j &= \Delta R_{j-1} \exp_{R}\left(\bm{\omega}_{\mathrm{m}_j}\right), \text{ with } \Delta R_0 = I, \\
\Delta \mathcal{R}_{\mathrm{IMU}} &= \left\lbrace \Delta R_j \lvert \forall j \text{ s.t. } t_{\mathrm{m}_j} \in(t_l-\Delta t_e,t_l]\right\rbrace.
\end{align}
The rotations $\Delta \mathcal{R}_{\mathrm{IMU}}$ enable us to
use \acs{slerp}~\cite{shoemake1985slerp} on $\Delta \mathcal{R}_{\mathrm{IMU}}$ to pre-orient $\mathcal{P}_l$ to a common reference time $t_{\mathrm{ref}}$:
\begin{align}
\widetilde{\bm{p}} &= \mathrm{slerp} (\Delta\mathcal{R}_{\mathrm{IMU}}, t_{\mathrm{ref}} )\inv \, \mathrm{slerp}(\Delta\mathcal{R}_{\mathrm{IMU}}, t_{\bm{p}}) \, \bm{p}.\label{eq:preorient}
\end{align}
Then, the oriented scan $\widetilde{\mathcal{P}}_l$ is embedded into a local multi-resolution sparse lattice $\mathcal{S}_l$ with tetrahedral cells and adaptive side length as in \cite{quenzel2021mars}.
Each cell stores a \acs{surfel} $s$ with mean $\bm{\mu}_s\in \RD$ and covariance $\Sigma_s\in\mathbb{R}^{3\times 3}$ for the embedded points.
The \acs{surfel} normal $\bm{n}_s$ corresponds to the Eigenvector $\bm{v}_0$ of the smallest Eigenvalue $\lambda_0$ of $\Sigma_s$.

We insert the \acsp{surfel} in $\mathcal{S}_l$ into the sliding registration window $\mathcal{W}_l$, as shown in \reffig{fig:lio_system}.
The window contains the last $L$ scans~(\refsec{sec:win}) with their trajectory represented by a non-uniform continuous-time B-spline $T_\mathcal{X}(t)$~(\refsec{sec:spline}).
From $\mathcal{S}_l$, the scan covariance $C_l$ (\refsec{sec:nsc}) is computed to weight a motion prior $\Delta T$ from poses of robot odometry $T_{\mathrm{m}}$. Then, the prior $\Delta T$ and the preintegrated \acs{IMU} measurement $\Delta_{pre}$ (\refsec{sec:imu}), from angular rate $\angRotM$ and linear acceleration $\linAccM$, aid the initialization of spline knots $\mathcal{X}$.
The spline allows motion compensation for \acsp{surfel} within the window $\mathcal{W}_l$ (\refsec{sec:ut}).

The registration (\refsec{sec:reg}) aligns the compensated window $\overline{\mathcal{W}}_l$ with the local \acs{surfel} map $\mathcal{M}$ by optimizing the knots $\mathcal{X}$ of the trajectory spline $T_\mathcal{X}(t)$.
The local \acs{surfel} map $\mathcal{M}$ contains \acsp{surfel} of selected spatially separated scans.
If necessary, the oldest raw scan $\mathcal{P}_{l-L+1}$ is pointwise motion compensated towards $t_{l-L+1}$ and re-embedded before its \acsp{surfel} $\overline{\mathcal{S}}_{l-L+1}$ are added as a keyframe to the keyframe storage (\refsec{sec:kf}).

\begin{figure*}
\centering
\if\WithFigures1
\resizebox{1.0\linewidth}{!}{\begin{tikzpicture}
[content_node/.append style={font=\sffamily,minimum size=1.5em,minimum width=6em,draw,align=center,rounded corners,scale=0.65},
label_node/.append style={font=\sffamily,scale=0.5},
group_node/.append style={font=\sffamily,dotted,align=center,rounded corners,inner sep=1em,thick},>={Stealth[inset=0pt,length=4pt,angle'=45]},node distance=1cm]

\pgfmathsetmacro{\dH}{1cm}
\pgfmathsetmacro{\dV}{0.75cm}

\definecolor{darkred}{rgb} {0.8,0.0,0.0}
\definecolor{red}{rgb}     {0.5,0.0,0.0}
\definecolor{green}{rgb}   {0.0,0.5,0.0}
\definecolor{darkblue}{rgb}{0.0,0.0,0.5}
\definecolor{grey}{rgb}    {0.5,0.5,0.5}
\colorlet{blue}          {blue!80!white}

\node(P0)[fill=none] at (0,0) {}; %green!15!white

% MARS
\begin{scope}
\node(P1)[fill=none,label=below:$X_0$] at (P0) {}; %green!15!white
\node(P2)[fill=none,right of=P1,xshift=0.1*\dH] {};
\node(P3)[fill=none,right of=P2,xshift=-0.2*\dH] {};
\node(P4)[fill=none,right of=P3,xshift=0.2*\dH,label=below:$X_1$] {};
\node(P5)[fill=none,right of=P4,xshift=-0.1*\dH] {};
\node(P6)[fill=none,right of=P5,xshift=-0.3*\dH] {};
\node(P7)[fill=none,right of=P6,label=below:$X_2$] {};
\node(P8)[fill=none,right of=P7,xshift=-0.2*\dH]{};

\node(Q1)[fill=none,below of=P1] {}; %green!15!white
\node(Q2)[fill=none,below of=P2,label=below:$X_0$] {};
\node(Q3)[fill=none,below of=P3] {};
\node(Q4)[fill=none,below of=P4] {};
\node(Q5)[fill=none,below of=P5,label=below:$X_1$] {};
\node(Q6)[fill=none,below of=P6] {};
\node(Q7)[fill=none,below of=P7] {};

\node(L1)[fill=none,below of=Q1,label=below:$t_{l-N}$] {};
\node(L2)[fill=none,below of=Q2,label=below:$t_{l-2}$] {};
\node(L3)[fill=none,below of=Q3,label=below:$t_{l-1}$] {};
\node(L4)[fill=none,below of=Q4,label=below:$t_{l}$] {};
\node(L5)[fill=none,below of=Q5,label=below:$t_{l+1}$] {};
\node(L6)[fill=none,below of=Q6,label=below:$t_{l+2}$] {};

\draw[|-|,thick] (P1) -- (P4.0);
\draw[-|,thick,dashed] (P4.180) -- (P7);
\draw[(-|,thick,blue] ([yshift=-0.1*\dV]P1.0) -- ([yshift=-0.1*\dV]P4.0);
\draw[|-|,thick] (Q2) -- (Q5.0);
\draw[-,thick,dashed] (Q5.180) -- (Q7);
\draw[(-|,thick,blue] ([yshift=-0.1*\dV]Q2.0) -- ([yshift=-0.1*\dV]Q5.0);

\draw[-,thick] (L1) -- (L6);
\draw[|-,thick] (L1) -- (L3);
\draw[|-,thick] (L2) -- (L4);
\draw[|-,thick] (L3) -- (L5);
\draw[|-,thick] (L4) -- (L6);
\draw[|->,thick] (L5) -- (L6.0);
\end{scope}

\node(PUO)[fill=none,right of=P0,xshift=4.6*\dH] at (P0) {}; %green!15!white
%uniform 
\begin{scope}
\node(P1)[fill=none,right of=P0,xshift=0*\dH] at (PUO) {}; %green!15!white
\node(P2)[fill=none,right of=P1,label=below:$X_{k-N}$] {};
\node(P3)[fill=none,right of=P2,label=below:$X_{k-2}$] {};
\node(P4)[fill=none,right of=P3,label=below:$X_{k-1}$] {};
\node(P5)[fill=none,right of=P4] {};
\node(P6)[fill=none,right of=P5] {};

\node(Q1)[fill=none,below of=P1] {}; %green!15!white
\node(Q2)[fill=none,below of=P2] {};
\node(Q3)[fill=none,below of=P3,label=below:$X_{k-2}$] {};
\node(Q4)[fill=none,below of=P4,label=below:$X_{k-1}$] {};
\node(Q5)[fill=none,below of=P5,label=below:$X_{k}$] {};
\node(Q6)[fill=none,below of=P6,label=below:$X_{k+1}$] {};

\node(L1)[fill=none,below of=Q1,label=below:$t_{l-N}$] {};
\node(L2)[fill=none,right of=L1,xshift=0.1*\dH,label=below:$t_{l-2}$] {};
\node(L3)[fill=none,right of=L2,xshift=-0.2*\dH,label=below:$t_{l-1}$] {};
\node(L4)[fill=none,right of=L3,xshift=0.2*\dH,label=below:$t_{l}$] {};
\node(L5)[fill=none,right of=L4,xshift=-0.1*\dH,label=below:$t_{l+1}$] {};
\node(L6)[fill=none,right of=L5,xshift=-0.3*\dH,label=below:$t_{l+2}$] {};

\draw[(-|,thick] (P2) -- (P3.0);
\draw[-|,thick,dashed] (P3.180) -- (P4.0);
\draw[-|,thick,blue] ([yshift=-0.1*\dV,xshift=-0.1*\dH]P1.0) -- ([yshift=-0.1*\dV]P3.0);

\draw[(-|,thick] (Q3) -- (Q4.0);
\draw[-|,thick] (Q4) -- (Q5.0);
\draw[-|,thick,dashed] (Q5.180) -- (Q6.0);
\draw[(-|,thick,blue] ([yshift=-0.1*\dV,xshift=0.1*\dH]Q1.0) -- ([yshift=-0.1*\dV]Q4.0);
\draw[|-|,thick,darkred] ([yshift=0.1*\dV,xshift=0.1*\dH]Q4.0) -- ([yshift=0.1*\dV]Q5.0);

% t_j to t_j+1 is within s_j-1 to s_j

\draw[-,thick] (L1) -- (L6);
\draw[|-,thick] (L1) -- (L3);
\draw[|-,thick] (L2) -- (L4);
\draw[|-,thick] (L3) -- (L5);
\draw[|-,thick] (L4) -- (L6);
\draw[|->,thick] (L5) -- (L6.0);
\end{scope}

\node(PCO)[fill=none,right of=PUO,xshift=6.0*\dH] at (PUO) {}; %green!15!white
% Coco-LIC
\begin{scope}
\node(P1)[fill=none,xshift=0*\dH] at (PCO) {};
\node(P01)[fill=none,right of=P1,xshift=-0.75*\dH] {};
\node(P02)[fill=none,right of=P01,xshift=-0.75*\dH,label=below:$X_{k-5}$]{}; %green!15!white
\node(P03)[fill=none,right of=P02,xshift=-0.75*\dH] {};
\node(P2)[fill=none,right of=P03,xshift=-0.75*\dH,label=below:${}_{\ldots}$] {};
\node(P3)[fill=none,right of=P2,label=below:$X_{k-2}$] {};
\node(P4)[fill=none,right of=P3,label=below:$X_{k-1}$] {};
\node(P5)[fill=none,right of=P4] {};
\node(P6)[fill=none,right of=P5] {};

\node(Q1)[fill=none,below of=P1] {}; %green!15!white
\node(Q01)[fill=none,below of=P01] {}; %green!15!white
\node(Q02)[fill=none,below of=P02,label=below:$X_{k-5}$] {}; %green!15!white
\node(Q03)[fill=none,below of=P03] {}; %green!15!white
\node(Q2)[fill=none,below of=P2,label=below:${}_{\ldots}$] {};
\node(Q3)[fill=none,below of=P3,label=below:$X_{k-2}$] {};
\node(Q4)[fill=none,below of=P4,label=below:$X_{k-1}$] {};
\node(Q5)[fill=none,below of=P5,label=below:$X_{k}$] {};
\node(Q6)[fill=none,below of=P6,label=below:$X_{k+1}$] {};

\node(L1)[fill=none,below of=Q1,label=below:$t_{l-N}$] {};
\node(L2)[fill=none,right of=L1,xshift=0.1*\dH,label=below:$t_{l-2}$] {};
\node(L3)[fill=none,right of=L2,xshift=-0.2*\dH,label=below:$t_{l-1}$] {};
\node(L4)[fill=none,right of=L3,xshift=0.2*\dH,label=below:$t_{l}$] {};
\node(L5)[fill=none,right of=L4,xshift=-0.1*\dH,label=below:$t_{l+1}$] {};
\node(L6)[fill=none,right of=L5,xshift=-0.3*\dH,label=below:$t_{l+2}$] {};

\draw[(-|,thick] (P2) -- (P3.0);
\draw[-|,thick,dashed] (P3.180) -- (P4.0);
\draw[|-|,thick,blue] ([yshift=-0.1*\dV,xshift=-0.1*\dH]P02.0) -- ([yshift=-0.1*\dV]P03.0);
\draw[-|,thick,blue] ([yshift=-0.1*\dV,xshift=-0.1*\dH]P03.0) -- ([yshift=-0.1*\dV]P2.0);
\draw[-|,thick,blue] ([yshift=-0.1*\dV,xshift=-0.1*\dH]P1.0) -- ([yshift=-0.1*\dV]P3.0);

\draw[(-|,thick] (Q3) -- (Q4.0);
\draw[-|,thick] (Q4) -- (Q5.0);
\draw[-|,thick,dashed] (Q5.180) -- (Q6.0);
\draw[(-|,thick,blue] ([yshift=-0.1*\dV,xshift=0.1*\dH]Q1.0) -- ([yshift=-0.1*\dV]Q4.0);

\draw[|-|,thick,blue] ([yshift=-0.1*\dV,xshift=-0.1*\dH]Q02.0) -- ([yshift=-0.1*\dV]Q03.0);
\draw[-|,thick,blue] ([yshift=-0.1*\dV,xshift=-0.1*\dH]Q03.0) -- ([yshift=-0.1*\dV]Q2.0);

\draw[|-|,thick,darkred] ([yshift=0.1*\dV,xshift=0.1*\dH]Q4.0) -- ([yshift=0.1*\dV]Q5.0);

% t_j to t_j+1 is within s_j-1 to s_j

\draw[-,thick] (L1) -- (L6);
\draw[|-,thick] (L1) -- (L3);
\draw[|-,thick] (L2) -- (L4);
\draw[|-,thick] (L3) -- (L5);
\draw[|-,thick] (L4) -- (L6);
\draw[|->,thick] (L5) -- (L6.0);
\end{scope}

\node(PNO)[fill=none,right of=PCO,xshift=5.1*\dH] at (PCO) {}; %green!15!white
% non uniform
\begin{scope}
\node(P1)[fill=none,xshift=0*\dH] at (PNO) {};
\node(P2)[fill=none,right of=P1,xshift=0.1*\dH,label=below:$X_{k-2}$] {};
\node(P3)[fill=none,right of=P2,xshift=-0.2*\dH,label=below:$X_{k-1}$] {};
\node(P4)[fill=none,right of=P3,xshift=0.2*\dH,label=below:$X_{k}$] {};
\node(P5)[fill=none,right of=P4,xshift=-0.1*\dH] {};
\node(P6)[fill=none,right of=P5,xshift=-0.3*\dH] {};

\node(Q1)[fill=none,below of=P1] {}; %green!15!white
\node(Q2)[fill=none,below of=P2] {};
\node(Q3)[fill=none,below of=P3,label=below:$X_{k-1}$] {};
\node(Q4)[fill=none,below of=P4,label=below:$X_{k}$] {};
\node(Q5)[fill=none,below of=P5,label=below:$X_{k+1}$] {};
\node(Q6)[fill=none,below of=P6] {};

\node(L1)[fill=none,below of=Q1,label=below:$t_{l-N}$] {};
\node(L2)[fill=none,below of=Q2,label=below:$t_{l-2}$] {};
\node(L3)[fill=none,below of=Q3,label=below:$t_{l-1}$] {};
\node(L4)[fill=none,below of=Q4,label=below:$t_{l}$] {};
\node(L5)[fill=none,below of=Q5,label=below:$t_{l+1}$] {};
\node(L6)[fill=none,below of=Q6,label=below:$t_{l+2}$] {};

\draw[(-|,thick] (P2) -- (P3.0);
\draw[-|,thick,dashed] (P3.180) -- (P4.0);
\draw[-|,thick,blue] ([yshift=-0.1*\dV,xshift=-0.1*\dH]P1.0) -- ([yshift=-0.1*\dV]P3.0);
\draw[(-|,thick] (Q3) -- (Q4.0);
\draw[-|,thick,dashed] (Q4.180) -- (Q5.0);
\draw[(-|,thick,blue] ([yshift=-0.1*\dV,xshift=0.1*\dH]Q1.0) -- ([yshift=-0.1*\dV]Q4.0);

% t_j to t_j+1 is within s_j-1 to s_j

\draw[-,thick] (L1) -- (L6);
\draw[|-,thick] (L1) -- (L3);
\draw[|-,thick] (L2) -- (L4);
\draw[|-,thick] (L3) -- (L5);
\draw[|-,thick] (L4) -- (L6);
\draw[|->,thick] (L5) -- (L6.0);
\end{scope}

\node(H1)[fill=none,right of=P0,xshift=5*\dH]{}; %green!15!white
\node(H2)[fill=none,below of=H1,yshift=-3*\dV]{};
\draw[-,very thick] (H1.90) -- (H2.90);

\node(H3)[fill=none,right of=H1,xshift=5.15*\dH]{}; %green!15!white
\node(H4)[fill=none,below of=H3,yshift=-3*\dV]{};
\draw[-,very thick] (H3.90) -- (H4.90);

\node(H5)[fill=none,right of=H3,xshift=5.0*\dH]{}; %green!15!white
\node(H6)[fill=none,below of=H5,yshift=-3*\dV]{};
\draw[-,very thick] (H5.90) -- (H6.90);

\node(a) [below of=P0,yshift=-2.75*\dV,xshift=2.5*\dH] {a) shifted~\cite{quenzel2021mars}};
\node(b) [below of=P0,yshift=-2.75*\dV,xshift=9.05*\dH] {b) uniform~\cite{lv2021clins,nguyen2023slict}};
\node(c) [below of=P0,yshift=-2.75*\dV,xshift=15.05*\dH] {c) non-uniform~\cite{lang2023cocolic}};
\node(c) [below of=P0,yshift=-2.75*\dV,xshift=20.8*\dH] {d) non-uniform (Ours)};
\node(t0) [below of=P0,yshift=1.25*\dV,xshift=-0.5*\dH,rotate=90] {$t$};
\node(tp1) [below of=P0,yshift=0*\dV,xshift=-0.5*\dH,rotate=90] {$t+1$};

\end{tikzpicture}
 }
\fi
\caption[Influence of knot placement]{Influence of knot placement: 
a) Knots $X_k$ influence the time (black/blue) between $t_{X_0}$ and $t_{X_1}$. 
\acs{MARS}~\cite{quenzel2021mars} necessitates reinitialization of its knots at every timestep since the uniform spline requires a constant $\Delta t$ (black) between knots while the knot times $t_{X}$ move forward with variable $\Delta t_l$.
b) Uniform knot placement with fixed $\Delta t$ , \eg as in CLINS~\cite{lv2021clins} and SLICT~\cite{nguyen2023slict}, may enforce continuity by appending new and fixing previous knots.
The difference between scan time $t_l$ and furthest knot $X_{k+1}$ [red in b)] impairs constraints on $X_{l-1}$.
c) Coco-LIC~\cite{lang2023cocolic} varies the number of knots under variable acceleration, but retains uniform knot placement under uniform motion as in b. 
d) Our non-uniform window has minimal difference and thus constrains the furthest knot better.}
\label{fig:knot_influence}
\end{figure*}
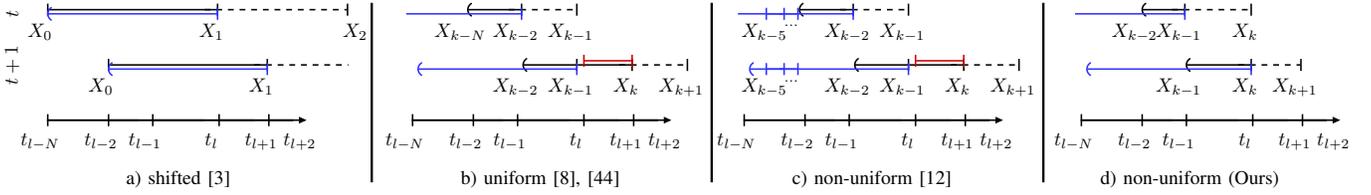

\subsection{Non-Uniform Continuous-time Trajectory}\label{sec:spline}
\textcite{sommer2020cvpr} define an $N$th order continuous-time cumulative B-spline trajectory $T_{\mathcal{X}}(t)\in \SeD$ with rotation $R(t)\in \SoD$, translation $\bm{p}(t)\in\RD$ and knots $\mathcal{X}$. 
Each knot $X_k \in \mathcal{X}$ is a tuple $\left(R_{k},\bm{p}_{k}\right)$ of the composite manifold $\SoR$ and temporally placed at time $t_k$. 
Then, $N$ temporally uniform-spaced knots $\left\lbrace X_k, X_{k+1},..., X_{k+N-1} \right\rbrace$ describe the trajectory interval $[t_k,t_{k+1})$ with duration $\Delta t>0$.

Although each control point $X_i\in\mathcal{X}$ is a tuple of $\left(R_i,\bm{p}_i\right) \in \SoR$ defined on two separate splines, the result $T_{\mathcal{X}}\left(t\right)$ is a rigid transform in $\SeD$~\cite{sommer2020cvpr}.

In the non-uniform case, $\Delta t_k = t_{k+1}-t_k > 0$ may vary, thus, leading to interval-specific B-spline basis\footnote{\textcite{sommer2020cvpr} refers to basis matrices~\cite{qin1998bm} as blending matrices. \textcite{qin1998bm} provides a recursive algorithm to compute non-uniform basis matrices for order $N$ and closed forms up to $N=4$.} matrices~\cite{qin1998bm} and requiring $2(N-1)$ timestamps:
\begin{align}
\left\lbrace t_{k-N+2},\ldots,t_k,\ldots t_{k+N-1} \right\rbrace.\label{eq:stamps}
\end{align}

We use this non-uniform continuous-time trajectory $T_{\mathcal{X}}(t)$ to represent the sensor pose $T_{\mathrm{m,s}}$ relative to the local \acs{surfel} map $\mathcal{M}$.
Hence, a point $\bm{p}_{\mathrm{s}}$ in the sensor frame corresponds to the point $\bm{p}_{\mathrm{m}} = T_{\mathrm{m,s}}\cdot \bm{p}_{\mathrm{s}}$ in the local map frame.
Initially, we set the world frame to coincide with the local map frame $T_{\mathrm{w,m}}=I$.
To maintain the locality of the local map, shifting gradually changes $T_{\mathrm{w,m}}$ by an integer multiple of the coarsest cell size while the orientation remains unchanged.

As is common in state estimation with inertial sensors~\cite{xu2021fastlio2,qin2020lins,sommer2020cvpr}, we select the \acs{IMU} as the reference sensor and transform scans prior to their lattice embedding into the reference frame with the \acs{IMU}-\acs{LiDAR} extrinsic $T_\mathrm{s,l}$. If no \acs{IMU} is available, we set $T_\mathrm{s,l}=I$.

\subsection{Sliding Registration Window}\label{sec:win}

\begin{figure}
\centering
\if\WithFigures1
\resizebox{1.0\linewidth}{!}{\begin{tikzpicture}
[content_node/.append style={font=\sffamily,minimum size=1.5em,minimum width=6em,draw,align=center,rounded corners,scale=0.65},
label_node/.append style={font=\sffamily,scale=0.5},
group_node/.append style={font=\sffamily,dotted,align=center,rounded corners,inner sep=1em,thick},>={Stealth[inset=0pt,length=4pt,angle'=45]},node distance=1.5cm and 1.0cm]

\pgfmathsetmacro{\dH}{1.0cm}
\pgfmathsetmacro{\dV}{0.75cm}

\definecolor{darkred}{rgb} {0.8,0.0,0.0}
\definecolor{red}{rgb}     {0.5,0.0,0.0}
\definecolor{green}{rgb}   {0.0,0.5,0.0}
\definecolor{blue}{rgb}    {0.0,0.0,0.5}
\definecolor{grey}{rgb}    {0.5,0.5,0.5}

\node(P0)[fill=none] at (0,0){};
\begin{scope}
\node(P1)[fill=none] at (P0) {};
\node(P2)[fill=none,right of=P1,yshift=-0.4cm] {};
\node(P3)[fill=none,right of=P2,yshift=0.1cm] {};
\node(P4)[fill=none,right of=P3,yshift=-0.2cm] {};
\node(P5)[fill=none,right of=P4,xshift=-0.5*\dH,yshift=0.1cm] {};
\node(P6)[fill=none,right of=P5,xshift=-0.5*\dH,yshift=0.2cm] {};

\node(Q1)[fill=blue!15!white,below of=P1,yshift=\dV] {};
\node(Q2)[fill=blue!15!white,below of=P2,yshift=\dV] {};
\node(Q3)[fill=blue!15!white,below of=P3,yshift=\dV] {};
\node(Q4)[fill=blue!15!white,below of=P4,yshift=\dV] {};
\node(Q5)[fill=blue!15!white,below of=P5,yshift=\dV] {};
\node(Q6)[fill=none,below of=P6,yshift=\dV] {};

\node(R1)[fill=none,below of=Q1,yshift=\dV] {};
\node(R2)[fill=none,below of=Q2,yshift=\dV] {};
\node(R3)[fill=none,below of=Q3,yshift=\dV] {};
\node(R4)[fill=none,below of=Q4,yshift=\dV] {};
\node(R5)[fill=none,below of=Q5,yshift=\dV] {};

\node(K1)[fill=none,below of=R1,yshift=-0.25*\dV] {};
\node(K2)[fill=none,right of=K1,yshift=-0.4cm] {};
\node(K3)[fill=none,right of=K2,yshift=0.1cm] {};
\node(K4)[fill=none,right of=K3,yshift=-0.2cm] {};
\node(K5)[fill=none,right of=K4,xshift=-0.5*\dH,yshift=0.1cm] {};
\node(K6)[draw=none,fill=none,right of=K5,xshift=-0.5*\dH,yshift=0.2cm] {};

\node(L1)[fill=blue!15!white,below of=K1,yshift=\dV] {};
\node(L2)[fill=blue!15!white,below of=K2,yshift=\dV] {};
\node(L3)[fill=blue!15!white,below of=K3,yshift=\dV] {};
\node(L4)[fill=blue!15!white,below of=K4,yshift=\dV] {};
\node(L5)[fill=blue!15!white,below of=K5,yshift=\dV] {};
\node(L6)[fill=none,below of=K6,yshift=\dV] {};

\node(M1)[fill=none,below of=L1,yshift=\dV] {};
\node(M2)[fill=none,below of=L2,yshift=\dV] {};
\node(M3)[fill=none,below of=L3,yshift=\dV] {};
\node(M4)[fill=none,below of=L4,yshift=\dV] {};
\node(M5)[fill=none,below of=L5,yshift=\dV] {};
\node(M6)[fill=none,below of=L6,yshift=\dV] {};

\node(A0)[below of=R1,yshift=-0.75*\dV]{};
\node(A1)[right of=A0,xshift=3*\dH]{};
\node(A2)[above of=A0]{};
\node(A3)[right of=A0,yshift=0.5*\dV]{};
\node(A4)[above of=A3,yshift=-0.8*\dV]{};
	
\draw[draw=none,very thick,darkred,pattern=north east lines,pattern color=darkred] (A2.180) rectangle (A4);
\draw[draw=none,fill=green] (P2 |- A2) circle (0.15em);
\draw[draw=none,fill=green] (P3 |- A2) circle (0.15em);
\draw[draw=none,fill=green] (P4 |- A2) circle (0.15em);

\draw[-,very thick,decorate,decoration={brace, mirror,raise=0.15cm}] (P2 |- A2) -- node[below,midway,yshift=-0.2cm]{opt. scans}(P4 |- A2);
\draw[-,very thick,decorate,decoration={brace, mirror,raise=0.65cm}] (A2.180) -- node[below,midway,yshift=-0.6cm]{spline $t_i$}(P4 |- A2);

\node(PiN) [below of=Q1,yshift=1.5*\dV]{$P_{i-L}$};
\node(Pim2)[below of=Q2,yshift=1.5*\dV]{$P_{i-2}$};
\node(Pim1)[below of=Q3,yshift=1.5*\dV]{$P_{i-1}$};
\node(Pip0)[below of=Q4,yshift=1.5*\dV]{$P_{i}$};
\node(Pip1)[below of=Q5,yshift=1.5*\dV]{$P_{i+1}$};

\begin{scope}
\draw [clip,draw=none] (P2) rectangle (R4);
\draw [very thick] plot [smooth] coordinates {(P1)(Q2)(Q3)(Q4)(R5)};
\end{scope}
\begin{scope}
\draw [clip,draw=none] (P1) rectangle (R2);
\draw [dashed,very thick,darkred] plot [smooth,tension=1] coordinates {(P1)(Q2)(Q3)(Q4)(R5)};
\end{scope}
\begin{scope}
\draw [clip,draw=none] (P4) rectangle (R5);
\draw [dashed,darkred] plot [smooth,tension=1] coordinates {(P1)(Q2)(Q3)(Q4)(R5)};
\end{scope}

\node(A0)[below of=L1,yshift=-0.5*\dV,xshift=0.25cm]{};
\node(A1)[right of=A0,xshift=3*\dH]{};
\node(A2)[above of=A0]{};
\node(A3)[right of=A0,yshift=0.5*\dV]{};
\node(A4)[above of=A3,yshift=-0.8*\dV]{};

\draw[-,very thick,decorate,decoration={brace, mirror,raise=0.65cm}] (A2.180) -- node[below,midway,yshift=-0.6cm]{spline $t_i$ = opt. scans}(P4 |- A2);

\begin{scope}
\draw [clip,draw=none] (K1) rectangle (P4);
\draw [very thick] plot [smooth] coordinates {(Q1)(Q2)(Q3)(Q4)(Q5)(Q6)};
\end{scope}

\begin{scope}
\draw [clip,draw=none] (K1) rectangle (M4);
\draw [very thick] plot [smooth] coordinates {(L1)(L2)(L3)(L4)(L5)(L6)};
\end{scope}

\end{scope}

\begin{scope}
\node(P1)[fill=none,right of=P0,xshift=5*\dH]{}; %green!15!white
\node(P2)[fill=none,right of=P1,yshift=-0.4cm] {};
\node(P3)[fill=none,right of=P2,yshift=0.1cm] {};
\node(P4)[fill=none,right of=P3,yshift=-0.2cm] {};
\node(P5)[fill=none,right of=P4,xshift=-0.5*\dH,yshift=0.1cm] {};
\node(P6)[fill=none,right of=P5,xshift=-0.5*\dH,yshift=0.2cm] {};

\node(Q1)[fill=blue!15!white,below of=P1,yshift=\dV] {};
\node(Q2)[fill=blue!15!white,below of=P2,yshift=\dV] {};
\node(Q3)[fill=blue!15!white,below of=P3,yshift=\dV] {};
\node(Q4)[fill=blue!15!white,below of=P4,yshift=\dV] {};
\node(Q5)[fill=blue!15!white,below of=P5,yshift=\dV] {};
\node(Q6)[fill=none,below of=P6,yshift=\dV] {};

\node(R1)[fill=none,below of=Q1,yshift=\dV] {};
\node(R2)[fill=none,below of=Q2,yshift=\dV] {};
\node(R3)[fill=none,below of=Q3,yshift=\dV] {};
\node(R4)[fill=none,below of=Q4,yshift=\dV] {};
\node(R5)[fill=none,below of=Q5,yshift=\dV] {};
\node(R6)[fill=none,below of=Q6,yshift=\dV] {};

\node(K1)[fill=none,below of=R1,yshift=-0.25*\dV] {};
\node(K2)[fill=none,right of=K1,yshift=-0.4cm] {};
\node(K3)[fill=none,right of=K2,yshift=0.1cm] {};
\node(K4)[fill=none,right of=K3,yshift=-0.2cm] {};
\node(K5)[fill=none,right of=K4,xshift=-0.5*\dH,yshift=0.1cm] {};
\node(K6)[draw=none,fill=none,right of=K5,xshift=-0.5*\dH,yshift=0.2cm] {};

\node(L1)[fill=blue!15!white,below of=K1,yshift=\dV] {};
\node(L2)[fill=blue!15!white,below of=K2,yshift=\dV] {};
\node(L3)[fill=blue!15!white,below of=K3,yshift=\dV] {};
\node(L4)[fill=blue!15!white,below of=K4,yshift=\dV] {};
\node(L5)[fill=blue!15!white,below of=K5,yshift=\dV] {};
\node(L6)[fill=none,below of=K6,yshift=\dV] {};

\node(M1)[fill=none,below of=L1,yshift=\dV] {};
\node(M2)[fill=none,below of=L2,yshift=\dV] {};
\node(M3)[fill=none,below of=L3,yshift=\dV] {};
\node(M4)[fill=none,below of=L4,yshift=\dV] {};
\node(M5)[fill=none,below of=L5,yshift=\dV] {};

\node(A0)[below of=Q1,yshift=-0.5*\dV,xshift=0.25cm]{};
\node(A1)[right of=A0,xshift=3*\dH]{};
\node(A2)[above of=A0]{};
\node(A3)[right of=A0,yshift=0.5*\dV]{};
\node(A4)[above of=A3,yshift=-0.8*\dV]{};

\draw [blue,very thick] plot [smooth] coordinates {(Q1)(Q2)(Q3)(Q4)(Q5)};
\draw [blue,very thick] plot [smooth] coordinates {(L1)(L2)(L3)(L4)(L5)};

\node(B2)[below of=M1,yshift=2*\dV]{};
\node(B5)[right of=B2,xshift=3*\dV]{};
\draw[-,very thick,decorate,decoration={brace, mirror,raise=0.5cm}] (B2.180) -- node[below,midway,yshift=-0.55cm]{spline $t_{i+1}$}(M5 |- B5);
\draw[-,very thick,green,decorate,decoration={brace, mirror}] (L1 |- B2) -- node[below,midway,yshift=-0.05cm]{fixed}(L2 |- B2);
\draw[-,very thick,decorate,decoration={brace, mirror}] (L2 |- B2) -- node[below,midway,yshift=-0.05cm]{opt. scans}(L5 |- B2);

\node(A55)[fill=none,draw=none,right of=P5,xshift=-0.75*\dH]{}; 
\begin{scope}
\draw [clip,draw=none] (K2) rectangle (M5);
\draw [very thick] plot [smooth] coordinates {(L1)(L2)(L3)(L4)(L5)(L6)};
\end{scope}
\begin{scope}
\draw [clip,draw=none] (L1) rectangle (M2);
\draw [very thick,green] plot [smooth] coordinates {(L1)(L2)(L3)(L4)(L5)};
\end{scope}
\begin{scope}
\draw [clip,draw=none] (L5) rectangle (A55|-K6);
\draw [dashed,darkred] plot [smooth,tension=1] coordinates {(L1)(L2)(L3)(L4)(L5)(L6)};
\end{scope}

% MARS:
\begin{scope}
\draw [clip,draw=none] (P3) rectangle (R5);
\draw [very thick] plot [smooth] coordinates {(R2)(Q3)(Q4)(Q5)(P6)};
\end{scope}
\begin{scope}
\draw [clip,draw=none] (R2) rectangle (Q3);
\draw [dashed,very thick,darkred] plot [smooth,tension=1] coordinates {(R2)(Q3)(Q4)(Q5)(P6)};
\end{scope}
\begin{scope}
\draw [clip,draw=none] (P5) rectangle (A55|-R6);
\draw [dashed,darkred] plot [smooth,tension=1] coordinates {(R2)(Q3)(Q4)(Q5)(P6)};
\end{scope}

\node(B0)[below of=R1,yshift=-0.75*\dV]{};
\node(B1)[right of=B0,xshift=4*\dH]{};
\node(B2)[above of=B0] at ( R2 |- B0){};
\node(B3)[right of=B0,yshift=0.5*\dV]{};
\node(B4)[yshift=-0.3*\dV] at (R3 |- B2){};

\draw[draw=none,very thick,darkred,pattern=north east lines,pattern color=darkred] (B2.180) rectangle (B4);
\draw[draw=none,fill=green] (R3 |- B2) circle (0.15em);
\draw[draw=none,fill=green] (R4 |- B2) circle (0.15em);
\draw[draw=none,fill=green] (R5 |- B2) circle (0.15em);
\draw[-,very thick,decorate,decoration={brace, mirror,raise=0.15cm}] (R3 |- B2) -- node[below,midway,yshift=-0.2cm]{opt. scans}(R5 |- B2);
\draw[-,very thick,decorate,decoration={brace, mirror,raise=0.65cm}] (B2.180) -- node[below,midway,yshift=-0.6cm]{spline $t_i$}(R5 |- B2);

\end{scope}

\node(a) [below of=P0,yshift=-5.5*\dV,xshift=0.25*\dH] {a) $t_i$};
\node(b) [below of=P0,yshift=-5.5*\dV,xshift=7*\dH] {b) $t_{i+1}$};

\node(H1)[fill=none,right of=P0,xshift=4.5*\dH]{};
\node(H2)[fill=none,below of=H1,yshift=-5.5*\dV]{};
\draw[-,very thick] (H1.270) -- (H2.90);

\node(a) [below of=P0,yshift=-0.5*\dV,xshift=-0.5*\dH,rotate=90] {MARS};
\node(b) [below of=P0,yshift=-4*\dV,xshift=-0.5*\dH,rotate=90] {Ours};

\end{tikzpicture}}
\fi
\vspace{-2mm}
\caption[Spline window comparison]{Spline window: a) \acs{MARS}~\cite{quenzel2021mars} only optimizes a $N$th order spline for $L$ scans (\eg$N=L=3$) for a single interval $\Delta t$ from scan $j-L$ until $j$ allowing discontinuity between $j-L$ and $j-L+1$ (left, red dashed). b) Our method enforces continuity and keeps previous intervals fixed while optimizing one interval per scan.}
\label{fig:timing_knots}
\end{figure}
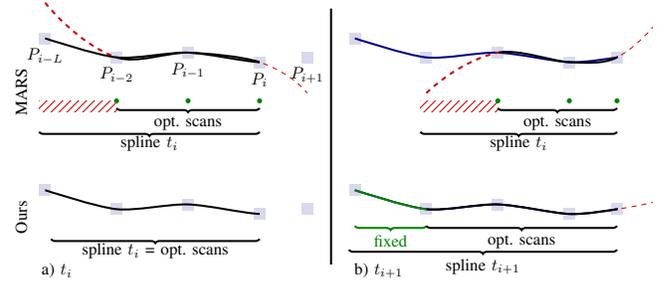

\acs{MARS}~\cite{quenzel2021mars} optimizes the trajectory $T_\mathrm{w,l}(t)$ of $L$ scans within the interval $[t_{l-L},t_{l}]$ using \acs{LM}~\cite{Transtrum2012LM} and requires reinitialization as $\Delta t$ change (\reffig{fig:knot_influence}) and $t_k$ (\reffig{fig:timing_knots}).
Instead, CLINS~\cite{lv2021clins} and SLICT2~\cite{nguyen2023slict} fixate older knots not contributing to the current scan resp. window, and add new knots uniformly such that $t_l \in [t_k,t_{k+1})$.
Although Coco-\acs{LIC}~\cite{lang2023cocolic} uses a non-uniform B-spline, it retains the notion of a fixed time interval $\Delta t$, \eg \SI{0.1}{\second}, and subdivides each interval $[t_{\kappa},t_{\kappa+1})$ into a variable number of uniformly placed knots depending on \acs{IMU} excitation.
Below a certain \acs{IMU} threshold, the subdivision leads to a uniform B-spline, as with CLINS.
Moreover, their \acs{IMU}-based selection strategy delays the optimization by up to $(N-1) \Delta t$ time intervals, \eg \SI{0.2}{\second} for low \acs{IMU} excitation.
This is a direct consequence of the basis matrix computation\footnote{Although not explicitly stated in \cite{lang2023cocolic}, the basis matrix for $[t_{\kappa},t_{\kappa+1})$ requires the knot time $t_{k+N-1}$ where $t_{k+N-1} > t_{\kappa+1}$.}, since \acs{IMU} measurements within $[t_{\kappa+N-1},t_{\kappa+N})$ influence the knot times \refeq{eq:stamps} required for $[t_{\kappa},t_{\kappa+1})$.

\begin{figure}
\centering
\if\WithFigures1
\resizebox{1.0\linewidth}{!}{\begin{tikzpicture}
\begin{semilogyaxis}
[name=plot1,grid=both,height=5cm,
 ylabel={$\kappa_l / \kappa_r$},
 xlabel={$t_l$},
 legend style={cells={align=left}},
 legend cell align=left,
 legend pos=outer north east,
 xtick={0,0.5,1},
 xticklabels={$t_{k}$,,$t_{k+1}$},
 every axis plot/.append style={very thick}]
\addplot [color=blue] table[x expr=\thisrowno{0},y expr=\thisrowno{1}] {data_last_knot_dist.txt}; 
\addlegendentry{uniform\\($t_l \not\approx t_{k+1}$)}%
\addplot [color=black] coordinates{ (0,1) (1,1) };
\addlegendentry{non-uniform\\($t_l \approx t_{k+1}$)}
\end{semilogyaxis}

\begin{axis}
[name=plot2,at={(plot1.right of north east)}, anchor=north west, xshift=1.5cm, grid=both,height=5cm,
 ylabel={weight $\lambda$},
 xlabel={$t$},
 legend cell align=left,
 legend pos=outer north east,
 xtick={-2,-1,0,1},
 xticklabels={$t_{k-2}$,$t_{k-1}$,$t_{k}$,$t_{k+1}$},
 every axis plot/.append style={very thick},xmin=-2]
\addplot [color=black,dashed] table[x expr=\thisrowno{0}-5,y expr=\thisrowno{7}] {data_knot_weight.txt};
\addlegendentry{$k-3$}
\addplot [color=black] table[x expr=\thisrowno{0}-5,y expr=\thisrowno{8}] {data_knot_weight.txt};
\addlegendentry{$k-2$}
\addplot [color=green] table[x expr=\thisrowno{0}-5,y expr=\thisrowno{9}] {data_knot_weight.txt};
\addlegendentry{$k-1$}
\addplot [color=orange] table[x expr=\thisrowno{0}-5,y expr=\thisrowno{10}] {data_knot_weight.txt};
\addlegendentry{$k$}
\addplot [color=red] table[x expr=\thisrowno{0}-5,y expr=\thisrowno{11}] {data_knot_weight.txt};
\addlegendentry{$k+1$}

\end{axis}

\node[scale=1.25, anchor=north east,xshift=-0.45cm,yshift=-0.31cm, rectangle, fill=white, align=center, font=\small\sffamily] (n_0) at (plot1.south west) {a)};

\node[scale=1.25, anchor=north east,xshift=-0.45cm,yshift=-0.31cm, rectangle, fill=white, align=center, font=\small\sffamily] (n_0) at (plot2.south west) {b)};

\end{tikzpicture} 
 }
\fi
\vspace{-2mm}
\caption[Ill-conditioning of uniform B-spline]{Ill-conditioning of uniform B-spline \wrt the last constraint.
The condition number $\kappa_l$ improves for $t_l$ approaching $t_{k+1}$ [a)] since the weight $\lambda$ [b)] increases for the newest knot ($k+1$).
}
\label{fig:last_knot_dist}
\end{figure}
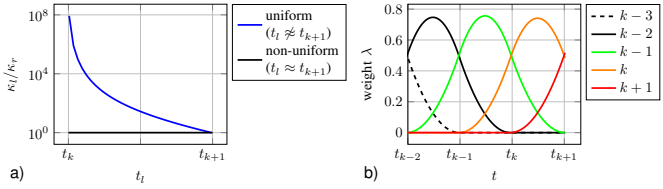

The normal equations used within \acs{LM} easily become ill-conditioned\footnote{An ill-conditioned linear system of equations is sensitive to small perturbations in the input data or due to limited numerical precision.
The computed solution may differ strongly from the exact result.
The condition number $\kappa(A) = \frac{\abs{\lambda_{\max}(A)}}{\abs{\lambda_{\min}(A)}}$ for a \ac{spd} matrix $A$ quantifies this sensitivity~\cite{higham2002numerical}.} if the current scan\footnote{On the Newer College ``Cloister'' sequence~\cite{zhang2021newerext}, around half of all scans have a $\Delta t$ slightly larger than the expected $\Delta t_e=\SI{0.1}{\second}$.} ends very close to the previous knot $(t_l-t_{k} \ll t_{k+1}-t_l)$.
We show this in \reffig{fig:last_knot_dist} by analysis of the condition number $\kappa(\mathrm{J}\mathrm{W}\mathrm{J}^\intercal)$.
For a spline with $N=3$ and a window of $L=3$ scans, we evaluate over \num{100} uniform time steps in $[t_{l-L},t_{l}]$ under the assumption of uniform data constraints (\eg $W=I_6$).
The ill-conditioning becomes apparent for $\kappa_l$ at $t_l$ relative to the reference $\kappa_r$ with $t_l \approx t_{k+1}$.
The reason behind this is the small B-spline knot weight at the end (resp. beginning) of the knot's local support.
We found that the optimization may approach local minima where the end knots are far away from reasonable values.

Without further constraints, this ill-conditioning leads to unrealistic motion, \eg with high acceleration.
We investigate the relevance of this situation by computing the difference between knot and scan time $(t_{k+1}-t_l)$ for CLINS\footnote{CLINS uses a $\Delta t=\SI{50}{\milli\second}$ for more dynamic datasets, such as hand-held ones, to create two knots per scan.} on the Newer College dataset~\cite{zhang2021newerext}.
Ideally, the difference is close to zero.
Our test showed that \SI{11.4}{\percent} of those time differences are above \SI{40}{\milli\second} (or \SI{80}{\percent} of $\Delta t$) on the ``Quad-Hard'' sequence.
The average is around \SI{10.8}{\milli\second} or \SI{21.7}{\percent} of $\Delta t$.
The situation is worse for the ``Cloister'' sequence, where the average is around \SI{21.6}{\milli\second} or \SI{43.18}{\percent} of $\Delta t$. Here, \SI{42}{\percent} of all time differences are above \SI{40}{\milli\second} (or \SI{80}{\percent} of $\Delta t$).
Accordingly, such ill-conditioning is a common occurence and not the exception.

We combine the approaches of \acs{MARS} and CLINS more robustly.
Our non-uniform spline sets the knot time such that $t_{k+1} = t_l + \epsilon$ with $\epsilon = \SI{1}{\ns}$ and thus $\kappa_l \approxeq \kappa_r$ prevents the above ill-conditioning. %$[t_{l-1},t_l) = (t_{k-1},t_k]$
Furthermore, we optimize the $L$ knots influencing the newest $L$ intervals such that our window is $\mathcal{W}_l = \left\lbrace S_{l-L+1}, \ldots S_l \right\rbrace$.
Each interval spans the time between two consecutive scans.
Knots $X_{k+2},\ldots,X_{k+N-2}$ are set with the expected constant $\Delta t_e$ and are temporally corrected once a new scan arrives.
Hence, we do not introduce a delay as in Coco-\acs{LIC}~\cite{lang2023cocolic}.

\subsection{Registration}\label{sec:reg}
We use the registration of \acs{MARS}~\cite{quenzel2021mars} including its adaptive \acs{surfel} resolution selection.
A normal distribution $\bm{e}_{sm}\sim \mathcal{N}\left(\bm{d}_{sm},\Sigma_{sm}+\sigma^2_l I\right)$ models the observation likelihood of scene \acs{surfel} $s\in\mathcal{S}_l$ for map \acs{surfel} $m\in\mathcal{M}$  with $\bm{d}_{sm}$:
\begin{align}
\bm{d}_{sm} &= T(t) \bm{\mu}_s - \bm{\mu}_m,\quad
\Sigma_{sm} = \Sigma_m + R(t)\Sigma_sR(t)^\intercal.\label{eq:rot_sigma}
\end{align}
We use the covariance as is for \textit{non-planar} \acsp{surfel}. For \textit{planar} \acsp{surfel}, we scale the covariance $\Sigma$ during registration based on their Eigendecomposition:
\begin{align}
\Sigma &= V D V^\intercal,\label{eq:eig_decomp}\quad
V=[\bm{v}_0,\bm{v}_1,\bm{v}_2],\quad
%D=\diag\left(\left[\lambda_i,\lambda_1,\lambda_2\right]^\intercal\right),
D=\diag\left(\bm{\lambda}\right),
\end{align} with sorted Eigenvalues $\lambda_0 \leq \lambda_1 \leq \lambda_2 \in \mathbb{R}$ and corresponding Eigenvectors $\veli \in \mathbb{R}^3$.
Replacing $D$ with a scaled matrix $\widetilde{D}$ based on the cell size $c_l$ on the \acsp{surfel}' level:
\begin{align}
\widetilde{\bm{\lambda}} &= \left[\min(\lambda_0,0.001), \frac{c_l}{2}, \frac{c_l}{2}\right]^\intercal,\quad
\widetilde{D} =\diag\left(\widetilde{\bm{\lambda}}\right),\label{eq:planar_cov}
\end{align}
reinforces the constraint in the normal direction while giving more leeway in the other directions. 

\acs{MARS} uses a \ac{GMM} to represent a scene \acs{surfel} $s\in\mathcal{S}_l$ observing multiple map \acsp{surfel} $A_s \in \mathcal{M}$ with $\bm{e}_{sm}$ while taking their similarity into account with weight $w_{sm}$ and jointly minimizes the data term $\mathcal{L}_{\mathrm{MARS}}$ for $L$ scans:
\begin{align}
\mathcal{L}_{\mathrm{MARS}}(l) &= \sum_{s\in{S_l}} \sum_{m\in{A_s}} w_{sm} \bm{d}_{sm}^\intercal \Sigma_{sm}\inv \bm{d}_{sm},\label{eq:mahalanobis}
\end{align}
with \ac{LM}~\cite{Transtrum2012LM} to update the knots $\mathcal{X}$.

We introduce another prior probability $p(\theta)$ into the \ac{GMM}'s similarity $p(\delta_{sm})$ based on the dot product between \acs{surfel} normal $\bm{n}$ and mean view direction $\bm{f}$ from sensor origin $\bm{o}$:
\begin{align}
%p(\theta) &\sim \mathcal{N}\left(\arccos\left(\bm{n}^\intercal\frac{\left(\bm{\mu}-\bm{o}\right)}{\norm{\left(\bm{\mu}-\bm{o}\right)}}\right), \left(\pi/8\right)^2\right).\label{eq:steep}
p(\theta) &\sim \mathcal{N}\left(\arccos\left(\bm{n}^\intercal\bm{f}\right), \left(\pi/8\right)^2\right).\label{eq:steep}
\end{align} 
This prior reduces the influence of surfaces measured under a steep angle whose normals are less reliable, \eg when measuring the floor or road far ahead.

To distribute constraints temporally more evenly and to follow the actual sensor trajectory more closely, we linearly subdivide each scan into $O$ time segments with $o\in [0,O-1]$:
\begin{align}
t_{\mathrm{seg}}(o) &= t_{l-1} + \frac{o}{O-1} \left(t_l - t_{l-1}\right) \in (t_{l-1},t_l].\label{eq:tm}
\end{align}
It is a reasonable assumption for a spinning \acs{LiDAR} that points fused within the same \acs{surfel} are also close in time.
Hence, we group \acsp{surfel} to segments \wrt their mean time.
Then, we register each segment, evaluated at its respective time $t_{\mathrm{seg}}(o)$, using the data term $\mathcal{L}_\mathrm{MARS}$.
This acts as a segment-wise undistortion during optimization.

\subsection{Optimization}
Complementary to \refeq{eq:mahalanobis}, we include constraints $\mathcal{L}_c$ and a marginalization prior $\mathcal{L}_\mathrm{marg}$ in the optimization.
The constraints $\mathcal{L}_{c}$ comprise weighted terms for \acs{IMU} $\mathcal{L}_\mathrm{IMU}$~\refeq{eq:imu}, zero-acceleration $\mathcal{L}_\mathrm{z}$~\refeq{eq:zero_acc}, and relative poses $\mathcal{L}_{\dDT}$~\refeq{eq:rel_pose}:
\begin{align}
\mathcal{L}_c &= w_\mathrm{IMU} \mathcal{L}_\mathrm{IMU} + w_\mathrm{z} \mathcal{L}_\mathrm{z} + w_\mathrm{\dDT} \mathcal{L}_{\dDT}.
\end{align}
We discuss these terms in greater detail in \refsec{sec:imu} and \refsec{sec:nsc}.

An \acs{IMU} provides angular velocity $\bm{\omega}_\mathrm{m}$ and linear acceleration $\bm{a}_\mathrm{m}$ readings of the gyroscope and accelerometer, respectively.
Fortunately, the spline allows direct evaluation of $\bm{\omega}(t)$ for $N\geq 2$ , resp. $\bm{a}(t)$ for $N\geq 3$,
with zero-mean additive biases $\bm{b}_\mathrm{gyr}(t)$, $\bm{b}_\mathrm{acc}(t)$, and gravity $\bm{g}_\mathrm{w}$:
\begin{align}
\bm{\omega}_\mathrm{m} &= \bm{\omega}(t) + \bm{b}_\mathrm{gyr}(t),\label{eq:angvel}\\
\bm{a}_\mathrm{m} &= R(t)^\intercal \left( \bm{a}(t) + \bm{g}_\mathrm{w}\right) + \bm{b}_\mathrm{acc}(t).\label{eq:accel}
\end{align}

We extend our state $\bm{\mathrm{x}}$ with time-varying accelerometer and gyroscope biases $\bm{b}_{\mathrm{acc}}(t),\bm{b}_{\mathrm{gyr}}(t)\in\RD$. 
Hence, two non-uniform B-splines represent the biases with a reduced order $N_{B}\in\left\lbrace 1,2\right\rbrace$ and bias knots $B_k=(\RD \times \RD)\in\mathcal{B}$.

Accelerometer biases are quite small ($<\abs{\num{0.1}}\SI{}{\metre\per\square\second}$) compared to the earth's gravity $\bm{g}_{\mathrm{w}} \approx \SI{9.81}{\metre\per\square\second}$.
Hence, gravity commonly dominates $\bm{a}(t)$ in \refeq{eq:accel} in our robotic applications ($\leq \abs{\num{2.5}}\SI{}{\metre\per\square\second}$ in \cite{quenzel2021mars}). 
Yet, $\bm{g}_{\mathrm{w}}$ acts in a specific direction, which makes its compensation straightforward based on the orientation $R(t)$.
Conversely, imperfect initialization of the orientation impairs the trajectory and map accuracy. 
Thus, we model the gravity direction
$\bm{g}$ on the unit 2-sphere $\mathcal{S}^2 = \left\lbrace \bm{x} \in \RD: \norm{\bm{x}}=1\right\rbrace$~\cite{xu2021fastlio2}. 
Then the gravity vector becomes $\bm{g}_{\mathrm{w}} = 9.81 \cdot \bm{g}$ with $\bm{g} \in \mathcal{S}^2$.
Initially, we set $\bm{g} = [0,0,-1]^\intercal$ and $R(t)$ according to a 3-axis magnetometer.
In the absence of the magnetometer, acceleration measurements of a stationary sensor allow the observation of roll and pitch of $R(t)$ while the yaw remains unobservable~\cite{geneva2020openvins} and can be chosen arbitrarily.
In practice, averaging over a short period in the beginning is sufficient~\cite{chen2021dlo,chen2023dlio,xu2021fastlio2}.

Overall, we optimize the following cost function using \ac{LM}:
\begin{align}
\argmin_{\mathcal{X},\mathcal{B},\bm{g}_{\mathrm{w}}} \sum_{l=0}^{L-1} \left(\sum_{o=0}^{O-1} \mathcal{L}_{\mathrm{MARS}}(l_o)\right) + \mathcal{L}_{c}(l)+ \mathcal{L}_\mathrm{marg}.\label{eq:lio_opt}
\end{align}

\subsection{Marginalization}
Following the standard procedure in \acs{LIO}~\cite{xu2021fastlio2,shan2020liosam} and \acs{VIO}~\cite{engel2016dso,usenko2020basalt,geneva2020openvins,demmel2021rootvo}, we marginalize old state variables leaving the sliding optimization window. These are the oldest knots for pose $X_{l-N}$ and bias $B_{l-N_B}$ within the window $\mathcal{W}_l$.
Entries for new variables are initialized to zero in the next iteration.
For our initialization, we perform a separate marginalization of all bias knots $\mathcal{B}$ and the gravity $\bm{g}_{\mathrm{w}}$ except for the remaining poses.

\subsection{Initialization}
In contrast to \acs{MARS}~\cite{quenzel2021mars}, we use associated \acsp{surfel} directly for the older scans.
For a new scan, we incorporate its constraints $\mathcal{L}_c$ and none of its \acsp{surfel}.
The linearization point is fixed for each knot---except for the new ones.
Thus, we employ First-Estimate Jacobians~\cite{huang2008fej} and fix \acs{surfel} associations beforehand. When all previous knots fit well from earlier registrations, any modification likely increases their respective errors.
With all knots but one quasi-fixed, the spline evaluations become linear interpolations for timestep $t$. 
As a result, only the newest knot may change to minimize the constraints' errors.
In some cases, we found oscillations occurring at twice the frequency of placed knots.
Hence, we increase the weight of the \acs{IMU} term by a factor of \num{10} for the newest scan to prevent early local minima and optimize:
\begin{align}
%X^{\mathrm{init}} &= 
\argmin_\mathcal{X} \sum_{l=0}^{L-1}\mathcal{L}_{c}(l)+ \sum_{l=0}^{L-2} \left(\sum_{o=0}^{O-1} \mathcal{L}_{\mathrm{MARS}}(l_o)\right) + \mathcal{L}_\mathrm{marg}.
\end{align}
For biases, new knots are set to the previous estimate.

\subsection{Inertial Spline Constraints}\label{sec:imu}
Constraints on the angular velocity $\bm{\omega}(t)$ and linear acceleration $\bm{a}(t)$ arise naturally from the gyroscope \refeq{eq:angvel} and accelerometer measurements \refeq{eq:accel}:
\begin{align}
\bm{d}_\mathrm{gyr} &= \bm{\omega}(t)- \left(\bm{\omega}_\mathrm{m} - \bm{b}_\mathrm{gyr}(t)\right),\label{eq:dgyr}\\
\bm{d}_\mathrm{acc} &= R(t)^\intercal \left( \bm{a}(t) + \bm{g}_\mathrm{w}\right)  - \left(\bm{a}_\mathrm{m} - \bm{b}_\mathrm{acc}(t)\right)\label{eq:dacc}.
\end{align}

However, \acsp{IMU}' high sampling frequencies lead to many additional terms in \refeq{eq:lio_opt} which increase the computational load for optimization.
Preintegration~\cite{leutenegger2015preint, usenko2020basalt, shan2020liosam} is a convenient method to combine multiple \acs{IMU} measurements.
Here, a single preintegrated pseudo-measurement has mean $\Delta_{\mathrm{pre}}$ and covariance $\Sigma^{\Delta_{\mathrm{pre}}} \in \mathbb{R}^{9\times 9}$.
The integration~\cite{usenko2020basalt} starts at time $t_{\mathrm{m}_i}$ with linearized biases $\bm{b}_{\mathrm{acc},i},\bm{b}_{\mathrm{gyr},i}$ using the bias-corrected \acs{IMU} measurements $\left(\overline{\bm{\omega}},\overline{\bm{a}}\right)_{\mathrm{m}_{j+1}}$.
We use the subscript $\mathrm{m}_j$ to emphasize the last integrated measurement with time $t_{\mathrm{m}_j}$:
\begin{align}
\Delta_{\mathrm{pre}} &= \left(\Delta {\bm{p}_{\mathrm{m}_j}}, \Delta {R_{\mathrm{m}_j}, \Delta {\bm{v}_{\mathrm{m}_j}} }\right)\in \RDSoR.
\end{align}
The covariance $\Sigma^{\Delta_\mathrm{pre}}$ is propagated accordingly with fixed measurement noises $\Sigma_{\mathrm{acc}}, \Sigma_{\mathrm{gyr}}$.

For simplicity, we use the shorthand $f_i$ (resp. $f_j$) for a B-spline function $f(t)$ evaluated at $t_{\mathrm{m}_i}$ (resp. $t_{\mathrm{m}_j}$) and drop the subscript for preintegrated variables from $t_{\mathrm{m}_i}$ to $t_{\mathrm{m}_j}$.
For now, we ignore the biases to define the errors for $\Delta_{\overline{\mathrm{pre}}}=\left(\Delta \overline{\bm{p}},\Delta \overline{R},\Delta \overline{\bm{v}}\right)$ with $\Delta t_{\mathrm{m}} = t_{\mathrm{m}_j} - t_{\mathrm{m}_i}$:
\begin{align}
\bm{d}_{\relPosPre} &= R_{i}^\intercal \overbrace{\left(\posj - \posi - \veli\Delta t_{\mathrm{m}} - \bm{g}_\mathrm{w}\frac{\Delta t_{\mathrm{m}}^2}{2}\right)}^{\relPosPre} - \Delta \overline{\bm{p}},\label{eq:dpos}\\
\bm{d}_{\relRotPre} &= \mathrm{Log}\left(\Delta \overline{R} \Rjt \Ri \right),\label{eq:drot}\\
\bm{d}_{\relVelPre} &= R_{i}^\intercal \left(\velj - \veli - \bm{g}_\mathrm{w}\Delta t_{\mathrm{m}} \right) - \Delta \overline{\bm{v}},\label{eq:dvel}.
\end{align}
While the Jacobians depending on a single time $t$ are given\footnote{\textcite{sommer2020cvpr} provide the right Jacobians for orientation $R(t)$, angular rate $\bm{\omega}(t)$ and angular acceleration $\bm{\alpha}(t)$, as well as position $\bm{p}(t)$, velocity $\bm{v}(t)$ and linear acceleration $\bm{a}(t)$.} in \cite{sommer2020cvpr}, we need to derive the right Jacobians $\Jfxdx$ \wrt knots $X_{k_i}$ and $X_{k_j}$ for the above \textit{relative} residuals $\bm{d}_{\Delta_\mathrm{pre}} = [\bm{d}_{\relPosPre}^\intercal,\bm{d}_{\relRotPre}^\intercal,\bm{d}_{\relVelPre}^\intercal]^\intercal$. 

\textcite{sommer2020cvpr} defined the composite manifold\footnote{The definition differs from \textcite{sola2021lie} who use right-$\oplus$ and right-$\ominus$, thus, leading to different Jacobians derived for $\SeD$ and $\SoR$.} for a spline knot $X \in \SoR$ with composition ($\circ$), a left-increment update ($\oplus$) and a right-decrement downdate ($\ominus$):
\begin{align}
\mathrm{Exp}(\bm{\tau}) &= \exp\left(\bm{\tau}^{\wedge}\right) = X \in \SoR,\\
\mathrm{Log}(X) &= \log\left(X\right)^{\vee} = \bm{\tau} \in \mathbb{R}^6,\label{eq:Log}\\
Y &= X \oplus \bm{\tau} = \mathrm{Exp}(\bm{\tau}) \circ X, \text{ with } \bm{\tau} \in \mathbb{R}^6\\
\bm{\tau} &= Y \ominus X = \mathrm{Log}(X\inv \circ Y), % A ominus B == Log( B^-1 circ A )
\end{align}
 as well as a right Jacobian $\Jfxdx \in \mathbb{R}^{6\times 6}$:
\begin{subequations}
\begin{align}
\Jfxdx(\bm{\tau}) &= \limTau{f(X\oplus \bm{\tau}) \ominus f(X)},\\
&= \limTau{\mathrm{Log}(f(X)\inv \circ f(X\oplus \bm{\tau}))},\\
&= \left.\frac{\partial \mathrm{Log}\left(f(X)\inv \circ f(X\oplus \bm{\tau})\right)}{\partial \bm{\tau}}\right|_{\bm{\tau}=\bm{0}}.
\end{align}
\end{subequations}
The adjoint $\mathrm{Adj}_X$ is a linear map defined as:
\begin{align}
\mathrm{Adj}_X &= (X \bm{\tau}^\wedge  X\inv)^\vee\label{eq:adj}.
\end{align}
The following identities\footnote{\refeq{eq:xtang} is given in \textcite[eq. (20)]{sola2021lie}. For $\SoD$, \refeq{eq:RwR} is a direct result from the adjoint \refeq{eq:adj} with \refeq{eq:skewT}.} are helpful for further derivations:
\begin{align}
\exp(X\bm{\tau}^\wedge X\inv) &= X \exp(\bm{\tau}^\wedge)X\inv,\label{eq:xtang}\\
\exp(\bm{\tau}^{\wedge})\inv &= \exp(-\bm{\tau}^{\wedge}) \underset{\bm{\tau\rightarrow \bm{0}}}{=} I-\bm{\tau}^{\wedge},\label{eq:invexp}\\
[\bm{w}]_\times \bm{v} &= -[\bm{v}]_\times \bm{w},\label{eq:skewT}\\
(R \bm{w})^{\wedge} R&= R \bm{w}^{\wedge}\label{eq:RwR}.
\end{align}

A relative B-spline function $\mathrm{g}=f_X(t_i,t_j)$ depends on two timestamps $t_i, t_j$.
Due to the local support of a B-spline, a knot $X_k$ has non-zero weight for none, one or both timestamps.
Hence, we denote the set of $N$ knots with non-zero weight at time $t$ as $\mathcal{X}(t)$.
Obviously, the Jacobian $\JrfxdX{X_k}$ of $\mathrm{g}$ \wrt $X_k$ is zero for $X_k \notin \left\lbrace\mathcal{X}(t_i)\cup \mathcal{X}(t_j)\right\rbrace$ since the weight of $X_k$ is zero.

The decoupled logarithm $\LOGD$ of $\SoR$ allows us to treat the Jacobians for the knot's rotation $\Rk$ and position $\posk$ independently: $\JrfxdX{X_k}=\left[\JrfxdX{\bm{p}_{k}},\JrfxdX{R_{k}}\right]$.
Thus, we begin with the derivation\footnote{The preintegration of Basalt uses a right-increment %(\TODO{ups?}\refeq{eq:drot_up}) 
and thus $\Rj \approx \Ri \Delta \overline{R}$ follows, which leads to \refeq{eq:drot}.} for the rotation:
\begin{align}
\relRotPre=\Delta \overline{R}\Rjt \Ri,
\end{align} in \refeq{eq:drot}.
First, we assume no overlap between $\mathcal{X}(t_i)$ and $\mathcal{X}(t_j)$ and consider both cases separately, starting with $\Rk\in\SoD$ evaluated at $t_i$:
\begin{subequations}
\begin{align}
\bm{J}_{\Rk,t_i}^{\Delta\widetilde{R}} &= \lim_{\bm{\tau} \rightarrow 0} \frac{\left(\Delta \widetilde{R}(X_{k}\oplus\bm{\tau})\right)\ominus \left(\Delta \widetilde{R}(X_{k})\right)}{\bm{\tau}},\\
 &= \lim_{\bm{\tau} \rightarrow 0} \frac{\left(\Delta \overline{R} \Rj\inv \Ri(X_{k}\oplus\bm{\tau}) \right)\ominus \left(\Delta \overline{R} \Rj\inv \Ri(X_{k})\right)}{\bm{\tau}},\label{eq:dRotlong}\\
% &= \lim_{\bm{\tau} \rightarrow 0} \frac{\mathrm{Log}\left(\left(\Delta \overline{R} \Rj\inv \Ri\right)\inv \Delta \overline{R} \Rj\inv \mathrm{Exp}(\bm{\tau}) \Ri\right)}{\bm{\tau}},\\
% &= \lim_{\bm{\tau} \rightarrow 0} \frac{\mathrm{Log}\left(\Rit {\Rj \Delta \overline{R}^\intercal \Delta \overline{R} \Rjt} \mathrm{Exp}(\bm{\tau}) \Ri\right)}{\bm{\tau}},\label{eq:dRotlong}\\
 &=\lim_{\bm{\tau} \rightarrow 0} \frac{ \mathrm{Log}\left(\Rit \mathrm{Exp}(\bm{\tau}) \Ri\right)}{\bm{\tau}} = \lim_{\bm{\tau} \rightarrow 0}\frac{\left(\Rit \bm{\tau}^\wedge \Ri\right)^\vee}{\bm{\tau}},\label{eq:dRotLogExp}\\
 &\stackrel{\refeq{eq:adj}}{=} \mathrm{Adj}_{\Rit} \bm{J}_{\Rk}^{\Ri} = \Rit \bm{J}_{\Rk}^{\Ri}.\label{eq:dRotRki}
\end{align}
\end{subequations}
Here, $R\inv = R^\intercal$ allows simplification in \refeq{eq:dRotlong} and cancels out many terms.
\refeq{eq:dRotLogExp} uses the identity in \refeq{eq:xtang} and the definition of $\mathrm{Log}(X)$ in \refeq{eq:Log}.
The equalities in \refeq{eq:dRotRki} stem from the definition of the adjoint \refeq{eq:adj} and $\mathrm{Adj}_R=R$ \cite[eq. (139)]{sola2021lie}, whereas \cite{sommer2020cvpr} provides $\bm{J}_{\Rk}^{\Ri}$.

The derivation follows similarly for $t_j$:
\begin{align}
\bm{J}_{\Rk,t_j}^{\Delta\widetilde{R}} &= -\Rit \bm{J}_{\Rk}^{\Rj}.\label{eq:dRotRkj}
\end{align}
Finally, we have for $X_k \in \left\lbrace \mathcal{X}(t_i) \cap \mathcal{X}(t_j) \right\rbrace$ with \refeq{eq:dRotRki} and \refeq{eq:dRotRkj} according to the product rule:
\begin{align}
\bm{J}_{\Rk}^{\Delta\widetilde{R}} &= \Rit \bm{J}^{\Ri}_{\Rk} - \Rit \bm{J}^{\Rj}_{\Rk}.
\end{align}
This is coincidently the general form for $\bm{J}_{\Rk}^{\Delta\widetilde{R}}$ whereas the special cases \refeq{eq:dRotRki} and \refeq{eq:dRotRkj} arise if $\mathcal{X}(t) = \emptyset$ and thus $\bm{J}^{R(t)}_{\Rk}=0_{3\times 3}$.

Evidently, the Jacobian of $\Delta\widetilde{R}$ \wrt the knot's position $\bm{J}_{\posk}^{\Delta\widetilde{R}}$ is zero and independent of $t$, 
which leads with the inverse right Jacobian $\bm{J}^{\mathrm{Log}(R)}_R=J_r\inv(\mathrm{Log}(R))$ for $\SoD$~\cite{sommer2020cvpr} to:
\begin{align}
\JrfxdXkij{\drelRotPre}{} &= \left[0_{3\times 3},J_r\inv\left(\drelRotPre\right) \bm{J}_{\Rk}^{\Delta\widetilde{R}}\right]\label{eq:Jrot}.
\end{align}

The errors for preintegrated position \refeq{eq:dpos} and velocity \refeq{eq:dvel} share a common form $\Delta\mathrm{h}$ with $f_{\posk}(t)$ depending on the knot position $\posk$ and some vector $\bm{e}\in \RD$:
\begin{align}
\Delta\mathrm{h}=\Rit \overbrace{\left( f_{\posk}(t_j) - f_{\posk}(t_i) + \bm{e} \right)}^{\bm{\mathrm{d}}}.
\end{align}
Since the target domain of $\Delta\mathrm{h}$ is $\RD$, the $\ominus$ becomes a subtraction:
\begin{subequations}
\begin{align}
\bm{J}_{\Rk,t_i}^{\Delta\mathrm{h}} &= \lim_{\bm{\tau} \rightarrow 0} \frac{\left(\Delta \mathrm{h}(X_k\oplus\bm{\tau})\right)\ominus \left(\Delta \mathrm{h}(X_k)\right)}{\bm{\tau}},\\
% &= \lim_{\bm{\tau} \rightarrow 0} \frac{ \left(\mathrm{Exp}(\bm{\tau})\Ri\right)\inv \left(f_{\posk}(t_j) - f_{\posk}(t_i) + \bm{c}\right)-\Delta\mathrm{h}}{\bm{\tau}},\\
 &= \lim_{\bm{\tau} \rightarrow 0} \frac{\Rit (I-\bm{\tau}^\wedge){\bm{\mathrm{d}}}-\Delta\mathrm{h}}{\bm{\tau}}\label{eq:dHRotRk}
 = \lim_{\bm{\tau} \rightarrow 0} \frac{-\Rit\bm{\tau}^\wedge {\bm{\mathrm{d}}}}{\bm{\tau}},\\
 &= \lim_{\bm{\tau} \rightarrow 0} \frac{-(\Rit\bm{\tau})^\wedge \Rit {\bm{\mathrm{d}}}}{\bm{\tau}}\label{eq:dHRotRwRk}
 = \lim_{\bm{\tau} \rightarrow 0} \frac{-[\Rit\bm{\tau}]_{\times} \Delta\mathrm{h}}{\bm{\tau}},\\
 &= \lim_{\bm{\tau} \rightarrow 0} \frac{[\Delta\mathrm{h}]_{\times} \Rit {\bm{\tau}}}{\bm{\tau}} = [\Delta \mathrm{h}]_{\times} \Rit \bm{J}^{\Ri}_{\Rk},\\
 \bm{J}_{\Rk,t_j}^{\Delta\mathrm{h}} &= 0_{3\times 3}.
\end{align}
\end{subequations}
Here, \refeq{eq:dHRotRk} uses the last equality of \refeq{eq:invexp} while \refeq{eq:RwR} expands \refeq{eq:dHRotRwRk} and is subsequently simplified with \refeq{eq:skewT}.

The derivation for $\Delta\mathrm{h}$ \wrt $f_{\posk}(t)$ is straightforward due to $f_{\posk}(t)\in \RD$ and the target domain $\RD$ and thus omitted for brevity:
\begin{align}
 \bm{J}_{f_{\posk}}^{\Delta\mathrm{h}} &= -\Rit \bm{J}^{f_{\posk}(t_i)}_{\posk} + \Rit \bm{J}^{f_{\posk}(t_j)}_{\posk}.
\end{align}
The Jacobian of $\Delta\mathrm{h}$ \wrt $X_k$ is:
\begin{align}
\JrfxdXkij{\Delta\mathrm{h}}{} &= \left[
%\begin{matrix}
\Rit \left(\bm{J}^{f_{\posk}(t_j)}_{\posk} - \bm{J}^{f_{\posk}(t_i)}_{\posk}\right), [\Delta \mathrm{h}]_{\times} \Rit \bm{J}^{\Ri}_{\Rk}
%\end{matrix}
\right].\label{eq:JrelDiff}
\end{align}
For $\drelVelPre$ \refeq{eq:dvel}, its Jacobian $\JrfxdXkij{\drelVelPre}{}$ directly follows from \refeq{eq:JrelDiff} with $f_{\posk}(t)=\bm{v}(t)$. Similarly, we obtain $\JrfxdXkij{\drelPosPre}{}$ for \refeq{eq:dpos}:
\begin{align}
\JrfxdXkij{\drelPosPre}{} &= \left[
%\begin{matrix}
\Rit \left(\bm{J}^{\posj}_{\posk} - \bm{J}^{\posi}_{\posk}-\Delta t_{\mathrm{m}} \bm{J}^{\veli}_{\posk} \right), [\relPosPre]_{\times} \Rit \bm{J}^{\Ri}_{\Rk}
%\end{matrix}
\right].
\end{align}

For gravity $\bm{g}_{\mathrm{w}}$, the Jacobians are straightforward with $\bm{J}^{\bm{g}_{\mathrm{w}}}_{\bm{g}}$ due to $\bm{g}\in \mathcal{S}^2$:
\begin{align}
\bm{J}_{\bm{g}_{\mathrm{w}}}^{\bm{d}_{\Delta_{\mathrm{pre}}}} = \left[\left(-\Rit \frac{\Delta t_{\mathrm{m}}^2}{2}\right)^\intercal, 0_{3\times 3}, \left(-\Rit\Delta t_{\mathrm{m}}\right)^\intercal 
\right]^\intercal \bm{J}^{\bm{g}_{\mathrm{w}}}_{\bm{g}}.
\end{align}

To update our biases, we must consider the linearization of $\bm{b}_{\mathrm{acc},i}$ and $\bm{b}_{\mathrm{gyr},i}$.
Hence, this modifies $\Delta_{\overline{\mathrm{pre}}}$ \cite{usenko2020basalt}:
\begin{align}
\bm{d}_{b_{acc}} &= \bm{b}_{\mathrm{acc}}(t_{\mathrm{m}_i})-\bm{b}_{\mathrm{acc},i},\\
\bm{d}_{b_{gyr}} &= \bm{b}_{\mathrm{gyr}}(t_{\mathrm{m}_i})-\bm{b}_{\mathrm{gyr},i},\\
\Delta_{\overline{\mathrm{pre}}} &= 
\begin{pmatrix}
&\Delta{\bm{p}} + \bm{J}^{\Delta\bm{p}}_{\bm{b}_{\mathrm{acc},i}} \bm{d}_{b_{acc}} + \bm{J}^{\Delta\bm{p}}_{\bm{b}_{\mathrm{gyr},i}} \bm{d}_{b_{gyr}},\\
&\mathrm{Exp}\left(\bm{J}^{\Delta {R}}_{\bm{b}_{\mathrm{gyr},i}}\bm{d}_{b_{gyr}}\right) \Delta {R},\\
&\Delta{\bm{v}} + \bm{J}^{\Delta\bm{v}}_{\bm{b}_{\mathrm{acc},i}} \bm{d}_{b_{acc}} + \bm{J}^{\Delta\bm{v}}_{\bm{b}_{\mathrm{gyr},i}}\bm{d}_{b_{gyr}}
\end{pmatrix}.
\end{align}
Fortunately, this does not change the general form of the above Jacobians \wrt the knot $X_k$.
With these, we obtain the Jacobians \wrt bias knots $\bm{b}_{\mathrm{acc},k},\bm{b}_{\mathrm{gyr},k}$:
\begin{align}
\bm{J}_{\bm{b}_{\mathrm{acc},k}}^{\bm{d}_{\Delta_{\mathrm{pre}}}} &= -\bm{J}_{\bm{b}_{\mathrm{acc},i}}^{\Delta_{\mathrm{pre}}}
\bm{J}^{\bm{b}_{\mathrm{acc}}(t_{\mathrm{m}_i})}_{\bm{b}_{\mathrm{acc},k}},\\
\bm{J}_{\bm{b}_{\mathrm{gyr},k}}^{\bm{d}_{\Delta_{\mathrm{pre}}}} &= - \begin{bmatrix} I_{3\times 3} & 0_{3\times 3} & 0_{3\times 3} \\ 0_{3\times 3} & J_r\inv(\relRotPre) \relRotPre^\intercal & 0_{3\times 3} \\ 0_{3\times 3} & 0_{3\times 3} & I_{3\times 3} \end{bmatrix}
\bm{J}_{\bm{b}_{\mathrm{gyr},i}}^{\Delta_{\mathrm{pre}}}
\bm{J}^{\bm{b}_{\mathrm{gyr}}(t_{\mathrm{m}_i})}_{\bm{b}_{\mathrm{gyr},k}}.
\end{align}

Empirically, we found that solely relying on preintegration may lead to slippage in the estimation (esp. in rotation) due to the change from absolute \acs{IMU} measurements $(\linAccM,\angRotM)$ to purely relative ones.
In case of high rotational velocities, \eg $\geq \SI{120}{\degree\per\second}$, we switch to raw measurements.
Otherwise, we fuse all \acs{IMU} measurements between two scans except for the last one, which allows us to define the \acs{IMU} error $\mathcal{L}_\mathrm{IMU}$:
\begin{align}
\mathcal{L}_\mathrm{raw} &= \bm{d}^\intercal_\mathrm{acc} \Sigma_\mathrm{acc}\inv\bm{d}_\mathrm{acc} + \bm{d}^\intercal_\mathrm{gyr} \Sigma_\mathrm{gyr}\inv\bm{d}_\mathrm{gyr},\\
\mathcal{L}_\mathrm{pre} &= \bm{d}_{\Delta_\mathrm{pre}}^\intercal \left(\Sigma^{\Delta_\mathrm{pre}}\right)\inv \bm{d}_{\Delta_\mathrm{pre}},\\
\mathcal{L}_\mathrm{IMU} &= w_\mathrm{IMU} \left( \mathcal{L}_\mathrm{pre} (t_{\mathrm{m}_i},\ldots,t_{\mathrm{m}_{j-1}}) + \mathcal{L}_\mathrm{raw}(t_{\mathrm{m}_j})\right).\label{eq:imu}
\end{align}

In the absence of further sensor input, we regularize the spline for $N\geq 3$ with zero-acceleration soft-constraints on linear $\linAccZ(t)$ and angular acceleration $\angAccZ(t)$ in \acs{IMU} frame with covariance $\Sigma_\mathrm{z}$ and weight $w_\mathrm{z}$:
\begin{align}
\bm{d}_\mathrm{z}(t)&=\left[\linAccZ(t)^\intercal,\angAccZ(t)^\intercal\right]^\intercal,\\
\mathcal{L}_\mathrm{z}(i) &= w_\mathrm{z} \sum_{o=0}^{O-1} \bm{d}^\intercal_\mathrm{z}(t_{\mathrm{seg}}(o)) \Sigma_\mathrm{z}\inv \bm{d}_\mathrm{z}(t_{\mathrm{seg}}(o)).\label{eq:zero_acc}
\end{align}
This promotes uniform motion towards a constant velocity model without strictly enforcing it.

\subsection{Relative Motion Constraints}\label{sec:nsc}
In some situations, a robot supplies further motion estimates, \eg from the wheel or joint encoders or \ac{VO}.
Hence, we promote similarity between robot odometry poses $\DTm=T_{\mathrm{m}_j}\inv T_{\mathrm{m}_i}$ and the spline $\DT = T_j\inv T_i$ using a relative pose error $\dDT$ with weight matrix $W_{\dDT}$:
\begin{align}
\Delta \bm{p} &= \Rit\left(\posj - \posi\right),\\
\dDT &= \left[\begin{matrix} \DRmt \left(\relPos-\Dpm\right)\\ \mathrm{Log}\left(\DRmt\Rjt\Ri\right) \end{matrix}\right],\\
\mathcal{L}_{\dDT} &= w_{\Delta T} \dDT^\intercal W_{\dDT} \dDT.\label{eq:rel_pose}
\end{align}
From \refeq{eq:JrelDiff}, we get $\JrfxdXkij{\relPos}{}$ and obtain $\JrfxdXkij{\dDT}{}$ using \refeq{eq:Jrot} with $\Delta\overline{R}=\Delta R_\mathrm{m}^\intercal$:
\begin{align}
\JrfxdXkij{\dDT}{} &= \left[\begin{matrix} \DRmt \JrfxdXkij{\Delta \bm{p}}{}\\
\JrfxdXkij{\drelRotPre}{} \end{matrix}\right].
\end{align}

Without a covariance estimate for $\DT$, $W_{\dDT}$ provides an opportunity to improve the method's resilience.
The limited \ac{fov}, the limited range of \acs{LiDAR} sensors, and the environment's geometry can lead to an uneven constraint distribution for all \acp{dof}.
An example are tunnel-like structures when there are mostly measurements of the walls, floor and ceiling.
Here, constraints for the translational \ac{dof} along the tunnel are likely underrepresented or dominated by noise~\cite{zhang2016degeneracy,tuna2024xicp}.

\textcite{nashed2021rdslam} selectively constrain translation along the Eigenvectors $\bm{v}_i$ of the normals' covariance $C_{\bm{n}}$ if the condition number  $\kappa_i = \lambda_{max} / \lambda_i$ is above $\tau_\kappa = 10$. 
We empirically found an adaptive scaling \wrt the data term $\mathcal{L}_{\mathrm{MARS}}$ and the inclusion of the orientation direction to be worthwhile:
\begin{align}
C_l &= \frac{1}{\abs{S_l}} \sum_{s\in S_l}\bm{h}_s^\intercal \bm{h}_s \text{ with }
\bm{h}_s = \left[\bm{n}_s,\bm{\mu}_s \times \bm{n}_s\right]^\intercal.
\end{align}
Hence, we scale the Eigenvectors $\bm{v}_i$ of $C_l$ with $\kappa_i$ and use the weight matrix\footnote{Since $C_l$ is real and \ac{spd}, inverting its Eigenvalues $\lambda_i$ leads to the inverse $C_l\inv=V\Lambda\inv V^\intercal$. Thus, the scaling directly provides the information matrix $W$.} $W_{l}=\sum_i \bm{v}_i \kappa_i \bm{v}_i^\intercal$ for the relative pose error $\mathcal{L}_{\dDT}$ within $(t_{l-L}, t_l]$.
A similar $W_l$ is obtained for a position-only prior $\mathcal{L}_{\Dp}$ (resp. orientation-only $\mathcal{L}_{\DR}$) from the $\num{3}\times\num{3}$ top-left (resp. bottom-right) block matrix of $C_l$.

\begin{figure}
\centering
\if\WithFigures1
\resizebox{1.0\linewidth}{!}{\input{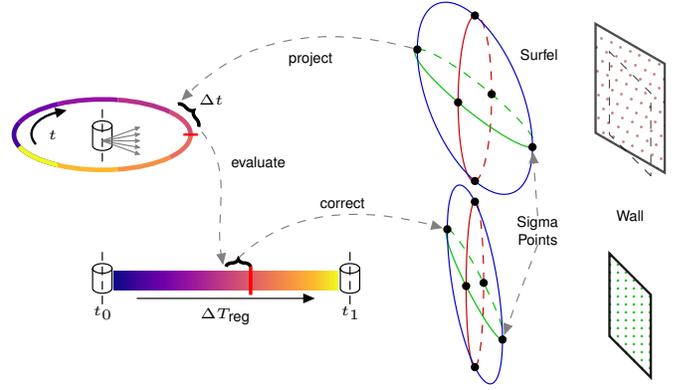}}
\fi
\vspace{-3mm}
\caption[Surfel motion compensation via \acf{UT}]{\Ac{surfel} motion compensation via \acf{UT}: sensor motion $\Delta T_{\mathrm{reg}}$ from $t_0$ to $t_1$, \eg towards a wall, distorts and skews scans and \acsp{surfel}. Sigma points $\mathcal{Y}_\Sigma$ of the \acs{surfel} covariance $\Sigma_s$ are projected into the \acs{LiDAR} to obtain the sigma points' time $t_{\bm{y}}$. Spline evaluation allows motion correction towards a segments' reference time $t_{\mathrm{seg}}$ (red) before \acsp{surfel} are re-fused.}
\label{fig:undistortion}
\end{figure}

\subsection{Unscented Transform for Motion Compensation}\label{sec:ut}
The spline allows us to compensate ego-motion towards a reference time $t_\mathrm{ref}$, given the points' capture time $t(\bm{p})$:
\begin{align}
\bar{\bm{p}} &= T_\mathcal{X}\left(t_{\mathrm{ref}}\right)\inv T_\mathcal{X}\left(t(\bm{p})\right)\bm{p}.\label{eq:compPoint}
\end{align}
Full compensation for all points with \refeq{eq:compPoint} is computationally prohibitively expensive during registration in real-time applications with modern \acsp{LiDAR}\footnote{\acs{CT-ICP}~\cite{dellenbach2021ct} reports an average computation time of \SI{430}{\milli\second} on Oxford Newer College~\cite{ramezani2020newer} to simultaneously optimize two poses for one grid sampled and linearly interpolated scan.}.
In our case, this also requires re-embedding of the point cloud. 

Instead, we adapt the \acf{UT}~\cite{julier1997ukf,wan2000unscented} for an efficient surfelwise undistortion, as shown in \reffig{fig:undistortion}.
The \ac{UT} generates from mean $\bm{\mu}$ and covariance $\Sigma$ a set of sigma points $\mathcal{Y}_\Sigma$ before applying a function $\bar{\bm{y}} = f(\bm{y}), \bm{y} \in \mathcal{Y}_\Sigma$ to these points.
Finally, the \ac{UT} recombines the transformed points to obtain the transformed mean $\bar{\bm{\mu}}$ and covariance $\bar{\Sigma}$.
Here, we use the symmetric set of sigma points $\mathcal{Y}_\Sigma=\left\lbrace\bm{\mu}\pm \bm{L}_i, i\in [0,d-1]\right\rbrace$ with the Cholesky decomposition $LL^\intercal$ of the scaled covariance ($d \cdot \Sigma_s$) with $d=3$ and $\bm{L}_i$ referring to the $i$th column of $L$.
The \acs{surfel} mean $\bm{\mu}_s$ is deliberately excluded from $\mathcal{Y}_{\Sigma}$ to prevent previously occurring numerical round-off errors~\cite[eq. (30)]{wu2006ukfNum}.

Thus far, we lack the sigma points' time $t_{\bm{y}}$.
Hence, we transform each sigma point $\bm{y} \in \mathcal{Y}_\Sigma$ into the \acs{LiDAR} frame and project using \refeq{eq:PointTime} and \refeq{eq:PointColumn} with mean directional offset $\overline{o}_c$ from the \acs{surfel} points:
\begin{align}
t_{\bm{y}} = t(T_\mathrm{s,l}\inv \bm{y}, \overline{o}_c).
\end{align}
Then, \refeq{eq:compPoint} and \refeq{eq:preorient} provide the compensated sigma point $\bar{\bm{y}}_i$ with the \acsp{surfel}' segment time $t_{\mathrm{seg}}$:
\begin{align}
\Delta R_\mathrm{\bm{y},seg} &= \mathrm{slerp}(\Delta\mathcal{R}_{\mathrm{IMU}},t_{\bm{y}} )\inv \mathrm{slerp} (\Delta \mathcal{R}_{\mathrm{IMU}}, t_{\mathrm{seg}} ),\\
\bar{\bm{y}}_i &= T_\mathcal{X}\left(t_{\mathrm{seg}}\right)\inv T_\mathcal{X}\left(t_{\bm{y}}\right) \left[\Delta R_\mathrm{\bm{y},seg} \bm{y}_i\right].
\end{align}
Here, $\Delta R_\mathrm{\bm{y},seg}$ counteracts the previous pre-orientation of $\widetilde{\mathcal{P}}_l$ in \refeq{eq:preorient}.
Afterwards, transformed sigma points $\bar{\mathcal{Y}}_\Sigma$ are re-fused as in \cite{wan2000unscented} to obtain compensated mean $\bar{\bm{\mu}}$ and covariance $\bar{\Sigma}$:
\begin{align}
\bar{\bm{\mu}} &= \sum_{\bar{\bm{y}}\in\bar{\mathcal{Y}}_\Sigma} \frac{1}{2d} \bar{\bm{y}},\quad \bar{\Sigma} = \sum_{\bar{\bm{y}}\in\bar{\mathcal{Y}}_\Sigma} \frac{1}{2d} \left(\bar{\bm{y}}-\bar{\bm{\mu}}\right) \left(\bar{\bm{y}}-\bar{\bm{\mu}}\right)^\intercal.
%\bar{\bm{\mu}} &= \sum_{\bar{\bm{y}}\in\bar{\mathcal{Y}}_\Sigma} \frac{1}{2d} \bar{\bm{y}},\\
%\bar{\Sigma} &= \sum_{\bar{\bm{y}}\in\bar{\mathcal{Y}}_\Sigma} \frac{1}{2d} \left(\bar{\bm{y}}-\bar{\bm{\mu}}\right) \left(\bar{\bm{y}}-\bar{\bm{\mu}}\right)^\intercal.
\end{align}

Although applicable during registration, \eg after each \acs{LM}-iteration, we found updating the sliding registration window $\mathcal{W}_l$ prior to registration to be sufficient.

\subsection{Keyframe Generation and Reuse}\label{sec:kf}
We combine the sliding keyframe window of \acs{MARS}~\cite{quenzel2021mars} with the keyframe storage of \acs{DLO}~\cite{chen2021dlo} and take advantage of the regular structure of the lattice.
All keyframes share an aligned common lattice with keyframe-specific shifts $\bm{\nu}_{\mathrm{k}} \in \mathbb{Z}^{d+1}$ to facilitate surfelwise fusion and ensure runtime efficiency without costly reintegration.
Moreover, shifts allow us to maintain the local property of our map $\mathcal{M}$ and keyframes.
For unseen areas, the storage approach is equivalent to the sliding keyframe window of \acs{MARS}, while the traversal of known regions allows the reuse of previous keyframes.

\acs{DLO}'s adaptive distance threshold may be consistently too large in situations with obstructions, like pillars, while the \acs{LiDAR} measures different surfaces with little overlap before reaching the threshold.
For this reason, MAD-\acs{ICP}~\cite{ferrari2024madicp} selects a recent scan to update its map if the percentage of successfully registered points drops below a certain threshold.
Similarly, the SOD metric of AS-\acs{LIO}~\cite{zhang2024aslio} computes the overlap between the voxelized scan and map to adapt the length of the sliding window before merging the window into the map.
Instead, we create a keyframe from the oldest scan within the registration window if less than \SI{80}{\percent} of \acsp{surfel} in $\mathcal{S}$ are associated.

As the measured ranges and thus necessary map sizes vary with the surrounding environment, we adapt the map size by changing the coarse cell size $c_0$. 
While $c_0=\SI{4}{\metre}$ is sufficient for open surroundings, half the size (\SI{2}{\metre}) is more than enough for close quarters.  
This coincides by design with the coarse $c_1$  due to $c_l = c_0 \cdot 2^{-l}$ with $l \in [0,\ldots, 3]$.
With four levels, the finest outdoor cell size $c_3$ is \SI{0.5}{\metre} compared to \SI{0.25}{\metre} indoors.
Furthermore, both coarse cell sizes have three map resolutions in common.
Thus, we seamlessly transition between narrow and wide areas by fusing map levels with the same cell size.
The current coarse cell size $c_0$ depends on the mean \acs{surfel} distance filtered with an \acl{EWMA}:
\begin{align}
\overline{d}_m &= (1-\gamma)\overline{d}_m + \frac{\gamma}{\abs{S_l}}\left(\sum_{s\in S_l}\norm{\bm{\mu}_s}\right),\\
c_0 &= \left\lbrace\begin{matrix}
\SI{4}{\metre}, &\text{if } \overline{d}_m \geq \SI{3}{\metre},\\
\SI{2}{\metre}, &\text{else}, &
\end{matrix}\right.
\end{align} similar to \acs{DLO}'s adaptive distance threshold. 

After a positive keyframe decision, we correct the ego-motion for the keyframe scan pointwise with \refeq{eq:compPoint} using $t_{s_j}$ as the reference time.
In contrast to scans in $\mathcal{W}_l$, we embed each keyframe locally at the closest vertex after rotating the points into the common map frame.
As a result, we set the keyframe-specific shift $\bm{\nu}_\mathrm{k}$ to the closest vertex $\bm{c}_{\mathrm{o}} \in \mathbb{Z}^{d+1}$.
Using the coarsest level has the advantage that a shift on any finer resolution is a whole-numbered multiple of the coarse shift.

Our local map selection initially adds a third of the map window from the closest keyframes, sorted by ascending distance.
For the remaining keyframes in range, we compute the \acs{iou} between a keyframe and current scan using the coarsest common map level.
Embedding the current scan's coarse \acs{surfel} mean $\bm{\mu}_s$ into each keyframe would be computationally involved.
Instead, we embed them once into the common map lattice.
Afterwards, we only require the cell shift between a keyframe's origin and the scan's origin.
Adding this cell shift to the embedded indices provides the corresponding indices within the keyframe as if we embedded the coarse \acsp{surfel} directly.
We fill the map $\mathcal{M}$ with up to $F$ keyframes from all overlapping ones while preferring older keyframes to reduce drift over time.

\subsection{Implementation}\label{sec:kron_sym}

Instead of the full $3\times 3$ covariance, we use the symmetry of the covariance matrix $\Sigma_s = \Sigma_s^\intercal$ for vectorization.
This allows us to store the lower triangular matrix of $\Sigma_s$ using the following $vec_L\left(\cdot\right)\in\RE$ and recover it with the inverse $sym_L\left(\bm{a}\right)$-operation:
\begin{align}
vec_L\left(A\right) &= [A_{00},A_{11},A_{22},0,A_{10},A_{20},A_{21},0]^\intercal,\label{eq:vecL}\\
sym_L\left(\bm{a}\right) &= \begin{pmatrix}
a_{0},a_{4},a_{5},\\ 
a_{4},a_{1},a_{6},\\ 
a_{5},a_{6},a_{2}
\end{pmatrix}.
\end{align}
For $A=A^\intercal$, it is easy to verify the identities 
$sym_L\left(vec_L\left(A\right)\right)=A$, and $vec_L\left(sym_L\left(\bm{a}\right)\right)=\bm{a}$.
The vectorization simplifies memory alignment and use of \acs{SIMD} instruction sets, \eg \acs{AVX}, with the vector class library~\cite{vcl}.
Addition and computation of the outer product in the surfel computation % \refeq{eq:surfel_add} 
become a single \acs{FMA} operation per point $\posi$.

For an efficient recomputation of $\Sigma$ using $\widetilde{D}$ and \refeq{eq:eig_decomp}, we make use of the Kronecker product $\otimes$ and its relation to the $vec$-function that stacks the columns of a matrix~\cite{lancaster1985kron}:
\begin{align}
vec(A \cdot B \cdot C) &= \left(C^\intercal \otimes A\right) \cdot vec(B).\label{eq:kron_vec}
\end{align}
Combining \refeq{eq:eig_decomp} and \refeq{eq:planar_cov} with \refeq{eq:kron_vec} gives:
\begin{align}
vec(V \cdot \widetilde{D} \cdot V^\intercal) &= (V\otimes V)\cdot vec(\widetilde{D}).\label{eq:kron_eig}
\end{align}
The diagonal matrix $\widetilde{D}$ has just three non-zero entries ($\widetilde{\bm{\lambda}}$).
Hence, not the full $V\otimes V$ is required:
\begin{align}
(V\otimes V)\cdot vec(\widetilde{D}) &= \begin{bmatrix}V \cdot \left(\widetilde{\bm{\lambda}}^\intercal \odot \left[v_{00},v_{01},v_{02}\right]\right)^\intercal\\
V \cdot \left(\widetilde{\bm{\lambda}}^\intercal \odot \left[v_{10},v_{11},v_{12}\right]\right)^\intercal\\
V \cdot \left(\widetilde{\bm{\lambda}}^\intercal \odot \left[v_{20},v_{21},v_{22}\right]\right)^\intercal
\end{bmatrix},\label{eq:kron_reduced_eig}
\end{align} where the Hadamard product $\odot$ \cite{KRON} performs element-wise multiplication.
Since $V\cdot \widetilde{D}\cdot V^\intercal$ is symmetric, we rephrase \refeq{eq:kron_reduced_eig} with \refeq{eq:vecL} and $E_V = \left[\bm{v}_{0:}^\intercal, \bm{v}_{0:}^\intercal, \bm{v}_{1:}^\intercal \right] \odot \left[\bm{v}_{1:}^\intercal, \bm{v}_{2:}^\intercal, \bm{v}_{2:}^\intercal\right]$ as:
\begin{align}
%E_V &= \left[\bm{v}_{0:}^\intercal, \bm{v}_{0:}^\intercal, \bm{v}_{1:}^\intercal \right] \odot \left[\bm{v}_{1:}^\intercal, \bm{v}_{2:}^\intercal, \bm{v}_{2:}^\intercal\right],\\
%E_V &= \left[\left(\bm{v}_{0:} \odot \bm{v}_{1:}\right)^\intercal, \left(\bm{v}_{0:} \odot \bm{v}_{2:}\right)^\intercal, \left(\bm{v}_{1:} \odot \bm{v}_{2:}\right)^\intercal\right],\\
vec_L(V \cdot \widetilde{D} \cdot V^\intercal) &= \left[\left(V \odot V\right)^\intercal, \bm{0}, E_V^\intercal, \bm{0} \right]^\intercal \cdot \widetilde{\bm{\lambda}},\label{eq:kron_sym}
\end{align}
where $\bm{v}_{i:}$ is the $i$th row vector of $V$.
This allows an efficient \acs{SIMD} implementation using two \acs{FMA} and three multiply instructions.

Similar optimizations with \refeq{eq:kron_vec} are applicable for \refeq{eq:rot_sigma} to rephrase $R\Sigma_s R^\intercal$ into a single matrix-vector-product.
We introduce $H_1$ to map symmetric matrices from $vec_L(\cdot)\in\RE$ to $vec(\cdot)\in\RX$ and 
$H_2$ for mapping $vec(\cdot)\in\RX$ to $vec_L(\cdot)\in\RE$ using the basis vectors $\bm{e}_i = \left[0,\ldots,\delta_{ii},\ldots,0\right]^\intercal $ with the Kronecker delta $\delta_{ij}$:
\begin{align}
H_1 & = \left[\bm{e}_0, \bm{e}_4, \bm{e}_5, \bm{e}_4, \bm{e}_1, \bm{e}_6, \bm{e}_5, \bm{e}_6, \bm{e}_2
\right]^\intercal,\\
H_2 &= \left[\bm{e}_0, \bm{e}_4, \bm{e}_5,\;\,\bm{0}, \bm{e}_1, \bm{e}_6, \;\,\bm{0},\;\,\bm{0}, \bm{e}_2\right].
\end{align}
Thus, simplifying the use of the Kronecker product with $vec_L(\cdot)$ such that:
\begin{align}
Z &= H_2 \left( R \otimes R \right) H_1,\\
sym\left(\left( R \otimes R \right) vec\left(\Sigma_s\right)\right) &= sym_L\left(Z \cdot vec_L(\Sigma_s)\right).
\end{align}
The matrix $Z$ is constant for all \acsp{surfel} $(\gtrsim\num{100})$ at time $t$, allowing the precomputation of $Z$ and frequent reuse for \refeq{eq:rot_sigma}:
\begin{align}
vec_L(\Sigma_m + R\Sigma_s R^\intercal) &= Z \cdot vec_L(\Sigma_s) + vec_L(\Sigma_m).
\end{align}
An efficient implementation needs just six \acs{FMA} instructions since $vec_L(\cdot)$ has two zero-entries.

\subsubsection*{Mahalanobis Distance}
Each registration iteration computes more than \num{10000} Mahalanobis distances in \refeq{eq:mahalanobis}.
Every distance requires the inversion of a symmetric $3\times 3$ matrix $A$.
In general, explicit matrix inversion is discouraged~\cite{higham2002numerical}, \eg when solving linear systems as the obtained solutions are less accurate for ill-conditioned matrices.
However, a \acs{surfel} integrates only local information due to the subdivision by the lattice and the fusion of a small number of spatially distributed scans.
Furthermore, the \ac{GMM} includes a resolution-depending scaling term $\sigma^2_lI$ that reduces the condition number.
As a result, we use the analytical inverse for $A$~\cite{gantmakher1960matrices}:
\begin{align}
A\inv &= \frac{1}{\det(A)} \begin{bmatrix}
\left(\bm{A}_{1:}\times \bm{A}_{2:}\right)^\intercal\\ 
\left(\bm{A}_{2:}\times \bm{A}_{0:}\right)^\intercal\\ 
\left(\bm{A}_{0:}\times \bm{A}_{1:}\right)^\intercal
\end{bmatrix}.\label{eq:sym_inv}
\end{align}
The symmetry of $A$ allows further simplification with $a_{01} = a_{10}$, $a_{02} = a_{20} $ and $a_{12} = a_{21}$.
We rephrase \refeq{eq:sym_inv} with $\bm{m} = vec_L(A)$ as:
\begin{subequations}
\begin{align}
\bm{a}_0&=[m_1,  m_0,  m_0, 0,  m_5,  m_4,  m_5, 0]^\intercal,\\
\bm{b}_0&=[m_2,  m_2,  m_1, 0,  m_6,  m_6,  m_4, 0]^\intercal,\\
\bm{a}_1&=[m_6,  m_5,  m_4, 0,  m_4,  m_5,  m_6, 0]^\intercal,\\
\bm{b}_1&=[m_6,  m_5,  m_4, 0,  m_2,  m_1,  m_0, 0]^\intercal,\\
\bm{c} &= \bm{a}_0 \odot \bm{b}_0 - \bm{a}_1 \odot \bm{b}_1,\label{eq:inv_c}\\
%\det(A) &= m_0 c_0 + m_4 c_4 + m_5 c_5,\\
%A\inv &= sym_L\left( \frac{1}{\det(A)} \bm{c} \right).\label{eq:inv_det}
A\inv &= sym_L\left( \frac{1}{m_0 c_0 + m_4 c_4 + m_5 c_5} \bm{c} \right).\label{eq:inv_det}
\end{align}
\end{subequations}
An efficient implementation needs one multiply and one \acs{FMA} operation for \refeq{eq:inv_c} in addition to the dot product for $\det(A)$ and one division.

\subsubsection*{Embedding into the Permutohedral Lattice}\label{sec:embedd}
The embedding of every point $\pos$ into the permutohedral lattice~\cite{quenzel2021mars} is the most costly process during map creation.
After lifting $\pos$ and scaling with $\sigma_l\inv$, we compute the closest remainder-0 point $\bm{\mathrm{y}} \in H_d$ of the simplex\footnote{see Lemma 2.9 and Fig. 3 in \cite{adams2010lattice}}.
For this, we first unroll the rank computation into $d+1$ parallel rounds.
The rank $\bm{r}_H=P_H \cdot \left[0,\ldots,d\right]^\intercal$ represents an unsorted permutation of $(0,\ldots, d)$ with permutation matrix $P_H$.

The next step sets the distances $\bm{d}$ to the permuted position ($P_H \bm{b} := \bm{d}$) to obtain the barycentric coordinates.
This typical ``scatter''-operation ($b[ind[i]]:=d[i]$) has only been recently added for memory access within AVX512~\cite{intel2023simd}, but remains unavailable for registers.

Instead, permuting the distances inversly\footnote{The inverse of a permutation matrix is its transpose~\cite{pissanetzky1984permute}.} ($\bm{b} := P_H^\intercal\bm{d}$) allows to use efficient gather-operations ($b[i]:=d[idx[i]]$), \eg shuffle of SSE3~\cite{intelSSE3shuffle} and \acs{AVX} permute~\cite{intelAVXpermute} instructions.
We verified by enumeration for $d=3$ 
that sorting the rank $\bm{r}_H$ and setting $idx[i] = ind(sorted[i])$ indeed computes the correct index. 
Hence, we encode the original index ($0,\ldots,d$) in the lower $\log_2(d+1)=2$ bits after shifting the rank by the same number of bits.
Then, we sort blocks of $d+1=4$ integers in parallel using three min- and max-operations.

After extracting the original index offset and adding the block offset, we shuffle according to the new index $idx[i]$.
The barycentric coordinates become readily available as the difference between neighboring entries\footnote{see Proposition 4.2 in \cite{adams2010lattice}}.
We retain $\bm{\mathrm{y}}$ as the one with the highest barycentric weight.

\bgroup
\newcolumntype{C}[1]{>{\centering\arraybackslash\hspace{0pt}}m{#1}}
\newcolumntype{A}{>{\raggedright\arraybackslash\hspace{0pt}}m{1.1cm}}
\newcolumntype{Z}{>{\centering\arraybackslash}X}
\newcolumntype{B}{>{\raggedright\arraybackslash}X}
\newcolumntype{L}{>{\raggedleft\arraybackslash}X}
\renewcommand{\arraystretch}{1.2}  %Vertical space: 1 is the default, change 
\begin{table}
\begin{threeparttable}[b]
\scriptsize%\tiny
\centering
\caption[RMS-ATE evaluation of LO/LIO methods on NC]{\acs{RMS}-\acs{ATE}~[\si{\metre}] evaluation on the Newer College~\cite{ramezani2020newer} dataset\tnote{$\ast$}}
\label{tab:lio_nc}
\setlength{\tabcolsep}{1pt}
\begin{tabularx}{\linewidth}{AZZZZZZZZZZ}
\toprule
  & \RotD{MARS} & \RotD{GenZ-ICP} & \RotD{Fast-LIO} & \RotD{Point-LIO} & \RotD{SLICT2} & \RotD{DLIO} & \RotD{SE-LIO} & \RotD{iG-LIO} & \RotD{RESPLE} & \RotD{LIO-MARS} \\
\midrule\midrule
\ac{LIO}            & \xmark & \xmark & \cmark & \cmark & \cmark & \cmark & \cmark & \cmark & \cmark & \cmark \\
CT\tnote{$\dagger$} & \cmark & \xmark & \xmark & \xmark & \cmark & \xmark & \cmark & \xmark & \cmark & \cmark \\
\midrule
% rounded to 3:
01\_Short & 1.110 & 0.419 & \pc{0.340} & 0.376 & 0.363 & 0.355 & \pb{0.291} & \pa{0.282} & 0.529 & 0.601\\
 02\_Long & 3.198 & 1.086 & \pb{0.329} & \tbfX & 0.399 & 0.384 & \pa{0.301} & \pc{0.338} & 0.460 & 0.396\\
 05\_Quad & 0.292 & 0.122 & \pc{0.111} & 0.171 & \pa{0.109} & 0.120 & 0.113 & \pa{0.109} & \tbfX & 0.132\\
 06\_Spin\tnote{$\star$} & 0.105 & \tbfX & 0.091 & \tbfX & 0.093 & 0.134 & 0.087 & 0.093 & 0.088 & 0.095\\
 07\_Park & 2.278 & 0.191 & \pa{0.125} & \pc{0.138} & 0.139 & 0.139 & 0.140 & \pb{0.129} & \tbfX & 0.169\\
\midrule\midrule
Fail [\%] & \tbfZ & 20.00 & \tbfZ & 24.00 & \tbfZ & \tbfZ & \tbfZ & \tbfZ & 12.00 & \tbfZ\\
Avg. Rank &7.40 & 5.80 & \pb{1.80} & 5.40 & 3.20 & 3.40 & \pc{2.60} & \pa{1.40} & 7.00 & 5.60\\
Overall & 10. & 8. & \pb{2.} & 6. & 4. & 5. & \pc{3.} & \pa{1.} & 9. & 7.\\
\bottomrule
\end{tabularx}
\begin{tablenotes}
\item[$\ast$] Algorithms are grouped by \acs{LO}/\acs{LIO} and ordered according to publication date. An ``X'' marks divergence. Lower values are better ($\downarrow$) with \pc{third}, \pb{second} and \pa{best} highlighted.
\item[$\dagger$] Continuous-time trajectory.
\item[$\star$] Inconsistent ground-truth, excluded from ranking. Failures manually inspected and verified.
\end{tablenotes}
\end{threeparttable}
\end{table}
\egroup

\bgroup
\newcolumntype{C}[1]{>{\centering\arraybackslash\hspace{0pt}}m{#1}}
\newcolumntype{A}{>{\raggedright\arraybackslash\hspace{0pt}}m{1.1cm}}
\newcolumntype{Z}{>{\centering\arraybackslash}X}
\newcolumntype{B}{>{\raggedright\arraybackslash}X}
\newcolumntype{L}{>{\raggedleft\arraybackslash}X}
\renewcommand{\arraystretch}{1.1}  %Vertical space: 1 is the default, change 
\begin{table}[t]
\begin{threeparttable}[b]
\scriptsize%\tiny
\centering
\caption{\acs{RMS}-\acs{ATE}~[\si{\metre}] evaluation on the Newer College extension~\cite{zhang2021newerext}\tnote{$\ast$}}
\label{tab:lio_newer_college}
\setlength{\tabcolsep}{1pt}
\begin{tabularx}{\linewidth}{AZZZZZZZZZZ}
\toprule
 & \RotD{MARS} & \RotD{GenZ-ICP} & \RotD{Fast-LIO} & \RotD{Point-LIO} & \RotD{SLICT2} & \RotD{DLIO} & \RotD{SE-LIO} & \RotD{iG-LIO} & \RotD{RESPLE} & \RotD{LIO-MARS} \\
\midrule\midrule
\ac{LIO}            & \xmark & \xmark & \cmark & \cmark & \cmark & \cmark & \cmark & \cmark & \cmark & \cmark \\
CT\tnote{$\dagger$} & \cmark & \xmark & \xmark & \xmark & \cmark & \xmark & \cmark & \xmark & \cmark & \cmark \\
\midrule
% rounded to 3:
 Math-E & 0.151 & 0.139 & \pb{0.080} & 0.153 & 0.145 & \pc{0.131} & 0.138 & \pa{0.062} & 0.191 & 0.157\\
 Math-M & 0.187 & 0.206 & \pb{0.106} & 0.181 & \pc{0.126} & 0.145 & 0.206 & \pa{0.101} & 0.189 & 0.141\\
 Math-H & 0.135 & 1.883 & \pb{0.066} & 0.138 & 0.132 & \pc{0.070} & 0.110 & \pa{0.062} & 0.134 & 0.112\\
 Mine-E & 0.087 & 0.079 & 0.049 & \pc{0.045} & 0.054 & \pa{0.036} & 0.089 & 0.052 & 0.055 & \pb{0.041}\\
 Mine-M & 0.094 & 0.092 & \pb{0.046} & \pb{0.046} & 0.051 & 0.047 & 0.312 & 0.055 & 0.058 & \pa{0.044}\\
 Mine-H & 0.114 & 0.130 & \pb{0.053} & \pa{0.052} & 0.068 & 0.071 & 0.120 & \pc{0.056} & 0.384 & \pc{0.056}\\
 Quad-E & 0.152 & 0.079 & \pc{0.070} & 0.074 & 0.071 & \pb{0.069} & 0.082 & \pc{0.070} & 0.077 & \pa{0.067}\\
 Stairs & \tbfX & \tbfX & \tbfX & 0.222 & \tbfX & \pb{0.117} & \tbfX & 0.333 & \pc{0.134} & \pa{0.103}\\
 Quad-M & 0.117 & 0.098 & \pa{0.060} & 0.066 & 0.064 & \pc{0.062} & \tbfX & \pc{0.062} & 0.073 & \pb{0.061}\\
 Quad-H & 0.306 & 0.117 & \pa{0.056} & 0.066 & 0.103 & 0.069 & \tbfX & \pb{0.062} & 0.083 & \pc{0.063}\\
 Park & 2.931 & 0.812 & 0.326 & 1.376 & 0.331 & 0.308 & \pc{0.298} & \pa{0.263} & 2.143 & \pb{0.288}\\
 Cloister & 0.303 & 0.166 & \pb{0.073} & 0.103 & 0.116 & \pc{0.095} & 0.108 & \tbfX & 0.103 & \pa{0.069}\\
\midrule\midrule
Fail [\%]& 8.33 & 8.33 & 5.00 & \tbfZ & 1.66 & \tbfZ & 25.00 & 3.33 & \tbfZ & \tbfZ\\
Avg. Rank & 8.67 & 8.17 & \pb{3.00} & 5.08 & 5.92 & \pc{3.42} & 7.75 & \pc{3.42} & 7.08 & \pa{2.83}\\
Overall & 10. & 9. & \pb{2.} & 5. & 6. & \pc{3.} & 8. & \pc{3.} & 7. & \pa{1.}\\
\bottomrule
\end{tabularx}
\begin{tablenotes}
\item[$\ast$] Algorithms are grouped by \acs{LO}/\acs{LIO} and ordered according to publication date. An ``X'' marks divergence. Lower values are better ($\downarrow$) with \pc{third}, \pb{second} and \pa{best} highlighted.
\item[$\dagger$] Continuous-time trajectory
\end{tablenotes}
\end{threeparttable}
\end{table}
\egroup

\section{Evaluation}\label{sec:lio_eval}
All experiments were conducted on a laptop with an AMD Ryzen 7 5800H and \SI{48}{\giga\byte} of \ac{RAM}.
We use the Oxford Newer College (Extension) dataset~\cite{ramezani2020newer,zhang2021newerext}, the DRZ Living Lab dataset~\cite{quenzel2021mars}, and the Multi-Campus dataset (MCD)~\cite{nguyen2024mcdviral} for evaluation.
The selected datasets pose unique challenges due to their different characteristics.
The handheld sensor motion of Newer College is, in general, slower yet more abrupt and fits through narrower passages like corridors.
In contrast, \ac{UAV} flights achieve higher accelerations and rotational speeds while exhibiting more continuous movement.
MCD exhibits a mixture of both with handheld and ground vehicle sequences around three campuses.

We compare our system, called \acs{LIO-MARS}, to multiple other methods, which can be divided into \acp{LO} and \acp{LIO}.
For the real-time \ac{LO} baselines, we evaluated \acs{MARS}~\cite{quenzel2021mars}, \acs{DLO}~\cite{chen2021dlo}, KISS-\acs{ICP}~\cite{vizzo2022kissicp} and GenZ-\acs{ICP}~\cite{lee2024genzicp}.
From these, we report only the overall best performing \ac{LO} GenZ-\acs{ICP} and our system's direct predecessor \acs{MARS}.
The online \ac{LIO} systems include Fast-\acs{LIO}2~\cite{xu2021fastlio2}, \acs{DLIO}~\cite{chen2023dlio}, \acs{SE-LIO}~\cite{yuan2023selio}, SLICT2~\cite{nguyen2024slict2}, Point-\acs{LIO}~\cite{he2023pointlio}, iG-\acs{LIO}~\cite{chen2024iglio} and RESPLE~\cite{cao2025resple}.

We further tested Traj-\acs{LO}~\cite{zheng2024trajlo}, CLINS~\cite{lv2021clins} and Coco-\acs{LIC}~\cite{lang2023cocolic}.
These systems only provided interfaces for offline processing of \acs{ROS} bags\footnote{\url{http://wiki.ros.org/rosbag}}.
Coco-\acs{LIC} routinely diverged using its \ac{LIO} mode.
Traj-\acs{LO} and CLINS were too slow for real-time processing while their results remained below average.
As such, we only report the results for two \acs{LO} and the seven online \acs{LIO} systems.

For better comparability, we disabled the loop-closing of SLICT2, as none of the other methods have explicit loop-closing.
All methods receive data from the same rotating \acs{LiDAR} and \ac{IMU} in case of multiple sensors.
If available, the algorithms use per dataset the recommended parameters by the respective authors.
In their absence, we adapt the parameters from a similar dataset and set the intrinsics for \acs{IMU} and \acs{LiDAR} as well as extrinsics according to the dataset's calibration.
Each algorithm uses a single parameter set per dataset without per-sequence adaptation.

Our evaluation uses Evo~\cite{grupp2017evo} to compute the \ac{RMS}-\ac{ATE}~\cite{zhang2018eval} \wrt the dataset's reference poses after $\SeD$-alignment.
Per dataset, each sequence runs in real-time with a single algorithm under evaluation.
After processing the scan, we store the current pose. We repeat the evaluation five times for each method to report the average \ac{RMS} \ac{ATE}.
If, during a single run, the length of the estimated trajectory differs from the ground-truth length by more than \SI{25}{\percent} or the mean error is above \SI{50}{\metre}, we mark the algorithm as diverged for the sequence.
We report the percentage of failed runs per dataset.

Given the large number of methods under evaluation, a clear overall winner may not be apparent.
A single run with a higher error or divergence can impact the result too negatively when comparing the mean \ac{RMS}-\ac{ATE} over all sequences.
Discarding the diverged runs would bias the comparison towards successful runs without penalty for failure.
Per sequence, we thus rank the $n$ algorithms based on their \ac{RMS}-\ac{ATE} from first to $n$th place and compute the average rank over all sequences.
All diverged methods receive the last place for the corresponding sequence.

\begin{figure}[t]
\centering
\if\WithFigures1
  \resizebox{1.0\linewidth}{!}{\input{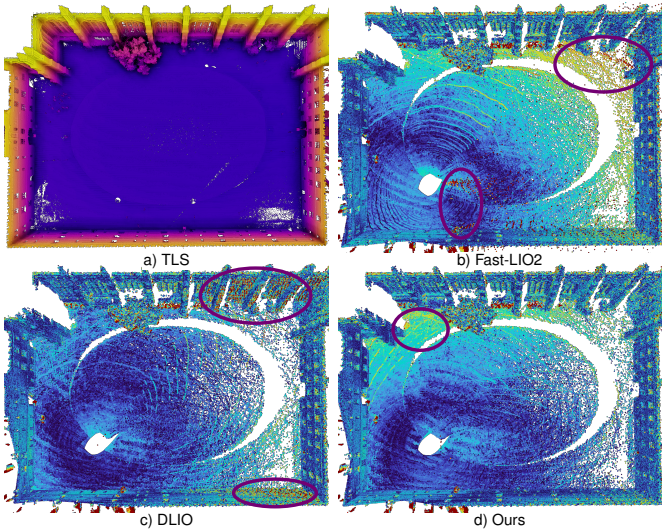}}
\fi
\caption[Comparison of undistortion after aggregation]{Comparison after compensation on ``{06\_Spin}'' of Newer College~\cite{ramezani2020newer}: height-colored ground-truth point cloud from \ac{TLS} BLK360 [a)], aggregation of every 25th scan compensated by Fast-\acs{LIO}2 [b)], \acs{DLIO} [c)] and our \acs{LIO-MARS} [d)]. Color in [b)-d)] encodes point-to-plane error \wrt ground-truth \acs{TLS} map from low (blue) to high (red, $\geq\SI{0.25}{\metre}$). Ellipses highlight areas with noticable differences between methods.}
\label{fig:nc_spin}
\end{figure}

\subsection{Newer College Dataset}
This dataset~\cite{ramezani2020newer} contains five sequences\footnote{The sequences are not consecutively numbered: 01, 02, 05, 06, 07.} captured with a handheld Ouster OS1-64 \acs{LiDAR} with integrated \acs{IMU}.
\textcite{zhang2021newerext} prepared a multi-camera-inertial-\acs{LiDAR} extension with an Ouster OS0-128 for another \num{12} sequences, partially recreating the original dataset.
During the evaluation, all algorithms receive the OS1-64 scans and its \acs{IMU} for the five Newer College sequences and the OS0-128 scans with an external Alphasense \acs{IMU} for the extension.
The reference poses stem from \ac{ICP} registration\footnote{\textcite{ramezani2020newer} do not report motion compensation.} against \ac{TLS} point clouds.
\Reftab{tab:lio_nc} presents the results for the five original sequences, whereas \reftab{tab:lio_newer_college} shows the \num{12} newer sequences.
The values for \acs{MARS} slightly differ from \cite{quenzel2021mars} due to several bug fixes after its original publication and rerunning the method.

As expected, the \ac{LIO} systems outperform their \ac{LO} counterparts since the \acs{IMU} provides valuable complementary information to \acs{LiDAR}.
Furthermore, methods with keyframe reuse or map reuse have an inherent advantage on the longer sequences (01, 02, and 07).
The keyframe re-creation for previously visited areas in \acs{MARS} leads to drift over time.
 
As GenZ-\acs{ICP} builds upon KISS-\acs{ICP}, it struggles with the unsteadier characteristics of handheld sensors, which is quite different from its typical automotive driving scenario.
Frequent obstruction of a small scan portion behind the sensor constitutes an additional challenge for Point-\acs{LIO} and the continuous-time methods.
Interestingly, the continuous-time methods perform slightly worse than the conventional methods on the first two sequences, with the only exception being SE-\acs{LIO}.
In these sequences, our method accumulates some drift in the parkland leading back to the parkland mound.

The sequence ``06\_Spin'' poses a particular challenge, as reflected by the high number of diverged solutions.
The sequence contains varying high rotational velocities of up to \SI{3.5}{\radian\per\second} where motion compensation is essential.
\Reffig{fig:nc_spin} shows the aggregation of every \num{25}th scan after compensation by Fast-\acs{LIO}2, \acs{DLIO}, and our approach.
Each point cloud contains around \num{2800000} points.
We subsample the clouds to \SI{5}{\centi\metre} resolution using CloudCompare for better alignment with the \ac{TLS} map.
After manual initialization, CloudCompare's \ac{ICP} fine registration aligns the subsampled clouds against the \ac{TLS} map with \num{50000} random sampled points.
We set the ``final overlap'' to \SI{90}{\percent} and enabled ``farthest point removal''.
Afterwards, we apply the estimated pose to the original cloud and compute point-to-plane errors \wrt the \ac{TLS} map to assess the compensation and registration quality.

Since no scan is visibly missaligned, we threshold the point-to-plane errors to \SI{0.25}{\metre} (resp. \SI{0.5}{\metre} and \SI{1}{\metre}) to reduce the influence of non-represented parts within the quad.
\Reftab{tab:nc_spin_rmse} shows that our map consistently has the lowest \ac{RMSE} before \acs{DLIO} and Fast-\acs{LIO}2 without any distinct red areas except for the growing tree and bushes.
Even if we compensate during registration in between the first and second iteration instead of prior to registration, we obtain a better result than the competing methods.
Fast-\acs{LIO}2 exhibits some spurious measurements and incorrectly compensated scans.

\bgroup
\newcolumntype{C}[1]{>{\centering\arraybackslash\hspace{0pt}}m{#1}}
\newcolumntype{A}{>{\raggedright\arraybackslash\hspace{0pt}}m{1.1cm}}
\newcolumntype{Z}{>{\centering\arraybackslash}X}
\newcolumntype{B}{>{\raggedright\arraybackslash}X}
\newcolumntype{L}{>{\raggedleft\arraybackslash}X}
\renewcommand{\arraystretch}{1.2}  %Vertical space: 1 is the default, change 
\begin{table}
\begin{threeparttable}[b]
\scriptsize
\centering
\caption{\acs{RMS} point-to-plane distance [\si{\centi\metre}] on ``06\_Spin'' \cite{ramezani2020newer}\tnote{$\ast$}}
\label{tab:nc_spin_rmse}
\begin{tabularx}{\columnwidth}{LZZZZ}
\toprule
$d_\mathrm{max}$ [\si{\centi\metre}] & Fast-LIO2 & DLIO & LIO-MARS & LIO-MARS\tnote{$\dagger$} \\
\midrule\midrule
\num{100} & 7.85 & 7.65 & \pa{7.59} & \pb{7.63} \\\hdashline
\num{50}  & 6.50 & 6.41 & \pa{6.31} & \pb{6.35} \\\hdashline
\num{25}  & 5.93 & 5.82 & \pa{5.73} & \pb{5.78} \\
\bottomrule
\end{tabularx}
\begin{tablenotes}
\item[$\ast$] for every 25th scan \wrt \ac{TLS} map. Lower values are better ($\downarrow$) with \pb{second} and \pa{best} highlighted.
\item[$\dagger$] Motion compensation on adaptive selected \acsp{surfel} between first and second registration iteration.
\end{tablenotes}
\end{threeparttable}
\end{table}
\egroup

The sequences of the extension~\cite{zhang2021newerext} showcase more variability \wrt the results in \reftab{tab:lio_newer_college}.
In the challenging ``Stairs'', the operator moved the handheld sensor setup from a hallway through a door into a narrow staircase and multiple flights of stairs upwards before heading back down again.
The close quarters make this sequence challenging as the ceiling or floor are temporarily not measured during turning.
This frequently leads to underconstrained directions in the optimization.
Furthermore, the \acs{LiDAR} measures the stairs from above and below.
This can erroneously pull both surfaces together if the map resolution is too coarse or incorrect correspondences are not rejected.

As part of their strategy to cope with the high amount of measurements per scan, most approaches use a single resolution per \acs{voxel}, \eg in conjunction with a \acs{voxel} filter and random or na{\"i}ve downsampling. 
As a result, Fast-\acs{LIO}2 performs well in most cases but diverges on the ``Stairs'' sequence.
A fine resolution (\SI{0.25}{\metre}) likely helps \acs{DLO} and \acs{DLIO} in this scenario, whereas a \SI{0.4}{\metre} to \SI{0.5}{\metre} resolution is more common.
KISS-\acs{ICP} and thus GenZ-\acs{ICP} follow the approach of \acs{CT-ICP} to store at most a fixed number of points per \acs{voxel}, which effectively allows the map to represent finer details too.
In contrast, SLICT2 and our method adjust data-dependently the size of a map cell dynamically.

For our map~(\refsec{sec:kf}), adapting the coarse cell size $c_0$ in combination with the resolution selection of MARS facilitates the handling of different environments.
Our system starts outside with its standard coarse \SI{4}{\metre} cell size such that the finest resolution $c_3$ has \SI{0.5}{\metre} cells.
After walking through the door into the staircase, \acs{LIO-MARS} sets $c_0$ to \SI{2}{\metre} (resp. $c_3=\SI{0.25}{\metre}$) as the mean \acs{surfel} distance reduces.
The process reverses in the end after walking out into the broader hallway.
Similarly, such adaptation happens when entering and leaving the narrow passage of the ``Cloister'' sequence.

Only two methods, \acs{DLIO} and \acs{LIO-MARS}, worked successfully on all sequences.
At the same time, our \acs{LIO-MARS} is the best performing continuous-time method over all sequences of the extension.

\subsection{Multi-Campus Dataset}
MCD~\cite{nguyen2024mcdviral} contains \num{12} handheld sequences recorded at the KTH and TUHH campuses using an Ouster OS1-64 \acs{LiDAR} and VectorNav VN-200 \acs{IMU}.
Six additional driving sequences were recorded at the NTU campus using an Ouster OS1-128 \acs{LiDAR} and VN-100 \acs{IMU}.

\bgroup
\newcolumntype{C}[1]{>{\centering\arraybackslash\hspace{0pt}}m{#1}}
\newcolumntype{A}{C{0.3cm}}
\newcolumntype{Z}{>{\centering\arraybackslash}X}
\newcolumntype{B}{>{\raggedright\arraybackslash}X}
\newcolumntype{L}{>{\raggedleft\arraybackslash}X}
\renewcommand{\arraystretch}{1.2}  %Vertical space: 1 is the default, change 
\begin{table}
\begin{threeparttable}[b]
\scriptsize
\centering
\caption{\acs{RMS}-\acs{ATE}~[\si{\metre}] evaluation for the Multi-Campus dataset~\cite{nguyen2024mcdviral}\tnote{$\star$}}
\label{tab:lio_mcd}
\setlength{\tabcolsep}{1pt}
\begin{tabularx}{\linewidth}{ABZZZZZZZZZZ}
\toprule
&  & \RotD{MARS} & \RotD{GenZ-ICP} & \RotD{Fast-LIO} & \RotD{Point-LIO} & \RotD{SLICT2} & \RotD{DLIO} & \RotD{SE-LIO} & \RotD{iG-LIO} & \RotD{RESPLE} & \RotD{LIO-MARS} \\
\midrule\midrule
\multicolumn{2}{l}{\ac{LIO}} & \xmark & \xmark & \cmark & \cmark & \cmark & \cmark & \cmark & \cmark & \cmark & \cmark \\
\multicolumn{2}{l}{CT\tnote{$\dagger$}} & \cmark & \xmark & \xmark & \xmark & \cmark & \xmark & \cmark & \xmark & \cmark & \cmark \\
\midrule
% rounded to 3:
\multirow{6}{*}{\RotD{KTH}}
& D\_06 & 6.997 & 0.753 & \pb{0.328} & 1.480 & \pc{0.576} & 3.330 & \tbfX & \pa{0.268} & 1.635 & 0.658\\
& D\_09 & \tbfX & \tbfX & \pa{0.126} & 0.377 & 0.387 & 8.463 & 6.195 & \pb{0.153} & 0.800 & \pc{0.361}\\
& D\_10 & 0.968 & \tbfX & \pb{0.392} & 2.016 & 1.125 & 3.027 & 26.068 & \pa{0.305} & 2.060 & \pc{0.551}\\
& N\_01 &10.076 & 0.947 & \pa{0.274} & 1.427 & \pc{0.467} & 3.674 & 4.254 & \pb{0.287} & 1.492 & 0.503\\
& N\_04 & \tbfX &\pc{0.316} & \pa{0.143} & 0.388 & 0.326 & 1.171 & 7.054 & \pb{0.181} & 0.829 & 0.362\\
& N\_05 & \tbfX & 1.081 & \pb{0.362} & 1.181 & 1.776 & 3.839 & 36.439 & \pa{0.293} & 1.581 & \pc{0.543}\\
 \midrule
\multirow{6}{*}{\RotD{NTU}}
& D\_01 & \tbfX & 5.934 & \pb{1.184} & 2.729 & 1.434 & 5.815 & \pc{1.295} & \pa{1.141} & 2.330 & 1.867\\
& D\_02 & \tbfX & 0.299 & 0.277 & \pa{0.098} & 0.392 & 0.340 & 0.275 & \pc{0.258} & \pb{0.125} & 0.424\\
& D\_10 & \tbfX & \pc{1.811} & 1.876 & 3.407 & 2.130 & 3.781 & \pa{1.231} & \pb{1.667} & 6.929 & 3.134\\
& N\_04 & \tbfX & \pa{1.222} & \pc{1.718} & 4.580 & 1.882 & 1.835 & 1.747 & \pb{1.541} & 4.286 & 2.793\\
& N\_08 & \tbfX & 2.026 & \pb{1.749} & 3.047 & 2.165 & 6.950 & 2.627 & \pa{1.710} & 6.137 & \pc{1.977}\\
& N\_13 & \tbfX & 1.009 & 0.900 & \pc{0.890} & 1.167 & 1.922 & 0.921 & \pb{0.756} & \pa{0.607} & 1.230\\
 \midrule
\multirow{6}{*}{\RotD{TUHH}}
& D\_02 & 1.909 & 0.580 & \pb{0.128} & 0.367 & 0.334 & 1.332 & 6.243 & \pa{0.122} & 0.488 & \pc{0.237}\\
& D\_03 &20.611 & 0.893 & \pb{0.757} & 0.829 & 1.375 & 2.973 & 8.398 & \pa{0.688} & \pc{0.812} & 1.175\\
& D\_04 &12.176 & 0.185 & \pb{0.098} & 0.202 & 0.149 & 0.707 & 7.267 & \pa{0.082} & 0.431 & \pc{0.138}\\
& N\_07 & 1.882 & 0.346 & \pa{0.137} & 0.390 & 0.330 & 4.830 & 3.576 & \pb{0.178} & 0.435 & \pc{0.225}\\
& N\_08 & 27.001 & \pc{0.701} & \pb{0.633} & 0.978 & 1.015 & 3.741 & \tbfX & \pa{0.619} & 4.435 & 1.301\\
& N\_09 & \tbfX & 0.173 & \pb{0.079} & \pc{0.125} & 0.127 & 0.389 & 3.896 & \pa{0.072} & 0.190 & 0.139\\
\midrule
\multicolumn{2}{l}{Fail [\%]} & 35.55 & 7.77 & \tbfZ & \tbfZ & \tbfZ & \tbfZ & 11.11 & \tbfZ & \tbfZ & \tbfZ\\
\multicolumn{2}{l}{Avg. Rank} & 9.39 & 5.39 & \pb{2.22} & 5.28 & 5.00 & 7.94 & 7.33 & \pa{1.50} & 6.22 & \pc{4.78}\\
\multicolumn{2}{l}{Overall} & 10. & 6. & \pb{2.} & 5. & 4. & 9. & 8. & \pa{1.} & 7. & \pc{3.}\\
\bottomrule
\end{tabularx}
\begin{tablenotes}
\item[$\star$] Algorithms are grouped by \acs{LO}/\acs{LIO} and ordered according to publication date. An ``X'' marks divergence. Lower values are better ($\downarrow$) with \pc{third}, \pb{second} and \pa{best} highlighted.
\item[$\dagger$] Continuous-time trajectory.
\end{tablenotes}
\end{threeparttable}
\end{table}
\egroup 

Most failures in \reftab{tab:lio_mcd} stem from zig-zagging after revisiting a previously seen area.
The zig-zagging leads to travel distances longer than \SI{125}{\percent} of the ground-truth distance.
Surprisingly, GenZ-\acs{ICP} performs very well in this setting.
The best results achieves iG-\acs{LIO}, followed by Fast-\acs{LIO}2 and our LIO-MARS in third place. 
Interesingly, LIO-MARS performs better in the handheld sequences then in the driving scenario.
Here, we noticed that most deviations are in the vertical direction.

\subsection{DRZ Living Lab}
This dataset~\cite{quenzel2021mars} was recorded onboard a flying DJI M210 v2 with an Ouster OS0-128 \acs{LiDAR}.
The DRZ Living Lab is located in a large industrial hall and accommodates facilities to test robots in difficult terrain or rescue scenarios.
A \ac{mocap} system measures the \ac{UAV} pose over time and provides the reference trajectories for evaluation.
This packed indoor environment leads to a high number of valid measurements compared to flights in outdoor environments with open skies.
Per scan, the ``F2'' sequence averages \num{92745} valid points, which equals to \SI{70.7}{\percent} of possible measurements.

\begin{figure}
  \centering
  \if\WithFigures1
  \resizebox{1.0\linewidth}{!}{\input{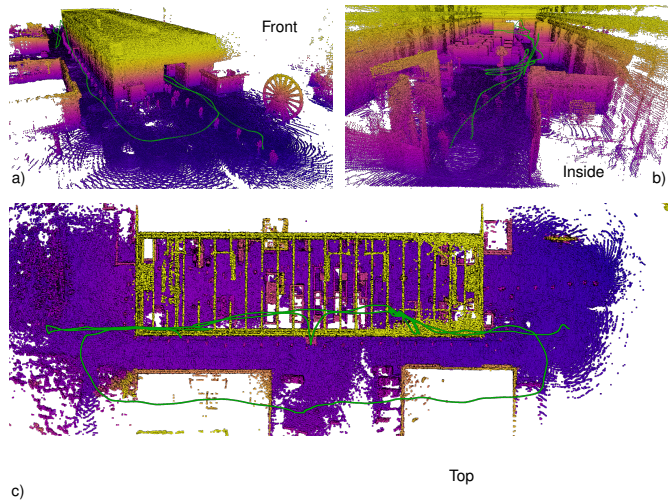}}
  \fi
\caption[DRZ Living Lab ``H3'']{Aggregated Pointcloud for ``H3'' from the DRZ Living Lab dataset with overlayed trajectory (green line), colored by height from low (blue) to high (yellow).
An \ac{UAV} with a \acs{LiDAR} started in front of the building [a)] and flew through the building [b)].
It traversed back through the alleyway and into the building a second time to land in the back.
The OS-0 \acs{LiDAR} provides dense 3D measurements within the hall.
Roof partially removed in c) for better visualization of the interior.
}
  \label{fig:quali_drz}
\end{figure}
 
\Reffig{fig:quali_drz} shows the reconstruction of ``H3'' using \acs{LIO-MARS}.
Here, the \ac{UAV} starts in the \ac{mocap} volume and flies through the hall and out the rear gate.
It then flies in the alleyway between the buildings towards the front entrance and back into the \ac{mocap} volume.
The forward direction along the alleyway is only partially constrained due to the \acs{LiDAR}'s limited range and the scene's geometry.

\Reftab{tab:lio_drz_halle} shows the results for all considered methods (see \refsec{sec:lio_eval}).
All algorithms perform reasonably well, even on the faster sequences with high rotational and linear velocities.
However, \acs{LIO-MARS} and Point-\acs{LIO} outperform the other \acp{LO} and \acp{LIO} algorithms.

Over all datasets, only \acs{DLIO} and \acs{LIO-MARS} did not fail on any sequence. %as shown in \reftab{tab:lio_ranks} and \reffig{fig:lio_ranks}.
In contrast, Point-\acs{LIO} failed in the more challenging ``06\_Spin'' and ''02\_Long'' whereas Fast-\acs{LIO}2 failed thrice in the ``Stairs'' sequences.

Although iG-\acs{LIO} and SE-\acs{LIO} performed well on the DRZ and original Newer College datasets, both diverged multiple times on the extension sequences.
Over all, \acs{LIO-MARS} achieves state-of-the-art performance on these \ac{UAV}, ground vehicle and hand-held sequences.

\bgroup
\newcolumntype{C}[1]{>{\centering\arraybackslash\hspace{0pt}}m{#1}}
\newcolumntype{A}{>{\raggedright\arraybackslash\hspace{0pt}}m{1.1cm}}
\newcolumntype{Z}{>{\centering\arraybackslash}X}
\newcolumntype{B}{>{\raggedright\arraybackslash}X}
\newcolumntype{L}{>{\raggedleft\arraybackslash}X}
\renewcommand{\arraystretch}{1.2}  %Vertical space: 1 is the default, change 
\begin{table}
\begin{threeparttable}[b]
\scriptsize
\centering
\caption{\acs{RMS}-\acs{ATE}~[\si{\metre}] evaluation for the DRZ Living Lab dataset\tnote{$\star$}}
\label{tab:lio_drz_halle}
\setlength{\tabcolsep}{1pt}
\begin{tabularx}{\linewidth}{AZZZZZZZZZZ}
\toprule
  & \RotD{MARS} & \RotD{GenZ-ICP} & \RotD{Fast-LIO} & \RotD{Point-LIO} & \RotD{SLICT2} & \RotD{DLIO} & \RotD{SE-LIO} & \RotD{iG-LIO} & \RotD{RESPLE} & \RotD{LIO-MARS} \\
\midrule\midrule
\ac{LIO}            & \xmark & \xmark & \cmark & \cmark & \cmark & \cmark & \cmark & \cmark & \cmark & \cmark \\
CT\tnote{$\dagger$} & \cmark & \xmark & \xmark & \xmark & \cmark & \xmark & \cmark & \xmark & \cmark & \cmark \\
\midrule
% Rounded to 3 digits as shown:
 H1   & 0.047 & 0.063 & \pb{0.019} & \pa{0.018} & 0.022 & 0.054 & 0.020 & \pb{0.019} & 0.029 & 0.020\\
 H2   & 0.048 & 0.057 & \pa{0.016} & \pc{0.018} & 0.025 & 0.052 & 0.022 & 0.019 & 0.029 & \pb{0.017}\\
 H3   & 3.760 & 0.324 & 0.031 & \pb{0.023} & 0.389 & 0.056 & \tbfX & \pc{0.025} & 0.034 & \pa{0.022}\\
 3P-S & 0.018 & 0.024 & 0.012 & \pa{0.008} & 0.011 & 0.014 & 0.012 & 0.011 & \pb{0.010} & \pb{0.010}\\
 3P-M & 0.037 & 0.048 & \pc{0.013} & \pa{0.010} & 0.016 & 0.036 & \pc{0.013} & 0.016 & 0.014 & \pa{0.010}\\
 3P-F & 0.056 & \tbfX & 0.039 & \pa{0.016} & \pc{0.020} & 0.061 & 0.022 & \pb{0.019} & 0.029 & 0.021\\
 S1   & 0.051 & 0.061 & 0.039 & \pa{0.025} & 0.029 & 0.048 & \pc{0.028} & 0.029 & \pc{0.028} & \pb{0.026}\\
 M1   & 0.079 & 0.107 & 0.044 & \pb{0.034} & \pc{0.036} & 0.075 & 0.040 & \pc{0.036} & 0.040 & \pa{0.032}\\
 F2   & 0.103 & 0.179 & 0.060 & \pa{0.054} & \pc{0.057} & 0.110 & 0.058 & 0.058 & 0.062 & \pb{0.056}\\
 F3   & 0.068 & 0.096 & 0.041 & \pb{0.027} & \pc{0.030} & 0.068 & \pc{0.030} & \pc{0.030} & 0.033 & \pa{0.025}\\
\midrule
Fail [\%] &\tbfZ & 10.00 & \tbfZ & \tbfZ & \tbfZ & \tbfZ & 10.00 & \tbfZ & \tbfZ & \tbfZ\\
Avg. Rank & 8.50 & 9.70 & 5.00 & \pa{1.50} & 4.70 & 8.20 & 4.80 & \pb{3.60} & 5.30 & \pb{2.00}\\
Overall & 9. & 10. & 6. & \pa{1.} & 4. & 8. & 5. & \pc{3.} & 7. & \pb{2.}\\
\bottomrule
\end{tabularx}
\begin{tablenotes}
\item[$\star$] Algorithms are grouped by \acs{LO}/\acs{LIO} and ordered according to publication date. An ``X'' marks divergence. Lower values are better ($\downarrow$) with \pc{third}, \pb{second} and \pa{best} highlighted.
\item[$\dagger$] Continuous-time trajectory.
\end{tablenotes}
\end{threeparttable}
\end{table}
\egroup

\bgroup
\newcolumntype{C}[1]{>{\centering\arraybackslash\hspace{0pt}}m{#1}}
\newcolumntype{Z}{>{\centering\arraybackslash}X}
\newcolumntype{L}{>{\raggedleft\arraybackslash}X}
\newcolumntype{A}{C{0.25cm}}
\newcolumntype{B}{C{0.45cm}}
\newcolumntype{D}{C{0.5cm}}
\renewcommand{\arraystretch}{1.1}  %Vertical space: 1 is the default, change 
\begin{table}[t!]
\begin{threeparttable}[b]
\scriptsize
\centering
\caption{Ablation on spline parameters for ``F2''\tnote{$\star$}}
\label{tab:lio_ablation}
\setlength{\tabcolsep}{1pt}
\begin{tabularx}{\columnwidth}{ADBAZZZZZZ}
\toprule
\multicolumn{4}{c}{Spline} & \acs{RMS}-\acs{ATE} & Map Emb. & Comp. & Spline Init. & Reg. & Avg. Time\\
$N$ & $\abs{\mathcal{W}}$ & $\abs{\mathcal{X}}$ & $O$ & [\si{\metre}] ($\downarrow$) & [\si{\milli\second}] ($\downarrow$) & [\si{\milli\second}] ($\downarrow$) & [\si{\milli\second}] ($\downarrow$) & [\si{\milli\second}] ($\downarrow$) & [\si{\milli\second}] ($\downarrow$)\\
\midrule\midrule
% 1 knot per scan
 2 & 3 & 3 & 8 &     0.0945  &    {10.9}&\pb{15.9}  &  9.2 &    17.0 &    59.2 \\
 2 & 4 & 4 & 8 &     0.0689  &     11.0    & 17.9   & 12.9 &    21.9 &    68.9 \\
 2 & 5 & 5 & 8 &     0.0701  & \pc{10.8}   & 30.7   & 17.2 &    27.5 &    92.4 \\
 2 & 6 & 6 & 8 &     0.0713  &    {10.9}   & 31.5   & 21.1 &    32.9 &   102.8\\
\midrule
 3 & 2 & 2 & 2 &     0.0659  &     17.8 &\pa{15.3}&   {7.6}&\pc{13.1}&    59.7 \\
 3 & 2 & 2 & 4 &     0.0567  &     12.2 &\pc{16.4}&\pb{6.5}&\pa{11.0}&\pb{51.6}\\
 3 & 2 & 2 & 8 &     0.0567  & \pc{10.8}   & 18.0 &\pa{6.3}&\pa{11.0}&\pa{51.5}\\
\midrule
 3 & 3 & 3 & 2 &    {0.0562} &     17.9    & 16.6   & 10.0 &    18.0 &    68.6 \\
 3 & 3 & 3 & 4 & \pb{0.0560} &     12.4    & 17.7   &  8.8 &    16.9 &    61.4 \\
 3 & 3 & 3 & 8 &     0.0562  & \pc{10.8}   & 19.1   &  9.4 &    17.1 &    62.4 \\
\midrule
% 2 knots per scan
 3 & 3 & 6 & 2 &    {0.0572} &     17.8    & 16.5   & 12.5 &    18.7 &    71.8\\
 3 & 3 & 6 & 4 & \pc{0.0561} & \pb{10.2}   & 18.5   &\pc{7.3}&  17.4 &    60.0\\
 3 & 3 & 6 & 8 & \pc{0.0561} &  \pa{9.1}   & 17.3   & 10.3 &    15.8 &    58.5\\
\midrule 
% 1 knot per scan
 3 & 4 & 4 & 8 &     0.0562  & \pc{10.8}   & 20.1   & 13.7 &    22.0 &    72.6\\
 3 & 5 & 5 & 8 & \pa{0.0559} & \pc{10.8}   & 36.6   & 17.1 &    28.3 &    99.1\\
 3 & 6 & 6 & 8 &     0.0562  &    {10.9}   & 37.8   & 21.1 &    33.3 &   109.5\\
\midrule
% 2 knots per scan, N4
 4 & 3 & 6 & 4 &     0.1216  &     12.8    & 20.3   & 13.8 &    20.4 &    78.2\\
 4 & 3 & 6 & 8 &     0.1026  &     11.1    & 21.9   & 15.3 &    20.3 &    78.6\\
\bottomrule
\end{tabularx}
\begin{tablenotes}
\item [$\star$] ``F2'' sequence in the DRZ Living Lab. Lower values are better ($\downarrow$) with \pc{third}, \pb{second} and \pa{best} highlighted.
\end{tablenotes}
\end{threeparttable}
\end{table}
\egroup

\subsection{Ablation}
In the previous evaluations, \acs{LIO-MARS} uses a third-order spline ($N=3$), three scans in the current scan window $\mathcal{W}$, and six optimizable knots.
Our surfel compensation applies only to the new scan whereas the two previous scans in the current window are already compensated.

To better understand the effect of our design decisions, we evaluate different spline parameters on the ``F2'' sequence captured within the \ac{mocap} of the DRZ Living Lab.
\Reftab{tab:lio_ablation} reports the results including the resp. timing for individual components.

Allocating all resources can become dangerous on a robotic system as multiple processes will interfere with each other.
Hence, we limit the number of threads to four which is sufficient to process each map level in parallel.
This leaves enough computational resources for mapping, navigation, and robot control.
However, we expect a further runtime reduction when all threads can be used.

Increasing the number of segments $O$ reduces the time required to embed a point cloud into the map.
When more segments subdivide a scan, the number of \acsp{surfel} per segment decreases.
This, in turn, leads to faster access to individual cells since we store each segment in a separate small hash map.
Evidently, this time remains constant independent of the number of scans ($\abs{\mathcal{W}}$) within the sliding window $\mathcal{W}$, as shifting the window removes the oldest scan and adds a new scan.

Overall, our system performs best \wrt \ac{RMS}-\ac{ATE} with a spline of order $N=3$ and an equal number of scans or more in the sliding window  $\abs{\mathcal{W}} \geq N$.
Optimizing fewer than $N$ knots can lead to oscillations and reduce accuracy.
Our current implementation limits the number of optimizable knots to six, which sets the maximum scan window size to \num{6}.
Alternatively, three scans with two knots per scan allow more flexible trajectories, \eg as used by SLICT2.
This reduces the temporal knot distance from $\approx\SI{100}{\milli\second}$ to $\SI{50}{\milli\second}$.
However, our results show not much of an additional benefit \wrt accuracy for a reduced temporal knot distance.

As expected, a larger sliding window increases the overall computation time for the compensation, initialization, and registration.
Moreover, it extends the time until a previously unseen area becomes part of the map, which can degrade performance in larger or obstructed environments.

In comparison with \acs{MARS}~\cite{quenzel2021mars}, \acs{LIO-MARS} can optimize more iterations, five instead of three, in less time.
Moreover, our system increases the map's resolution while reducing its computational cost.

\bgroup
\newcolumntype{Z}{>{\centering\arraybackslash}X}
\renewcommand{\arraystretch}{1.1}  %Vertical space: 1 is the default, change 
\begin{table*}[t!]
\begin{threeparttable}[b]
\scriptsize
\centering
\caption{Ablation on IMU integration on the Newer College Extension~\cite{zhang2021newerext}\tnote{$\ast$}}
\label{tab:lio_ablation_rel}
\begin{tabularx}{\linewidth}{ZZZZZZZZZZZZZZ}
\toprule
IMU & \RotD{Math-E} & \RotD{Math-M} & \RotD{Math-H} & \RotD{Mine-E} & \RotD{Mine-M} & \RotD{Mine-H} & \RotD{Quad-E} & \RotD{Stairs} & \RotD{Quad-M} & \RotD{Quad-H} & \RotD{Park} & \RotD{Cloister} \\
\midrule \midrule
$\diamond$ 
& \pa{0.1460} & \pa{0.1407} &     0.1023  &     0.0431  & \pa{0.0422} & \pa{0.0559} &     0.0714  & \pb{0.1189} &     0.0631  &     0.0622  &     0.2935  &     0.0881  \\
$\dagger$
& 0.1895      & 0.1559      & \pa{0.0938} &     0.0419  & \pb{0.0423} &     0.0569  &     0.0695  &     0.1323  & \pb{0.0621} &     0.0607  & \pb{0.2883} &     0.0710  \\
$\ddagger$
& \pb{0.1574} & \pb{0.1411} &     0.1118  & \pb{0.0411} &     0.0436  & \pb{0.0563} & \pa{0.0675} & \pa{0.1026} & \pa{0.0608} &     0.0633  & \pa{0.2882} & \pa{0.0691} \\
\midrule
$\odot$
& 0.1895      &    0.1572   & \pb{0.0961} &     0.0412  &     0.0428  &     0.0590  & \pb{0.0690} &     0.1256  &     0.0629  & \pb{0.0606} &     0.3115  & \pb{0.0696} \\
$\boxdot$
& 0.1895      &    0.1420   &     0.0983  & \pa{0.0405} &     0.0431  &     0.0598  &     0.0695  &     0.1275  &     0.0625  & \pa{0.0604} &     0.3328  &     0.0733  \\
\bottomrule
\end{tabularx}
\begin{tablenotes}
\setlength{\columnsep}{0.4cm}
\setlength{\multicolsep}{0cm}
  \begin{multicols}{2}
\item[$\ast$] Lower values are better ($\downarrow$) with \pb{second} and \pa{best} highlighted.
\item[$\diamond$] Using only raw IMU constraints (\refeq{eq:dgyr} and \refeq{eq:dacc}).
\item[$\dagger$] Preintegrating all IMU measurements (\refeq{eq:dpos}, \refeq{eq:drot} and \refeq{eq:dvel}).
\item[$\ddagger$] Preintegrate IMU measurements except for last one (\refeq{eq:imu}).
\item[$\odot$] Same as $\ddagger$, w/o bias estimation.
\item[$\boxdot$] Same as $\odot$, w/o gravity estimation.
\end{multicols}
\end{tablenotes}
\end{threeparttable}
\end{table*}
\egroup

\subsubsection{Relative Motion Constraints}
So far, our experiments use the mixture of preintegrated and raw \acs{IMU} measurements from \refsec{sec:nsc}.
Here, we leverage the ``F2'' sequence from the DRZ dataset.
With the standard configuration, \acs{LIO-MARS} achieves an \ac{RMS}-\ac{ATE} of \SI{0.056}{\metre}.
Using only raw \acs{IMU} without preintegration results in an \acs{RMSE} of \SI{0.057}{\metre}.
\Reftab{tab:lio_ablation_rel} shows that our combined \acs{IMU} mixture \refeq{eq:imu} performs best on the Newer College extension~\cite{zhang2021newerext}.

% for single Rel Ori w/ undistort, no preint \acs{IMU}
On ``F2'' of the DRZ dataset, replacing the \acs{IMU} measurements with a single relative \acs{UAV} pose measurement $\Delta T$ during optimization increases the \acs{RMS}-\acs{ATE} to \SI{0.093}{\metre} in our experiment.
We attribute this increase to inaccuracies and sudden jumps in the height estimate when flying above the NIST crates or their side walls.
The DJI M210v2 provides a magnetometer-based orientation, a horizontal position from \ac{VIO} and an ultrasonic height measurement above ground.
Together, the \acs{UAV}'s odometry has an \acs{RMS}-\acs{ATE} of \SI{0.56}{\metre} including clear deviations in horizontal directions.

We already filter out spikes with high vertical velocities that are inconsistent with the motion direction.
Moreover, our system only includes relative motion and no absolute position.
However, smaller steps from objects on the ground still impact the result.
Comparing the measured height with the local map height can potentially reduce the discrepancy, but remains out of the scope of this work.

If we instead supply only a single relative orientation measurement $\Delta R$ per scan, we achieve an \acs{RMS}-\acs{ATE} of \SI{0.066}{\metre}.
According to \reftab{tab:lio_drz_halle}, this variant would place it directly behind full \acs{LIO} methods.
Hence, we can still obtain resonable results in the absence of an \acs{IMU} or improve resilience in underconstrained environments.

\subsection{Motion Compensation}
Compensating all \acsp{surfel} within the sliding window prior to registration is a time-costly process.
Hence, we compare against compensating only the adaptively selected \acsp{surfel}.
Additionally, we test the effect of the de-skewing and using the full covariance versus the adaptation for planar \acsp{surfel}~\refeq{eq:planar_cov}.
\Reftab{tab:lio_comp_ablation} highlights the benefit of adapting the covariances of planar \acsp{surfel}.
Furthermore, compensating selected \acsp{surfel} is much faster without losing accuracy.

\bgroup
\newcolumntype{C}[1]{>{\centering\arraybackslash\hspace{0pt}}m{#1}}
\newcolumntype{A}{C{0.8cm}}
\newcolumntype{B}{C{1.3cm}}
\renewcommand{\arraystretch}{1.1}  %Vertical space: 1 is the default, change 
\begin{table}[t!]
\begin{threeparttable}[b]
\scriptsize
\centering
\caption{Ablation on motion compensation for ``F2''\tnote{$\ast$}}
\label{tab:lio_comp_ablation}
\setlength{\tabcolsep}{1pt}
\begin{tabularx}{\linewidth}{AAAABBBB}
\toprule
\multicolumn{2}{c}{Comp.} & \multicolumn{2}{c}{Full} & \acs{RMS}-\acs{ATE} & Comp. & Reg. & Avg. Time\\
Pre. & Sel. & De-Skew & Cov. & [\si{\metre}] ($\downarrow$) & [\si{\milli\second}] ($\downarrow$) & [\si{\milli\second}] ($\downarrow$) & [\si{\milli\second}] ($\downarrow$)\\
\midrule\midrule
% 2 knots per scan
 \xmark & \xmark & \xmark & \cmark &   {0.0577} &  \pa{0.0} &    18.4 &    48.5\\
 \xmark & \xmark & \xmark & \xmark &   {0.0565} &  \pa{0.0} &\pa{16.1}&\pa{42.0}\\
 \xmark & \xmark & \cmark & \cmark &   {0.0572} & \pc{$<$0.1} &\pc{18.2}&\pc{48.0}\\
 \xmark & \xmark & \cmark & \xmark &   {0.0567} & \pc{$<$0.1} &\pa{15.6}&\pb{37.8}\\
\midrule
 \cmark & \xmark & \xmark & \cmark &   {0.0570} &     16.5  &    19.0 &    65.9 \\
 \xmark & \cmark & \xmark & \cmark &   {0.0570} &      4.4  &    19.0 &    53.3 \\
 \cmark & \xmark & \xmark & \xmark &   {0.0562} &     16.4  &    18.5 &    64.7 \\
 \xmark & \cmark & \xmark & \xmark &   {0.0562} &      4.4  &    19.0 &    52.9 \\
\midrule
 \cmark & \xmark & \cmark & \cmark &   {0.0569} &     18.2  &    18.7 &    67.1 \\
 \xmark & \cmark & \cmark & \cmark &   {0.0569} &      4.9  &    18.9 &    53.6 \\
 \cmark & \xmark & \cmark & \xmark &\pb{0.0561} &     18.5  &    17.4 &    60.0 \\
 \cmark\tnote{$\dagger$} & \xmark & \cmark & \xmark &\pa{0.0560} & 10.5 &  16.2 &    49.2 \\
 \xmark & \cmark & \cmark & \xmark &\pb{0.0561} &      4.2  &    16.6 &    43.4 \\
\bottomrule
\end{tabularx}
\begin{tablenotes}
\item[$\ast$] Statistics for varying motion compensation on the ``F2'' sequence in the DRZ Living Lab. Lower values are better ($\downarrow$) with \pc{third}, \pb{second} and \pa{best} highlighted.
\item[$\dagger$] Compensate only the new scan.
\end{tablenotes}
\end{threeparttable}
\end{table}
\egroup

To show the effect of our proposed \ac{UT} for motion compensation, we use the \ac{KLD}~\cite{hershey2007kld} $D_\mathrm{KL}\left(\mathcal{N}_q\lVert \mathcal{N}_r\right)$.
The reference distributions $\mathcal{N}_r$ stem from \acs{surfel} mean $\bm{\mu}_s$ and covariance $\bar{\Sigma}_s$ of compensated points using \refeq{eq:compPoint} after registration. 
Per \acs{surfel} $s\in\mathcal{S}_l$, we compute the \ac{KLD} for the raw points $\left(\mathcal{N}_{q_0}\right)$,
with \acs{IMU} pre-orientation $\left(\mathcal{N}_{q_1}=\left({\bm{\mu}}_s,{\Sigma}_s\right)\right)$ and after surfelwise compensation $\left(\mathcal{N}_{q_2}=\left(\bar{\bm{\mu}}_s,\bar{\Sigma}_s\right)\right)$.
Again, we evaluate on the rotation-heavy sequence ``06\_spin'' of the Newer College dataset~\cite{ramezani2020newer}.
Compared to the \acsp{surfel} from raw points ($\mathcal{N}_{q_0}$), our compensation reduces the \ac{KLD} \wrt  $\mathcal{N}_r$ for \SI{76.0}{\percent} of the \acsp{surfel}.
On average, the median \ac{KLD} improves per scan in \SI{94.6}{\percent} of the cases, with a mean reduction to \SI{21.4}{\percent}.

The benefit is less pronounced against the pre-oriented points.
After our compensation, \SI{53.4}{\percent} of the \acsp{surfel} have a lower \ac{KLD} \wrt $\mathcal{N}_r$.
On average, the median \ac{KLD} improves per scan in \SI{74.5}{\percent} of the cases with a mean reduction to \SI{91.9}{\percent}.
Interestingly, the covariances $\bar{\Sigma}_s$ are more similar in \SI{65.4}{\percent} of the cases versus \SI{57.1}{\percent} for the mean.

\subsection{Influence of Symmetry}

We compare the timing for the previous \ac{GMM}~\cite{quenzel2021mars} against a variant exploiting symmetry, \eg in inverse computation~\refeq{eq:sym_inv}, and finally, our vectorized version~(\refsec{sec:kron_sym}).
All versions run on sequence ``F2'' with up to four threads and five iterations during registration.
In \reftab{tab:gmm_timing}, we report the timing for the association, gradient computation, and evaluation separately, as the \ac{GMM} computation is split between them.
As expected, exploiting the symmetry accelerates evaluation and gradient computation.
Our $vec_L$ version outperforms both other variants and reduces the time spent on association.

\bgroup
\newcolumntype{C}[1]{>{\centering\arraybackslash\hspace{0pt}}m{#1}}
\newcolumntype{A}{>{\raggedright\arraybackslash\hspace{0pt}}m{1.1cm}}
\newcolumntype{Z}{>{\centering\arraybackslash}X}
\newcolumntype{B}{>{\raggedright\arraybackslash}X}
\newcolumntype{L}{>{\raggedleft\arraybackslash}X}
\renewcommand{\arraystretch}{1.2}  %Vertical space: 1 is the default, change 
\begin{table}
\begin{threeparttable}[b]
\scriptsize%\tiny
\centering
\caption{Timing Ablation [\si{\milli\second}] for optimized operations\tnote{$\ast$}} %~\refsec{sec:embedd}
\label{tab:splatting_timing}
\begin{tabularx}{\columnwidth}{C{2.5cm}ZZZ}
\toprule
%\diagbox[height=2\line]{Seq.}{Method} 
Method \textbackslash Step & Assoc. & Eval. & Grad. \\
\midrule\midrule
orig.   & 3.0975 & 4.6433 & 3.4701 \\\hdashline
w/ sym.    & \pb{3.0462} & \pb{2.7356} & \pb{2.6307} \\\hdashline
w/ $vec_L$ & \pa{1.8741} & \pa{2.1111} & \pa{1.9326} \\
\midrule
\end{tabularx}
\label{tab:gmm_timing}
\begin{tabularx}{\columnwidth}{%@{\extracolsep{\fill}}
C{2.5cm}ZZZZ}
 & \multicolumn{2}{c}{Single\tnote{1}} & \multicolumn{2}{c}{Parallel\tnote{1}} \\
Threads \textbackslash Order & L$\rightarrow$ P & P$\rightarrow$ L & L$\rightarrow$ P & P$\rightarrow$ L \\
\midrule\midrule
1 & 3475 & 3488 & \pa{1053} & \pb{1092} \\\hdashline 
4 & 1034 & 1124 & \pa{322} & \pb{479} \\
\bottomrule
\end{tabularx}
\begin{tablenotes}
\item[$\ast$] Avg. Timing [\si{\milli\second}] for \ac{GMM} computation per scan for \num{5} iterations on seq. ``F2''. Lower values are better ($\downarrow$) with \pb{second} and \pa{best} highlighted.
\item[$\dagger$] Timing [\si{\milli\second}] for splatting of \num{14.195e6} points (P) on \num{4} levels (L). Lower values are better ($\downarrow$) with \pb{second} and \pa{best} highlighted.
\item [1] Single~\cite{quenzel2021mars} and Parallel (\refsec{sec:embedd})
\end{tablenotes}
\end{threeparttable}
\end{table}
\egroup

We further evaluate the timing for our optimized splatting operation from \refsec{sec:embedd} on \num{14.195e6} points distributed over an area of \SI{200}{\metre\squared}.
Each point is splatted onto all four levels of our \acs{surfel} map with only a single thread or up to four threads in parallel. 
We also check whether to iterate first over all points and the level second (P$\rightarrow$L) or vice versa.
\Reftab{tab:splatting_timing} shows that our optimized implementation with a single thread provides a similar speedup to using four threads for regular splatting.
Moreover, our solution still profits from the higher thread count and from first processing all levels per point (L$\rightarrow$P).

\subsection{Further Qualitative Examples}
\begin{figure}[t]
  \centering
  \if\WithFigures1
  \resizebox{1.0\linewidth}{!}
  {\input{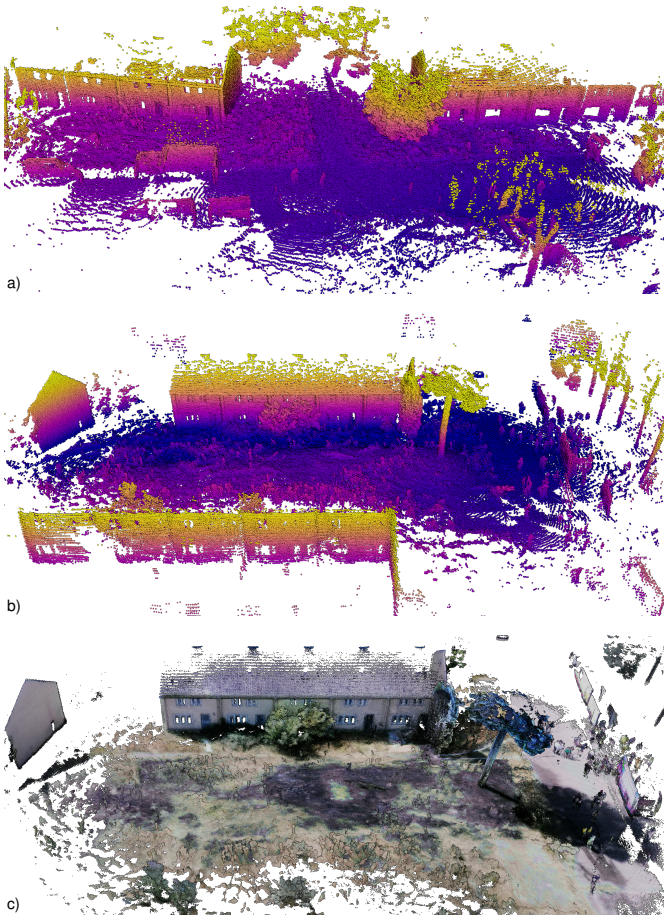}}
  \fi
  \caption[Fire fighting example]{Aggregated pointclouds from separate locations recorded during a fire fighting exercise with the fire brigade of the district Viersen at the abandoned Javelin Barracks in Elmpt, Germany.
The visible row houses consist of five attached units with a combined width of $\approx\SI{30}{\metre}$.
Map in a), b) are colored by height from low (blue) to high (yellow).
The map in c) corresponds to b) and is colored by projection into a $360^\circ$ camera.
  }
  \label{fig:quali_viersen}
\end{figure}
Recently, we deployed our \ac{UAV} during a forest fire training exercise by the fire brigade of the district Viersen at the abandoned Javelin Barracks in Elmpt, Germany.
\Reffig{fig:quali_viersen} shows the height-colored point clouds from multiple flights at separate locations.
Each $\approx\SI{30}{\metre}$ wide row house consists of five attached units in a state of disrepair with many windows and doors missing.
During the exercise, the trainers set the bushes in front of the left houses in \reffig{fig:quali_viersen} a) and close to the tree in b) on fire.
After the brush fire was extinguished, our \acs{UAV} inspected the scene and directly provided an overview for the firefighters to assess the extent of the burned vegetation.
\Reffig{fig:quali_viersen} b) shows the reconstructed \acs{LiDAR} map.

\section{Summary}
In this work, we presented a novel continuous-time \acl{LIO} called \acs{LIO-MARS} and verified our key claims.
Our system extends \acs{MARS}~\cite{quenzel2021mars} by introducing a non-uniform B-spline, active motion compensation, and tight coupling of \acs{LiDAR} and \acs{IMU}.

The non-uniform continuous-time B-spline adapts better to variations in scan timing without introducing additional delays.
For this, we propose a novel strategy for the temporal knot placement to better represent the sliding window used for state estimation at runtime.
As a consequence, we achieve better numerical stability during optimization compared to its uniform counterpart.

A temporal separation into intra-scan segments facilitates motion compensation at optimization time.
Meanwhile, an \acl{UT} compensates individual \acsp{surfel}, which leads to more concise covariances for measured surfaces. 

Leveraging complementary motion estimates further improves consistency and robustness.
For this, we derived the analytic Jacobians for relative motion constraints, like preintegrated \acs{IMU} measurements or relative poses.

Our system generates data-dependent new keyframes from the motion-compensated scans and selects a subset of keyframes for local map generation based on a \acs{surfel}-dependent overlap.
Moreover, the local map scale adapts to the measured distances in the sensor's vicinity to better represent open or narrow environments.

We modified the embedding of points into the permutohedral lattice using an inverse permutation and parallel min-max-sorting coupled with efficient \acs{SIMD} instructions.
In this way, our improved embedding achieves a $\approx\num{3.3}$-fold increase in throughput without changing the thread count.
Furthermore, we exploit the inherent symmetry within \acs{surfel} covariances and the \acl{GMM} computation and rephrase both using Kronecker sums and products for more efficient calculation.
As a consequence, the real-time performance per iteration improves on average by a factor of two.

In total, we tested \num{14} other current \ac{LO} and \ac{LIO} algorithms on multiple \acs{LiDAR} datasets in \ac{UAV}, ground vehicle and handheld sensor scenarios.
Overall, our system is the best-performing continuous-time method delivering state-of-the-art performance in real-time.

\section*{Acknowledgments}
We would like to express our gratitude to the fire brigade of the district Viersen for the oportunity to participate in their forest fire training exercises.
\if\WithAuthorInfo1
This work has been supported by the German Federal Ministry of Education and Research (BMBF) in the projects ``Kompetenzzentrum: Etablierung des Deutschen Rettungsrobotik-Zentrums (E-DRZ)'', grant 13N16477, and ``UMDenken: Supportive monitoring of turntable ladder operations for firefighting using IR images'', grant 13N16811.
\fi

\section*{References}
\printbibliography[heading=none]

\if\WithAuthorInfo1
\newpage
\section*{Biography Section}
\vspace{-22pt}
\begin{IEEEbiography}[{\includegraphics[width=1in,clip,keepaspectratio]{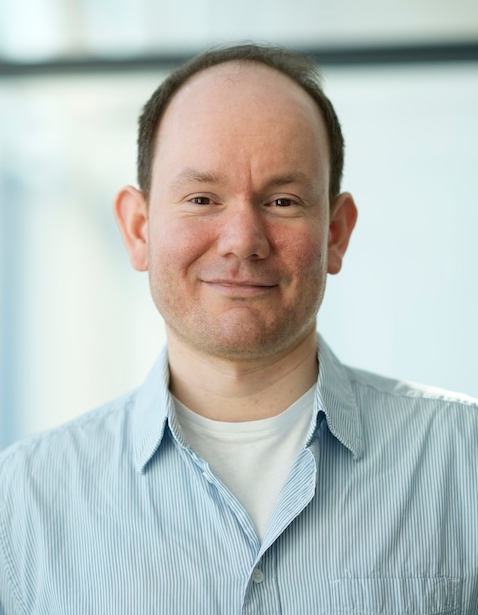}}]{Jan Quenzel}
received his M.Sc. degree in Computer Science from the University of L{\"u}beck in 2015.
Since August 2015, he is a member of the Autonomous Intelligent Systems Group at the University of Bonn.
His research focuses on real-time multi-modal odometry for unmanned aerial vehicles, scene reconstruction, and sensor calibration.
\end{IEEEbiography}
\vspace{-22pt}
\begin{IEEEbiography}[{\includegraphics[width=1in,height=1.25in,clip,keepaspectratio]{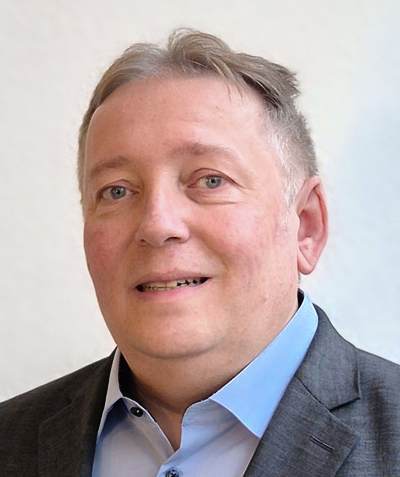}}]{Sven Behnke}
is since 2008 professor for Autonomous Intelligent Systems at the 
University of Bonn and director of the Computer Science Institute VI -- Intelligent Systems and Robotics. 
He heads the research area Embodied AI of the Lamarr Institute for Machine Learning and Artificial Intelligence and is ELLIS Fellow.
Prof. Behnke received his M.S. degree in Computer Science (Dipl.-Inform.) in 1997 from Martin-Luther-Universit\"at Halle-Wittenberg. 
In 2002, he obtained a Ph.D. in Computer Science (Dr. rer. nat.) from Freie Universit\"at Berlin.
He spent the year 2003 as postdoctoral researcher at the International Computer Science Institute, Berkeley, CA. From 2004 to 2008, Prof. Behnke headed the Humanoid Robots Group at Albert-Ludwigs-Universit\"at Freiburg. His research interests include cognitive robotics, computer vision, and machine learning. 
\end{IEEEbiography}
\fi

%\balance
\vfill

\end{document}